\documentclass[review]{elsarticle}
\pdfoutput=1
\usepackage{lineno,hyperref}
\usepackage{amsmath}
\usepackage{cleveref}[2012/02/15]
\crefformat{footnote}{#2\footnotemark[#1]#3}
\usepackage{booktabs}
\usepackage{multirow}
\usepackage{amssymb}
\usepackage{caption}
\usepackage{xcolor}
\usepackage{graphicx,wrapfig,lipsum}
\usepackage[export]{adjustbox}
\usepackage{soul}

\modulolinenumbers[5]

\journal{Pattern Recognition Journal}

\begin{document}

\begin{frontmatter}

\title{Query-Guided Networks for Few-shot Fine-grained Classification and Person Search}

\author[tum]{Bharti Munjal\corref{mycorrespondingauthor}}

\cortext[mycorrespondingauthor]{Corresponding author}
\ead{munjalbharti@gmail.com}

\author[sap]{Alessandro Flaborea}
\author[magic]{Sikandar Amin}
\author[tum,google]{Federico Tombari}
\author[sap]{Fabio Galasso}

\address[tum]{Department of Informatics, Technical University of Munich, Germany}
\address[sap]{Department of Computer Science, Sapienza University of Rome, Italy}
\address[magic]{Magic Leap Zurich, Switzerland}
\address[google]{Google Zurich, Switzerland}

\begin{abstract}
Few-shot fine-grained classification and person search appear as distinct tasks and literature has treated them separately.  But a closer look unveils important similarities: both tasks target categories that can only be discriminated by specific object details; and the relevant models should generalize to new categories, not seen during training.

We propose a novel unified Query-Guided Network (QGN) applicable to both tasks. QGN consists of a Query-guided Siamese-Squeeze-and-Excitation subnetwork which re-weights both the query and gallery features across all network layers, a Query-guided Region Proposal subnetwork for query-specific localisation, and a Query-guided Similarity subnetwork for metric learning.

QGN improves on a few recent few-shot fine-grained datasets, outperforming other techniques on CUB by a large margin. QGN also performs competitively on the person search CUHK-SYSU and PRW datasets, where we perform in-depth analysis.

\end{abstract}


\end{frontmatter}

%\linenumbers

\section{Introduction}
\label{intro}

Few-shot fine-grained classification and person search share important similarities, as they both require paying attention to the details, e.g.\ what distinguishes a person from other people, or a bird from other possibly similar races. Both fields have progressed largely in recent years~\cite{Kim_2021_CVPR,Wang_2020_CVPR}.
Few-shot learning eases the burden of large data collections when generalizing to new unseen (possibly rare) classes. Person search is useful for video surveillance, long term tracking and person verification.
Both tasks face the similar challenges of background clutter, illumination and viewpoint changes, occlusions, image blur and distortions, including non-rigid deformations of the object body pose~\cite{xiao2017joint,tang2020revisiting}.

Person search is the task of finding a specific person, as provided by a single query image, within a gallery image. It consists of localization within the gallery (detection) and re-identification (classification based on the single query example). Few-shot learning similarly stands for recognizing the queried object, either classifying or detecting, typically from a single or multiple (i.e., five) examples (1- and 5-shot learning). Fine-grained classification specifically describes the challenge of recognizing an object (bird, aircraft, dog etc.) from a few details (the shape of the beak, the pattern on the wings etc.). Person search is therefore a one-shot fine-grained classification task, which includes detection. Note that in few-shot fine-grained classification the \emph{query}-\emph{gallery} pair is termed \emph{support}-\emph{query} respectively, especially confusing for the role of the query. Throughout this work, we adopt the person search terminology and search a \emph{query} person or object within a \emph{gallery} image. See Sec.~\ref{sec:methodoverview} for more details.

We propose a novel unified Query-Guided Network (QGN) to address both person search and few-shot fine-grained classification. 
Query guidance is novel and stands for processing the query and gallery images jointly, with a Siamese network design and query-gallery interaction modules.
By contrast, prior literature in person search~\cite{xiao2017joint,Chen_2020_CVPR,Dong_2020_CVPR} and few-shot learning~\cite{snell2017nips,Mangla2020ChartingTR,chen2019closerlook} typically extracts separate features for the query and gallery images, which prevents their models from emphasizing query-specific patterns in the gallery search.

QGN proposes three query-gallery interaction modules: \textbf{i.}\ the Query-guided Siamese Squeeze-and-Excitation Network (QSSE) re-weights both the query and gallery channel features, jointly conditioned on both images; \textbf{ii.}\ the Query Similarity Network (QSimNet) learns a similarity metric which is specific for comparing with the query; \textbf{iii.}\ the Query-guided RPN (QRPN) is used for detection, to provide query-specific proposals (besides the classic RPN).

The modularity of QGN allows to evaluate the core idea of extensively using query guidance in retrieval for detection and classification tasks. In both cases, query guidance enhances the relevance of ID features in the network backbone, matching function and, if present, in the region proposal. We consider person search as the detection task (in any case, this subsumes person re-identification) and few-shot fine-grained recognition as the classification task (to the best of our knowledge, there is no established few-shot fine-grained object detection benchmark yet).

Query-guidance is novel in the few-shot context. We evaluate QGN on five-widely adopted few-shot fine-grained datasets: CUB~\cite{WelinderEtal2010}, Stanford Cars~\cite{KrauseStarkDengFei-Fei_3DRR2013}, FGVC-Aircraft~\cite{aircraftdataset}, Stanford Dogs~\cite{Khosla2012NovelDF}, and Oxford Flowers~\cite{10.1109/CVPR.2006.42}. QGN achieves state-of-the-art results on CUB, FGVC-Aircraft and Stanford Dogs. Particularly on CUB, QGN surpasses the current best S2M2~\cite{Mangla2020ChartingTR} by a large margin, i.e.\ 12pp and 5pp in 1- and 5-shot learning experiments, respectively. Moreover, when employing a shallower ResNet18, the performance of QGN surpasses S2M2, which employs the deeper WRN~\cite{Mangla2020ChartingTR}, by 3.1pp for 1-shot learning.

For person search, we add our query-guided components on top of a recently improved OIM implementation~\cref{nae}, and achieve competitive performance with the state of the art on the large scale CUHK-SYSU~\cite{xiao2017joint} and PRW~\cite{8099840} datasets. We report comparison with several competing person search techniques, including the ones following our original work~\cite{munjal2019cvpr}. 
Both in person search and in few-shot fine-grained classification, we perform an in-depth analysis, including diverse backbones (ResNet10, ResNet18, ResNet50, WRN-28-10). Furthermore, we demonstrate the intuition of our proposed query-guided components via qualitative visualizations on both tasks.

\section{Related Work}\label{sec:related}
We review prior art on few-shot learning, fine-grained classification and person search, emphasizing methods which condition the feature extraction upon the query. To the best of our knowledge, QGN is the first technique addressing both tasks and it is the first query-guidance approach for few-shot fine-grained classification.

\noindent \textbf{Few-shot learning.}
Few-shot learning aims to train models that can rapidly adapt and generalize to new concepts using only a few samples. The copious recent progress in the field can be loosely divided into five categories. In the first, \textit{metric-based} methods~\cite{sung2018cvpr,ZHAO2022108880} learn a shared embedding space for the comparison of the feature embeddings from the query and the gallery images. The proposed QSimNet resembles the relation module in the Relation Network~\cite{sung2018cvpr}, but the input features of query and gallery are jointly extracted and end-to-end trained.
In the second category, \textit{optimization based} methods~\cite{DBLP:journals/corr/FinnAL17} adjust the optimization algorithm to learn from a few examples. Here the most popular is MAML~\cite{DBLP:journals/corr/FinnAL17}, which optimizes the initialization of the gradient-descent-based learner. \textit{Data hallucination} may be a third direction, based on the data augmentation and the scarce provided data.

More recently,~\cite{chen2019closerlook} proposed a simpler \textit{transfer learning} approach using a distance-based classifier, which is competitive with other more sophisticated approaches. S2M2~\cite{Mangla2020ChartingTR} extends their work with self-supervision techniques~\cite{Su2020When}. 
Following \cite{chen2019closerlook,Mangla2020ChartingTR}, QGN also employs the \textit{non-episodic} training, hence it does not need to train separately for different few-shot protocols. 
Unlike transfer learning methods, QGN jointly processes the query and the gallery with a Siamese network model and it does not need any fine-tuning at inference time. 

Finally, the category of \textit{dynamic network conditioning} methods uses the query or gallery examples to either tune or condition the network by \textit{attention} based mechanism~\cite{hou2019cross} or \textit{generate network parameters}~\cite{Zhao_2018_ECCV}.
Matching networks~\cite{matching2016} apply conditioning as post-processing with a bidirectional LSTM.~\cite{LIANG2022108662} uses a weight-centric learning strategy to push samples closer to their corresponding classifier weights. Other approaches generate weights by means of kernel generator or by combining basis convolutional kernel filters~\cite{Zhao_2018_ECCV}. These techniques relate to QSSE, which we employ for feature extraction, however our approach is the sole to make use of both the query and gallery features from the very first layers. Similar to ours, CAM~\cite{hou2019cross} generates query-gallery cross-attention maps, but it focuses on image parts, rather than entire feature channels, as we do. Also, the correlation layer of CAM is applied only once at the output layer, due to its high memory and runtime requirements, while our simpler QGN is applied at all network layers, which results in the query-gallery interaction across both coarser and finer details.

\noindent \textbf{{Few-shot fine-grained classification.}}
Fine-grained differs mainly from general few-shot learning as it focuses on categories with subtle distinctive traits, e.g.\ species of birds, dogs, flowers, car models. This is more complex and less researched. Within this literature, \cite{tang2020revisiting} targets fine-grained few-shot recognition by learning pose normalized embedding and uses extra part annotations.~\cite{ijcai2020-152} uses attention modules after the feature extractor to infer spatial and channel attentions.~\cite{TANG2022108792} employs a multi-scale feature pyramid and a multi-level attention pyramid to extract features of different granularities. More recently, \cite{chen2019closerlook} evaluates the generic few-shot methods including ProtoNet~\cite{snell2017nips}, MatchingNet~\cite{matching2016}, RelationNet~\cite{sung2018cvpr}  and MAML~\cite{DBLP:journals/corr/FinnAL17} on few-shot fine-grained classification. S2M2~\cite{Mangla2020ChartingTR} also evaluates its approach on the fine-grained case.~\cite{SunCZZZWW20} propose a unifying loss for various fine-grained tasks. Unlike the above methods, our QGN is a Siamese model and it leverages query-gallery cross-attention. 

\noindent \textbf{{Person Search.}}
There are several person search techniques but they are distinct from the previous, as no methods address both tasks. In person search, we distinguish sequential methods~\cite{Dong_2020_CVPR,Chuchu2019}, which cascade the person detection and person re-identification sub-tasks, from joint methods~\cite{munjal2019knowledge,patcog2022108654}, which perform both sub-tasks with a single network. The latter lag a bit behind sequential models in performance and are more complex to train, since detection and re-identification are conflicting sub-tasks. However these require in general less memory and computational resources, and are therefore preferable for industrial applications. QGN belongs to this second category, but the proposed query-guidance components are applicable to a sequential method, too.

Among the joint models, QGN relates to \cite{xiao2017joint} which introduces Online Instance Matching (OIM) into Faster RCNN~\cite{ren2015faster} as an additional multi-task loss. OIM is the de-facto standard re-identification loss, adopted by most recent person search approaches~\cite{Kim_2021_CVPR,Yan_2021_CVPR} as well as by QGN. PGA~\cite{Yan_2021_CVPR} uses the class prototype as a guidance for person attention. AlignPS~\cite{Kim_2021_CVPR} proposes an anchor free framework for person search with a feature aggregation module. Similarly,
BINet~\cite{dong2020} and NAE~\cite{Chen_2020_CVPR} build on top of OIM. BINet~\cite{dong2020} employs an additional parallel branch that takes cropped patches and supervises the joint model with interaction losses. NAE~\cite{Chen_2020_CVPR} decomposes the embeddings of OIM into angle and norm to accomplish re-ID and detection respectively.~\cite{LI2021107862} uses a hierarchical distillation strategy to transfer knowledge from a stronger teacher model to a student model. QGN is the first to introduce query-gallery interaction modules at different stages of the network, as well as throughout the backbone.

\noindent \textbf{{Query-guided person search.}} Prior work from ours~\cite{munjal2019cvpr} was the first to introduce query guidance for person search. Afterwards, this has been adopted by a few techniques, including TCTS~\cite{cheng_2020_CVPR} and IGPN~\cite{Dong_2020_CVPR}. TCTS proposes an identity-guided query detector to produce query-like person boxes for the subsequent re-ID network.
IGPN replaces the standard two-stage detector with a query- or instance-guided detector. IGPN adopts the Siamese RPN which correlates the query and gallery feature maps. By contrast, the proposed QRPN takes the query image crop at the input and re-weights the feature channels of the gallery image, emphasizing the traits of the person which we are searching for. Also, both IGPN and TCTS are sequential approaches that use two different models for detection and re-identification, while ours is a joint approach. Note that the joint models require less resources as compared to the sequential approaches as both the model parameters and processing are shared by the backbone. Additionally, learning joint models provides an appealing multi-task objective and addressing this successfully may result in a better use of data, higher performance and a better direction towards general intelligence, i.e. networks which understand multiple aspects of the scene.

\section{Method}

\label{sec:methodoverview}
In this section, we first formulate few-shot fine-grained classification and person search tasks. Then we discuss the proposed model and the three query-guided modules, as well as the optimization details.

\subsection{{Problem formulation}}
Let us describe the few-shot fine-grained classification and the person search tasks in a unified way. 

The training and test sets for both tasks can be given as $D_{train} = \{(x_i, y_i)\}_{i=1}^{N_{train}}$ with $C_{train}$ classes and $D_{test} = \{(x_i, y_i)\}_{i=1}^{N_{test}}$ with $C_{test}$ classes, respectively.
Here, $x_i$ represents the images and $y_i$ their corresponding ground-truth annotations. In particular, $y_i$ stands for the object classes in the case of few-shot fine-grained classification; and it means the person-ID and its location in the image for the task of person search.
The set of $C_{train}$ and $C_{test}$ classes in $D_{train}$ and $D_{test}$ are disjoint, i.e.\ at test time the model needs to classify new classes and person-IDs. 

Following literature from both tasks, we employ an episodic evaluation protocol, where a subset $D_{novel}$ is sampled from $D_{test}$ with $C_{novel}$ novel classes in each episode. A part of $D_{novel}$, i.e.\ $K$ examples from each of the $C_{novel}$ classes, is considered as query. The remaining part of $D_{novel}$ is the gallery, where the model needs to find the queries.

In the \textbf{few-shot} case, $(K+L)C_{novel}$ examples are sampled per episode as $D_{novel}$\,, i.e.\ $C_{novel}$ classes with $K$ examples per class as query. This is termed $C_{novel}$-way $K$-shot classification. While another $L$ examples per class are used as gallery.
$C_{novel}$ also represents the complexity of the evaluation. Larger $C_{novel}$ means more competition among classes during classification.
On the other hand, $K$ represents the number of examples per class in $C_{novel}$ that we can use as query. Larger $K$ means more information per class.
Typically, $K$ is either $1$ (1-shot learning) or $5$ (5-shot learning).

In \textbf{person search}, an episode $D_{novel}$ is sampled per query example. Here, $D_{novel}$ includes all positive examples corresponding to that query and a large number of random negatives from $D_{test}$, e.g. for CUHK-SYSU~\cite{xiao2017joint} the size of $D_{novel}$ is typically $101(= 100$ gallery $+1$ query$)$.
Therefore, $C_{novel}=2$ and $K=1$. 
$C_{novel} = 2$ means person search follows a binary classification strategy i.e. either the gallery sample matches the query or not. $K = 1$ means only one example per class is given as a query at one time. Therefore, person search can be viewed as a special case of few-shot classification, i.e. 1-shot learning.

\noindent \textbf{Note}: The {terminology} used in few-shot classification literature is different from that of person search. In person search, the query image is the one for which the class (or ID) is already known, while the gallery image needs to be classified.
Whereas, in few-shot classification, the query is the image that needs to be classified and the support is the image for which the class is already known. 
To keep the terminology consistent, we adopt the \textit{query-gallery} convention of person search for few-shot case as well.

\begin{figure*}[t!] 
\begin{center}
    \vspace{-0.3cm}
	\includegraphics[trim=0cm 0.5cm 0cm 2.1cm, clip=true, width=1.0\linewidth]{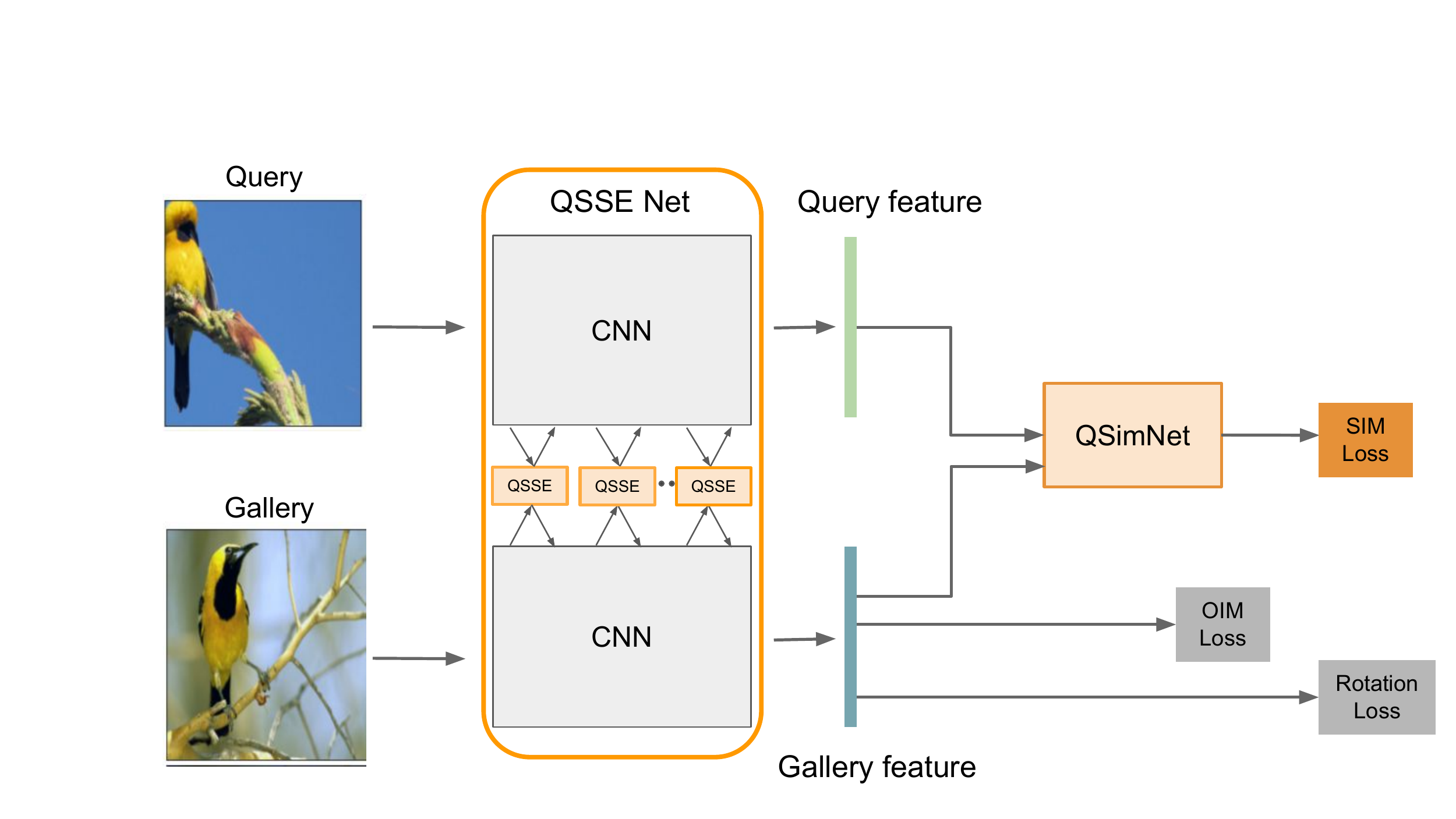}
	  
	\caption{Our proposed query-guided network for few-shot fine-grained classification. The \textit{bottom network} OIM~\cite{xiao2017joint} with auxiliary rotation loss is our baseline $\mathrm{OIM_{R}}$. We pair the baseline network with a siamese branch on top that takes a \textit{query} and guides the \textit{bottom network} at different levels. CNN here represents standard network architecture (ResNet10, ResNet18, WRN) followed by global average pooling. Note that we follow the person search terminology here: \textit{query} refers to the example for which we already know the class and \textit{gallery} needs to be classified. Our proposed query-guidance blocks are given in orange.
	 \vspace{-0.2cm}
	}\label{fig:fewshotnet}
    
\end{center}
\end{figure*}
\begin{figure*} 
\begin{center}
	\includegraphics[trim=0cm 7.5cm 0cm 0cm, clip=true, width=1\linewidth]{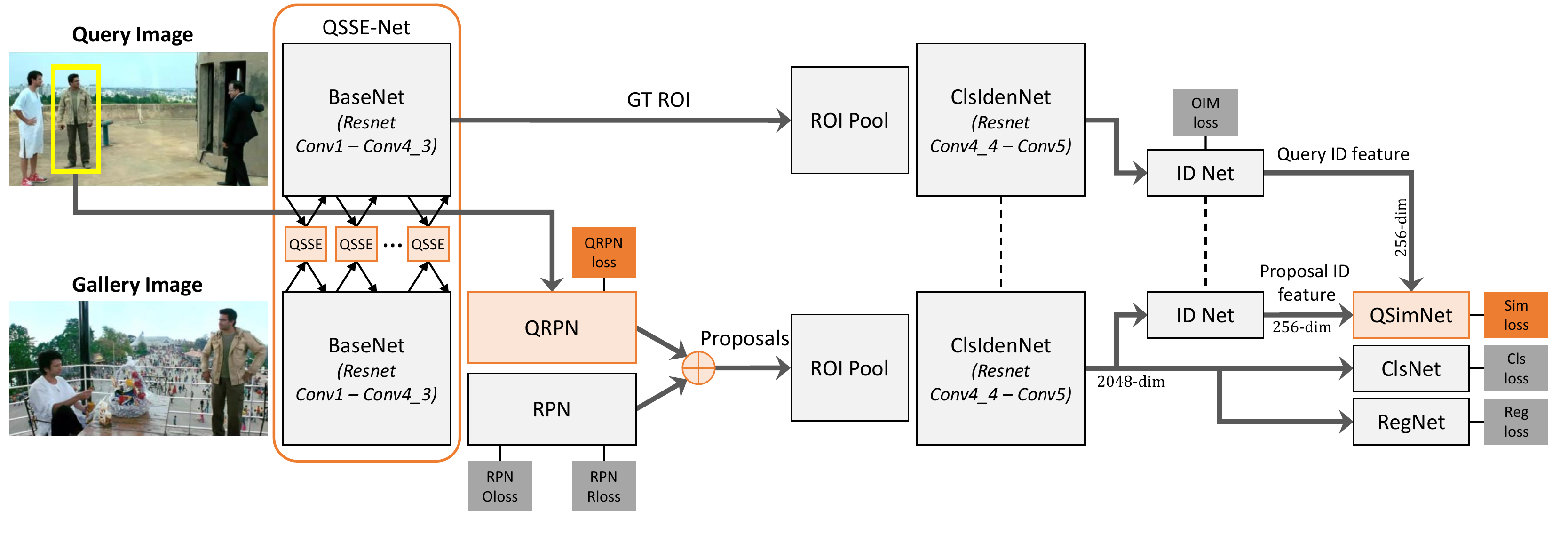}
    \vspace{-0.4cm}
	\caption{
	Our proposed query guided network architecture for person search. We pair the reference OIM~\cite{xiao2017joint} \textit{bottom network} with a novel Siamese \textit{top network}, to process the query and guide the person search at different levels of supervision (cf.\ Sec.~\ref{sec:methodoverview}).
	The novel query-guidance blocks of our approach, displayed in orange, are trained end-to-end with the whole network with specific loss functions (\textit{darker orange boxes}).
	}\label{fig:network}
	\vspace{-0.5cm}
\end{center}
\end{figure*}

\subsection{Query-guided Networks}

When provided with one or a few query samples, humans focus on its relevant and distinguishing features to find a corresponding gallery image and the object within it. Inspired by this, QGN proposes to process jointly the query and gallery images by a Siamese network design, and to model the query-gallery interactions by query-guided modules.

\noindent\textbf{Few-shot fine-grained classification} is accomplished by a Siamese network which processes the query and gallery images together, to produce an embedding for each of them, which is used to classify the gallery class to one of the novel classes in $D_{novel}$.
The relevant overall QGN model is illustrated in Figure~\ref{fig:fewshotnet}. The image embeddings are computed by two convolutional backbones. QGN contributes several Query-guided Siamese Squeeze-and-Excitation Network (QSSE) blocks, which relate the feature extraction at multiple layers of the backbone. Finally QGN realizes the classification of the embeddings by a Query Similarity Network (QSimNet), which learns the final metric similarity score. These components are described in detail in Sec.\ref{sec:qgn_components}. The implementation of each branch in the Siamese network draws details from~\cite{Mangla2020ChartingTR} and leverages for training the OIM loss~\cite{xiao2017joint}.

\noindent\textbf{Person search} is realized by two parallel Siamese detection networks, which extract the object crops from the query and gallery images, computes an embedding and compares those to assess whether they contain the same or different classes. The proposed QGN model is illustrated in Fig.~\ref{fig:network}. The image embeddings are extracted with convolutional backbones, leveraging the multi-layers query-gallery interaction by the QSSE. Then the object crops are extracted from the gallery by the proposed Query-guided Region Proposal Network (QRPN), i.e.\ proposals for bounding boxes tailored to the queried object, which integrates the proposals of a standard RPN~\cite{ren2015faster}. The top proposals are then passed to the subsequent network with a multi-task  head  for  classification (person vs non-person), localization refinement (regression offsets), and ID feature learning. Finally, the ID embeddings of query and each gallery proposal are compared by the QSimNet to distinguish same Vs different IDs. Details for the QGN components are provided in Sec.~\ref{sec:qgn_components}.

The implementation of each detection parallel branch follows details of~\cite{xiao2017joint}, including the OIM loss. Differently from the few-shot fine-grained, person search includes a detection task, so the entire query and gallery images are provided to the network, not just the person crops. Note that we do not need proposals for the query branch, since the query crop is given as input.

 \subsection{{Query-guided Network Components}}
\label{sec:qgn_components}

We propose three components to provide query-guidance at different stages of the Siamese networks. QSSE considers joint global context of the query and gallery to re-calibrate the channel features of the convolutional backbones. QRPN generates query-like proposals exploiting the query-crop specific patterns. QSimNet learns a distance metric to compare the query- gallery features.

In person search (Fig.~\ref{fig:network}), we adopt all three components. In few-shot fine grained classification (Fig.~\ref{fig:fewshotnet}), there is no need to generate candidate proposals and QGN consists only of QSSE and QSimNet. In both cases, all network parts are trained end-to-end.

\begin{figure}[t!] 
\begin{center}
	\includegraphics[scale=0.3]{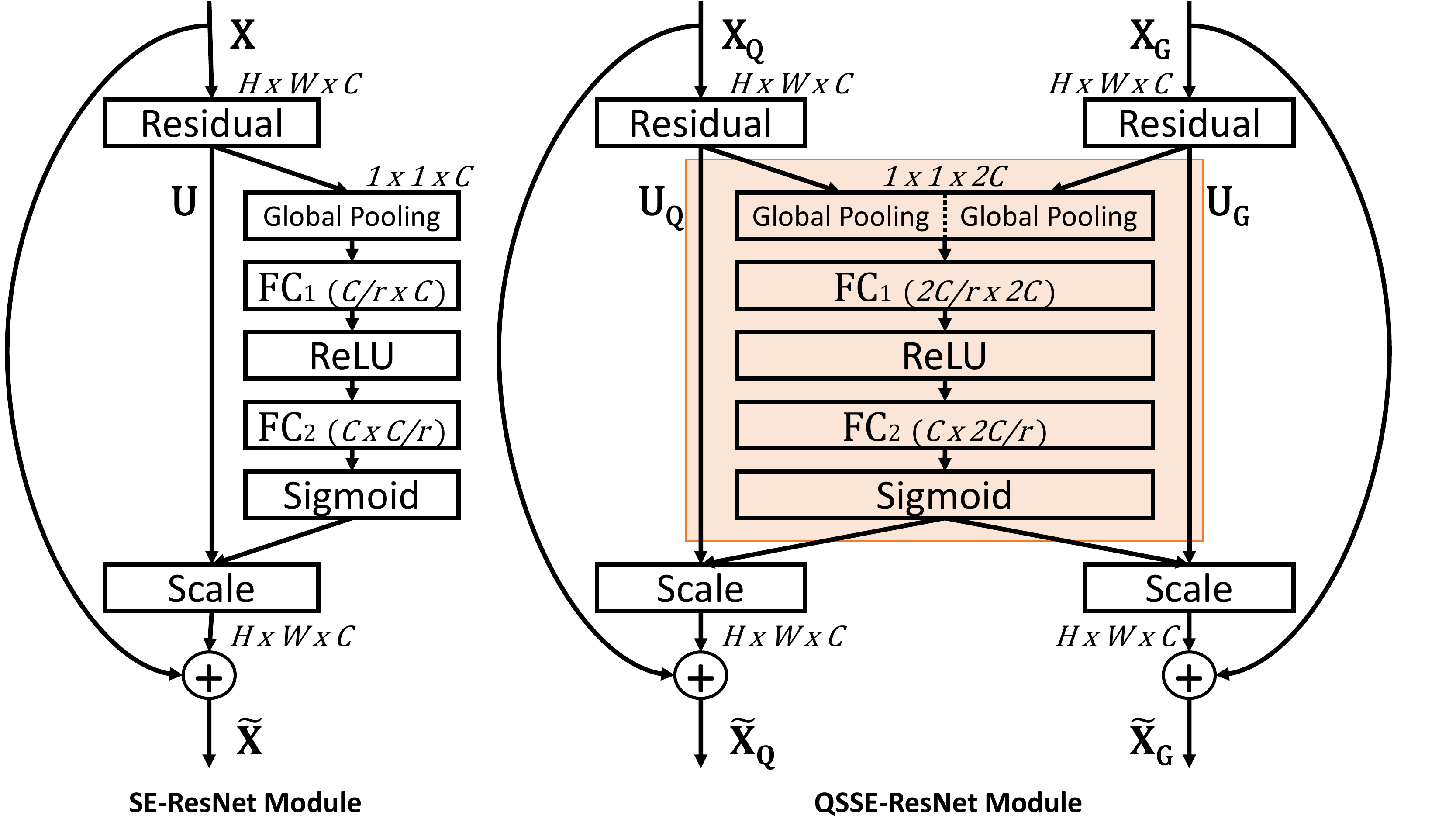}
%	\vspace{-0.5cm}
	\caption{On the left a standard SE block~\cite{Hu_2018_CVPR} is shown. On the right is our proposed Query-guided Siamese Squeeze-and-Excitation Network (QSSE-Net). 
	 The globally-pooled query and gallery features after the ResNet block are concatenated and jointly used to re-calibrate 
	 feature channels of both query and gallery. This way QSSE considers both intra- and inter-channel dependencies.   \vspace{-0.6cm}
	}\label{fig:qseblock}

\end{center}
\end{figure}

\subsubsection{{Query-guided Siamese Squeeze-and-Excitation Network (QSSE)}}\label{sec:qssenet}
The query and gallery objects in the images may be taken from different viewpoints and with different lighting conditions. Their embeddings should ideally disentangle these nuances.
To this goal, we propose the QSSE module, which leverages the interaction of query and gallery.
More specifically, as shown in Fig.~\ref{fig:network}, the QSSE modules, inserted at the output of each network block (e.g. residual block for ResNet), allow a joint re-calibration of the feature maps. 

The QSSE module draws inspiration from SE-Net~\cite{Hu_2018_CVPR}, extending it to pairs of images (Fig.~\ref{fig:qseblock}). In more detail, inside a QSSE, first a \emph{squeeze} operation is performed by global average pooling of query and gallery features. This operation summarizes the spatial information of each of the $C$ channels, giving descriptors $\mathrm{\bf z}_q$ and $\mathrm{\bf z}_g \in\mathbb{R}^{C}$ for query and gallery respectively. 

After this, an \emph{excitation} operation is performed where the two descriptors are first concatenated $[\mathrm{\bf z}_q,\mathrm{\bf z}_g]\in\mathbb{R}^{2C}$ and then passed through a non-linear bottleneck. The first layer $FC_1$ of the bottleneck is for dimensionality reduction, shrinking the dimension of the concatenated descriptor by a factor of $r$. 
This reduced feature ($\frac{2C}{r}$) is then passed through the ReLU operation ($\delta$) modeling non-linear dependencies between channels. Finally, the feature is expanded to $C$ dimensions by the next fully connected layer $FC_2$, followed by sigmoid activation ($\sigma$) to generate the weight vector $\mathrm{s} \in \mathbb{R}^C$. 
Mathematically, the Siamese squeeze-and-excitation operation is given by
\begin{equation}
\label{eq:qseblock}
\mathrm{\bf s} = \mathrm{F}_{ex}(\mathrm{\bf z}_q,\mathrm{\bf z}_g;\mathrm{\bf W}) = \sigma(~\mathrm{\bf W}_2 ~ \delta (~\mathrm{\bf W}_1 [\mathrm{\bf z}_q, \mathrm{\bf z}_g]~)~)
\end{equation} 

where the parameters of the first and second fully connected layers are, respectively, 
$\mathrm{\bf W}_1 \in \mathbb{R}^{\frac{2C}{r} \mathrm{x} 2C}$  and $\mathrm{\bf W}_2 \in \mathbb{R}^{C \mathrm{x} \frac{2C}{r} }$.

Following~\cite{Hu_2018_CVPR}, we set the reduction ratio $r$ to 16 in all our experiments. As shown in Fig.~\ref{fig:qseblock}, the \textit{scale} operation employs the weight vector $s$ to re-weight the residual outputs $\mathrm{\bf U_Q}$ (for query) and $\mathrm{\bf U_G}$ (for gallery), by channel-wise multiplication. These scaled outputs are then added to the original features $\mathrm{{\bf X}_Q}$ and $\mathrm{{\bf X}_G}$ via \textit{skip connections}, giving outputs $\mathrm{\widetilde{\bf X}_Q}$ and $\mathrm{\widetilde{\bf X}_G}$ respectively. Mathematically, the above operation is defined as
%Eq.~\ref{eq:scale}, 
\begin{equation}
\begin{split}
\label{eq:scale}
\mathrm{\widetilde{\bf X}_Q} = \mathrm{\bf X_Q} + \mathrm{\bf s} \odot \mathrm{\bf U_Q} \\
\mathrm{\widetilde{\bf X}_G} = \mathrm{\bf X_G} + \mathrm{\bf s} \odot \mathrm{\bf U_G}
\end{split}
\end{equation} 
where $\odot$ denotes the \textit{channel-wise} scaling operation.

\begin{figure}[t] 
\begin{center}
	\includegraphics[scale=0.3]{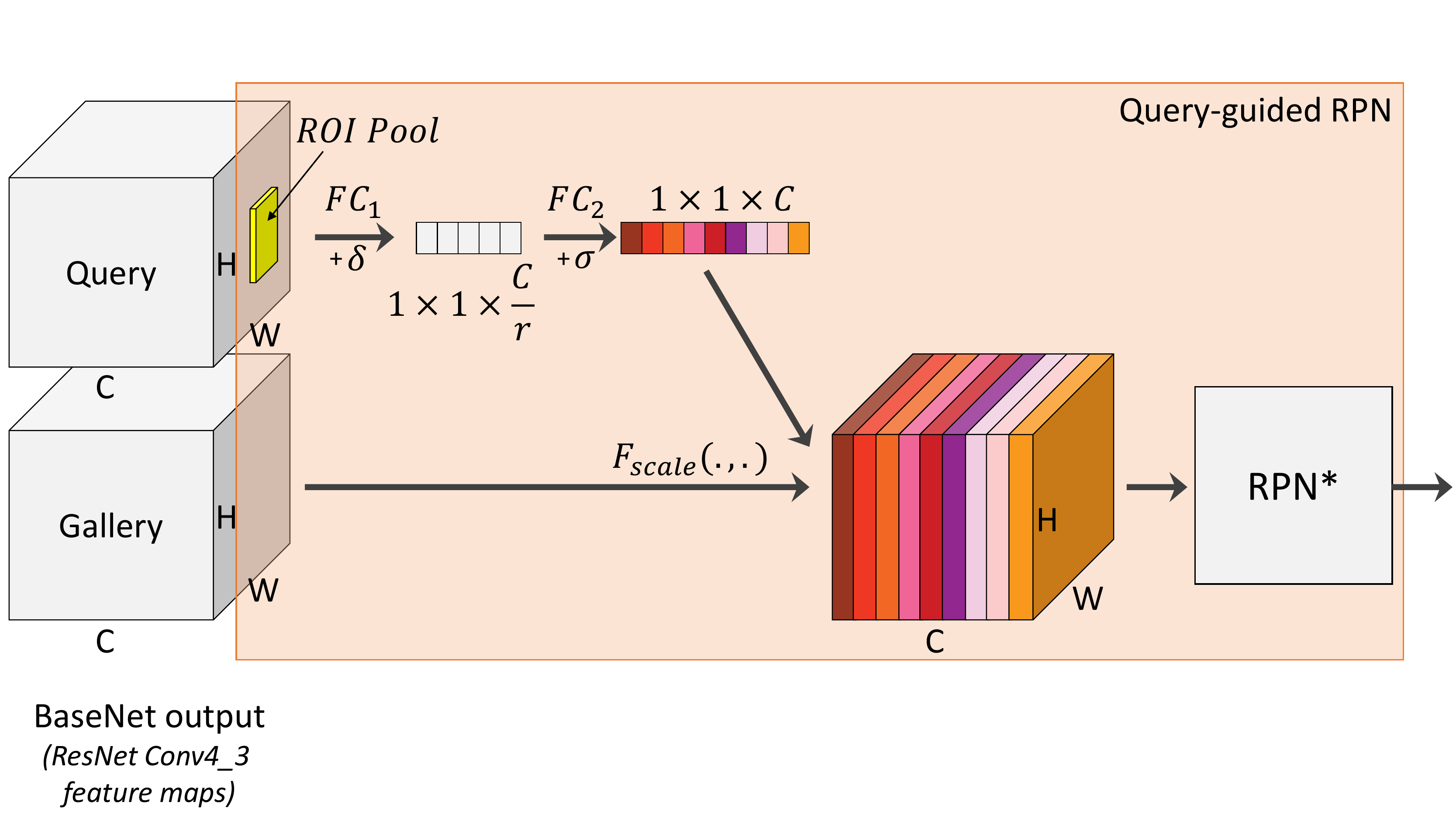}
   \vspace{-0.3cm}
	\caption{Query-guided Region Proposal Network (QRPN) adapts squeeze-and-excitation to generate weights from local query and re-calibrates gallery feature channels. The re-weighted gallery features are then passed to RPN$^*$ where RPN$^*$ is the standard RPN but does not compute regression offsets.   \vspace{-0.5cm}
	}\label{fig:qrpn}
   \vspace{-0.1cm}
\end{center}
\end{figure}

\subsubsection{{Query-guided RPN (QRPN)}}
\label{sec:qrpn}
QRPN is an attention-based region proposal network that leverages the local query features to generate query-like object proposals. QRPN consists of a channel-wise attention sub-network followed by a standard RPN~\cite{ren2015faster}, as shown in Fig.~\ref{fig:qrpn}. The attention network uses the cropped query features to re-weight the feature channels of the gallery image. The re-weighted features are then passed to a standard RPN to generate object proposals. 

In more detail, the query-crop features are first pooled using a ROI-pool~\cite{ren2015faster}. 
We then pass the pooled query features to a non-linear bottleneck. The first layer $FC_1$ of the bottleneck reduces the pooled features to $\mathbb{R}^{C/r}$, where $C=1024$ and $r=16$. 
Note that $FC_1$ is applied to all pixels of all the channels of the pooled map.
In this way, our attention mechanism leverages the spatially \emph{localized} query crop patterns to emphasize particular gallery channels. This also gives the network layer more freedom and lets the optimization dictate what specific local patterns to highlight, instead of just global features. This is in contrast with the \emph{squeeze} operation of SE-Net~\cite{Hu_2018_CVPR}.
The second fully connected layer $FC_2$ then expands the features back to $C$ dimensions, followed by a sigmoid ($\sigma$) activation to generate weights. Finally, the output weights are used to re-calibrate the gallery features and not the query itself. 

We further complement QRPN with the standard RPN in a parallel branch, that takes into account generic objectness score (cf. Fig \ref{fig:network}). This helps in retrieving further proposals when they are quite different from the query. The objectness score from RPN and query-similarity score from QRPN are summed up to generate final score for each anchor which is used for non-maximal suppression (NMS) at the stage of proposal generation. Note that both RPN included in QRPN and the parallel RPN follow the same design and use same anchors. 

QRPN is trained using \textbf{QRPN loss} which is a binary cross-entropy loss given as,
\begin{equation} \label{eq:qrpnloss}
L_{qrpn} = -\frac{1}{N}\sum_{N}\mathrm{log}(p_n^u)
\end{equation}  
where $p_n^u$ is the probability of the true class $u$ for the $n^{th}$ anchor out of a total of $N$ anchors.

\subsubsection{{Query-guided Similarity Net (QSimNet)}}\label{sec:qsimnet}

QSimNet is a deep query-dependent metric that is trained end-to-end with other network components. Unlike standard offline metrics such as the euclidean distance~\cite{xiao2017joint,Chen_2020_CVPR}, QSimNet alters the similarity measures for each query, to account for the relative importance of attributes such as e.g.\ color and shape.

\begin{figure}[t] 
\begin{center}
	\includegraphics[scale=0.26]{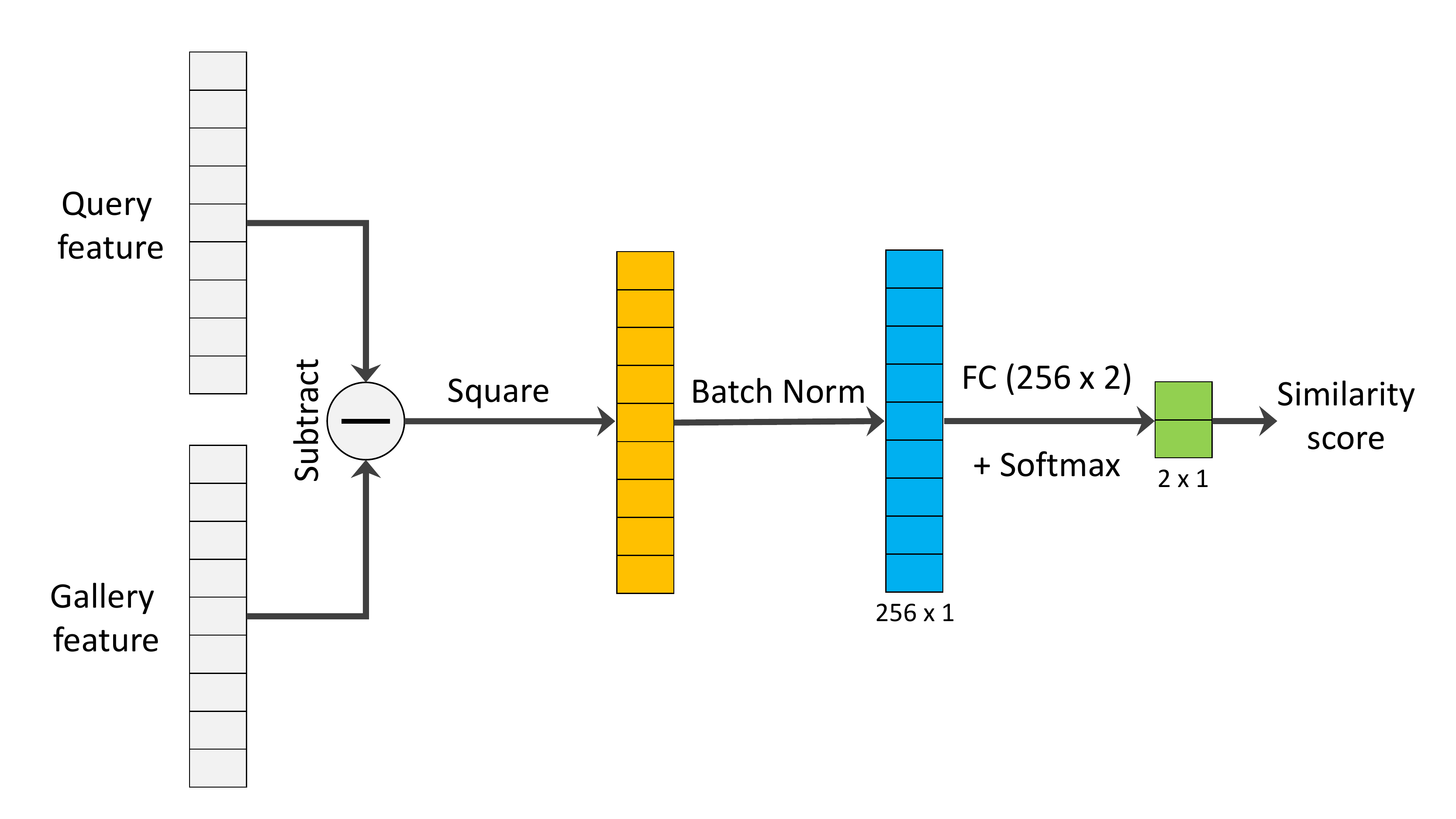}
	  \vspace{-0.5cm}
	\caption{Query-guided Similarity Network (QSimNet) estimates the similarity score between query and gallery features. For few-shot case, these features correspond to the output of CNN in upper and lower branches (Fig. \ref{fig:fewshotnet}), for person search, they correspond to the object features generated by ID Net (Fig.~\ref{fig:network}). QSimNet is trained end-to-end with other parts of the network.  \vspace{-0.5cm}
	}\label{fig:netsim}
   
\end{center}
\end{figure}

As shown in Fig.~\ref{fig:netsim}, QSimNet works by first calculating the L2 distance between the two features, i.e element-wise subtraction and square operation. This is followed by batch normalization and a fully connected layer with two outputs. Finally, a softmax is applied to generate similarity/dissimilarity scores.

QSimNet is trained using~\textbf{Sim loss} $L_{sim}$ which is defined as the binary
cross-entropy loss similar to $L_{qrpn}$. $L_{sim}$ is given as,
\begin{equation} \label{eq:simloss}
L_{sim} = -\frac{1}{N}\sum_{N}\mathrm{log}(p_n^t)
\end{equation}  

where N defines the number of pairs in the mini-batch and $p_n^t$ is the probability of the true class $t$ for the $n^{th}$ pair.

\subsection{Training Query-guided Networks}
\label{sec:modeltraining}

We discuss in details the optimization of QGN for each of the task.

\subsubsection{{Few-shot fine-grained classification}}
\label{sec:few_shot_method}

The QGN network is optimized in an end-to-end fashion, which considers both the classification backbone, as well as the QSSE and QSimNet.

\textit{Self-supervision} has been proven to improve few-shot learning in various recent works~\cite{Su2020When,Mangla2020ChartingTR} as it helps to overcome \textit{supervision-collapse}~\cite{Su2020When}, a phenomenon where training on the \textit{base} classes force the network to discard information irrelevant for the discrimination of \textit{base} classes, but crucial for the \textit{novel} classes. Various \textit{pretext} tasks have been proposed in literature for self-supervision. In this work, we opt rotation prediction~\cite{Su2020When} mainly because of its simplicity and effectiveness~\cite{Su2020When,Mangla2020ChartingTR}. 
In more details, each image in the batch is rotated by four angles (0\textdegree, 90\textdegree, 180\textdegree, 270\textdegree) and a 4-way rotation classifier is added on the top. 
The network is optimized with an additional rotation loss ($L_{rot}$), together with $L_{oim}$ and $L_{sim}$. The overall loss function $L_{fs}$ is therefore: 
\begin{equation}
\label{eq:fewshot}
L_{fs} = L_{oim} + L_{sim} + L_{rot} 
\end{equation}
Note that we do not follow an episode based training and use the same trained model, both for the 1- and 5-shot case. The inference architecture of the 1-shot case looks similar to the training phase (without the loss functions) as shown in Fig.~\ref{fig:fewshotnet}. We simply pass the query and gallery to the network to obtain their similarity score. 
However, in the 5-shot case, each of the 5 queries are passed to the CNN together with the gallery. 
This results in 5 different sets of feature vectors for each query and gallery. 
We compute the sum of these 5 features which are then normalised and passed to QSimNet to get the similarity score:
\begin{equation}
\begin{split}
\label{eq:qsimnetfewshot}
sim\_score = QSimNet(sum(f^{q_1},f^{q_2}...f^{q_N}), \\sum(f^{g_1},f^{g_2}...f^{g_N} ))
\end{split}
\end{equation}
where $f^{g_i}$ is the $ith$ gallery feature and $f^{q_i}$ is the corresponding  \textit{query} (support) feature, $i=1...N$.

\subsubsection{{Person search}}

The QGN end-to-end network training includes the detection network and the identification network, as well as QSSE, QRPN and QSimNet. The overall loss function $L_{ps}$ is: 
\begin{equation}
\label{eq:network}
\begin{split}
L_{ps} = L_{cls} &+ L_{reg} +  L_{rpn_o} + L_{rpn_r}\\
&+  L_{oim} +  L_{qrpn} +  L_{sim}
\end{split}
\end{equation}

where $L_{cls}$, $L_{reg}$, $L_{rpn_r}$ and $L_{rpn_o}$ are the standard Faster-RCNN losses~\cite{ren2015faster} for classification, regression, RPN regression and RPN objectness. The ID feature learning is supervised by standard OIM loss~\cite{xiao2017joint}, while our new components QRPN and QSimNet are supervised by $L_{qrpn}$ and $L_{sim}$ respectively. The losses are shown in Fig. \ref{fig:network} as dark gray or dark orange boxes.

During inference, it is typical for object detection pipelines to apply NMS at the end using final classification scores. However, we use the final similarity score from QSimNet for such NMS stage during inference. The classification score from ClsNet is only used to remove least confident detections with score less than 0.01.

\noindent \textbf{QRPN Anchor Sampling:} Since a typical gallery image can only contain one target-person matching the query crop, the number of positive anchors is significantly fewer as compared to the negatives. 
This leads to a skewed positive-to-negative ratio for training of the qrpn loss ($L_{qrpn}$). 
Therefore, we augment the target person in gallery via jittering i.e. the target box is moved randomly in the nearby region. Additionally, we keep a lower anchor-to-target IoU threshold of 0.6 for positive anchor sampling. To further reduce the number of negatives, we use a batch size of 128 instead of standard 256 hence improving the positive-to-negative ratio.
Note that the negative anchors are sampled from the background that do not cover other people in the gallery. This is because the non-target people in the gallery are positives for the standard RPN and it would lead to contrasting objectives for QRPN and RPN.

\section{Experimental evaluation}\label{sec:exp}

We experimentally evaluate QGN on recent datasets for few few-shot fine-grained classification and person search. On the few-shot fine-grained classification, QGN outperforms the current state of the art by a large margin. On the person search, QGN performs competitive with other approaches. In both cases, we provide novel qualitative visualizations of the query guidance.

\subsection{Experiments on few-shot fine-grained classification}

We evaluate QGN on the widely adopted Caltech-UCSD birds dataset (CUB)~\cite{WelinderEtal2010} and four other fine-grained datasets from different domains: Stanford Cars~\cite{KrauseStarkDengFei-Fei_3DRR2013}, FGVC-Aircraft~\cite{aircraftdataset}, Stanford Dogs~\cite{Khosla2012NovelDF}, and Oxford Flowers~\cite{10.1109/CVPR.2006.42}. Further to evaluating various backbones, we also provide a visualization of the QSSE.

\begin{table}[t]
\renewcommand{\arraystretch}{0.8}
\caption{Description of the five few-shot fine-grained datasets. Each row shows total number of images, total number of classes, followed by number of classes in train, val and test sets. 
\vspace{-0.4cm} 
}
\label{tab:few_shot_datasets}
\begin{center}
\resizebox{10cm}{!}{
\begin{tabular}{lcccccc}
\hline
\bf{Dataset} & \#\bf{images} & \#\bf{classes}& \#\bf{train} & \#\bf{val} & \#\bf{test}  \\
\hline
\hline
CUB (Birds) & 11,788& 200&100& 50 & 50\\
FGVC-Aircraft & 10,000& 100 & 50 &25& 25\\
Stanford Dogs & 20,580& 120& 60& 30 & 30\\
Oxford Flowers & 8,189& 102& 51& 25& 26\\
Stanford Cars & 16,185& 196 & 98& 49& 49\\

\hline

\end{tabular}
}

\end{center}
\vspace{-0.5cm}

\end{table}

\subsubsection{Benchmarks and Implementation details}
\label{sec:fewshotimpldetails}

The few-shot fine-grained datasets: {CUB}~\cite{WelinderEtal2010}, {Stanford Cars}~\cite{KrauseStarkDengFei-Fei_3DRR2013}, {FGVC-Aircraft}~\cite{aircraftdataset}, {Stanford Dogs}~\cite{Khosla2012NovelDF} and {Oxford Flowers}~\cite{10.1109/CVPR.2006.42}, are composed of 100-200 classes and a few thousands of images for each class. For CUB, we follow the split of~\cite{chen2019closerlook} as used by most previous approaches. For other four datasets, we follow the split of~\cite{Su2020When}. In Table~\ref{tab:few_shot_datasets}, we provide details of these datasets.

\noindent \textbf{Evaluation Criteria:} Following~\cite{Mangla2020ChartingTR}, we adopt an episodic few-shot evaluation and report the mean classification accuracy of $|D_{novel}|=600$ randomly generated 5-way 1-shot and 5-way 5-shot episodes with $L=15$ gallery per class.

\noindent \textbf{Implementation Details:}
We integrate the QSSE and QSimNet modules \cite{munjal2019cvpr} and the OIM loss~\cite{xiao2017joint} with the \textit{Rotation} self-supervision of \cite{Mangla2020ChartingTR}.
We experiment with three network architectures: ResNet10, ResNet18 and WRN-28-10 (width 28, scale factor 10). Following~\cite{chen2019closerlook,Mangla2020ChartingTR}, the image size is $224\times224$ for ResNet10/18 and $80\times80$ for WRN. The feature embedding is 512 for ResNet10/18 and it is 640 for WRN-28-10. In all experiments, the batch size is 8 (8 query-support pairs). The negative-to-positive ratio is 3 to 1, (3 query-support samples from the same class and 1 from different ones). We train for 120 epochs using the Adam optimizer with an initial learning rate of 0.001. During training, we augment the data via random crop, image jittering and random horizontal flip.

\begin{table}[t!]
\setlength{\tabcolsep}{4pt}
\renewcommand{\arraystretch}{0.9}
\caption{
Comparison on the few-shot fine-grained classification task on the \textbf{CUB} dataset using 5-way. Methods below the horizontal line  use either semi-supervised approach (additional unlabeled samples are used) or transductive inference (all unlabeled query samples are processed together). Our approach uses inductive inference where each query is processed independently. $\dagger$ denotes that the values are reported from the implementation in~\cite{chen2019closerlook}.
\vspace{-0.3cm}
}
\begin{center}
\resizebox{10cm}{!}{
\begin{tabular}{llccccc}
\hline
\bf{Setting} & \bf{Model} & \bf{Backbone} & \bf{1-shot} & \bf{5-shot}  & \bf{Publication}\\
\hline
\hline
& MatchingNet$^\dagger$~\cite{matching2016} & ResNet18 & 73.49   &   84.45 & NIPS16\\ 
& MAML$^\dagger$~\cite{DBLP:journals/corr/FinnAL17}   & ResNet18 & 68.42 &  83.47 & ICML17\\
& ProtoNet$^\dagger$~\cite{snell2017nips}  & ResNet18 & 72.99 & 86.64  & NIPS17\\
& RelationNet$^\dagger$~\cite{sung2018cvpr}  & ResNet18 & 68.58  & 84.05 & CVPR18\\
& Baseline++~\cite{chen2019closerlook}  & ResNet18 & 67.02   &   83.58 & ICLR19\\
In. & S2M2~\cite{Mangla2020ChartingTR}   & ResNet18 & 71.81 & 86.22& WACV20\\
& Proto+Jig~\cite{Su2020When}   & ResNet18 & - &  89.8& ECCV20\\

& Baseline++~\cite{Mangla2020ChartingTR} & WRN & 70.40 & 82.92& WACV20\\
& S2M2~\cite{Mangla2020ChartingTR}  & WRN & 80.68 & 90.85& WACV20\\
&\textbf{QGN (Ours)} & {ResNet10} &  {80.83}  &  {89.39}& Proposed\\
&\textbf{QGN (Ours)} & {ResNet18} &  83.82  &  91.22 & Proposed\\
&\textbf{QGN (Ours)} & {WRN} &  \textbf{84.15} &  \textbf{91.86} & Proposed\\

\hline
\hline
Tran./Semi &TEAM~\cite{trans2019} & ResNet18& 80.16& 87.17 & ICCV19\\
&ICI~\cite{Wang_2020_CVPR}  &
WRN & 91.11   &  92.98 & CVPR20\\ 
\hline

\end{tabular}
}
\end{center}
\vspace{-0.3cm}

\label{tab:cub_sota}
\end{table}

\begin{table}[t!]
\setlength{\tabcolsep}{4.5pt}

\renewcommand{\arraystretch}{0.8}
\caption{
Comparison on few-shot fine-grained classification on 5-way 5-shot. All models are built using ResNet18. $\dagger$ denotes the values are reported from the implementation in~\cite{Su2020When}.
\vspace{-0.5cm}
}

\label{tab:fine_grained_sota}

\begin{center}
\resizebox{10cm}{!}{
\begin{tabular}{lccccccc}
\hline
 \bf{Model} & \bf{CUB} & \bf{Cars} & \bf{Aircraft} & \bf{Dogs} & \bf{Flowers} & \bf{Publication}\\
\hline
\hline
 Softmax$^\dagger$  & 81.5  & 87.7& 89.2  &  77.6& 91.0 &   \\
 MAML$^\dagger$~\cite{DBLP:journals/corr/FinnAL17}  & 81.2 &   86.9 & 88.8 &  77.3 & 79.0&   ICML17\\
 ProtoNet$^\dagger$~\cite{snell2017nips} &  87.3 & 91.7 & 91.4& 83.0 & 89.2  & NIPS17\\
 Proto+Jig$^\dagger$~\cite{Su2020When}  & {89.8}& \textbf{92.4} & {91.8} & 85.7 & \textbf{92.2}  & ECCV20\\
\textbf{QGN (Ours)} & \textbf{91.2} & 91.3 & \textbf{92.0} & \textbf{85.9}  & 89.9 &  Proposed\\

\hline

\end{tabular}
}
\end{center}
\vspace{-0.2cm}

\end{table}

\subsubsection{Comparison to the state of the art}
In Table~\ref{tab:cub_sota}, we compare QGN to state-of-the-art few-shot fine-grained classification methods on the CUB dataset. QGN with the ResNet18 backbone achieves an accuracy of 83.82 and 91.22 for the 1-shot and 5-shot cases respectively, surpassing the previous best technique S2M2~\cite{Mangla2020ChartingTR} by the large margins of 12pp and 5pp.
These results also surpass the performance of S2M2 with the larger WRN backbone, by 3.1pp and 0.4pp respectively.
Similarly, QGN with the shallower ResNet10 backbone also surpasses S2M2 with the ResNet18 backbone by 9pp and 3.2pp. For completeness, we report in Table~\ref{tab:cub_sota} all most recent techniques. Methods below the double line either use additional unlabeled data (semi-supervised) or evaluate all queries together (transductive), hence they do not make a fair comparison to our approach. However, these techniques appear complementary to the proposed query guidance and they could be integrated into QGN in future work. 

In Table~\ref{tab:fine_grained_sota}, we compare QGN to other approaches on four other few-shot fine-grained datasets in addition to birds (CUB). As shown in the table, for 3 out of 5 datasets i.e birds, aircrafts and dogs, we outperform the previous best results by 1.4pp, 0.2pp and 0.2pp respectively.

\begin{table}[t!]
\setlength{\tabcolsep}{3.7pt}
\renewcommand{\arraystretch}{0.8}
\begin{center}
\caption{
Importance of each proposed model component, as evaluated on the \textbf{CUB} few-shot fine-grained classification dataset. The accuracy is reported as mean of 600 randomly generated episodes is reported.
\vspace{-0.1cm}
}
\label{tab:cub_ablation}

\resizebox{10cm}{!}{
\begin{tabular}{lcccccc}
\hline

&\multicolumn{2}{c}{\textbf{ResNet10}} &\multicolumn{2}{c}{\textbf{ResNet18}}  & \multicolumn{2}{c}{\textbf{WRN-28-10}} \\
\cmidrule(l){2-3} \cmidrule(l){4-5} \cmidrule(l){6-7}
 Method  & 1-shot & 5-shot & 1-shot & 5-shot & 1-shot & 5-shot \\
 \hline
 \hline

\textit{Rotation}~\cite{Mangla2020ChartingTR} & - & -& 72.40 & 84.83 &  77.61 & 89.32 \\ 
 $\mathrm{OIM_{R}}$ (\textit{Baseline})  & 77.76 & 87.88 & 80.27 & 89.81 &   81.45 & 90.15 \\ 
 + \textit{QSSE} & 78.79 & 88.92 & 80.72 & \bf{91.30} & 83.99& 91.42 \\
 + \textit{QSimNet} & 80.12 & 89.04 &82.20 & 90.89 & 83.05 & 91.81 \\
 + \textit{QSSE} + \textit{QSimNet} (=QGN) & \textbf{80.83} &  \textbf{89.39} & \textbf{83.82} & {91.22} &  \textbf{84.15} & \textbf{91.86} \\
 \hline

\end{tabular}
}
\end{center}
\vspace{-0.5cm}
\end{table}

\subsubsection{Ablation Studies}
\noindent\textbf{QGN components.} We evaluate the effectiveness of query-guided components applicable to few-shot classification, QSSE and QSimNet, with ablation studies.

\noindent \textbf{CUB}. In Table~\ref{tab:cub_ablation}, we consider analysis of QGN with backbones ResNet10, ResNet18 and WRN-28-10. The reference baseline combines the OIM classifier with an auxiliary rotation prediction for self-supervision. We dub this model $\mathrm{OIM_{R}}$. This coincides with \cite{Mangla2020ChartingTR}, which we indicate as \textit{Rotation}, except for replacing the cosine classifier with OIM. For ResNet18,
$\mathrm{OIM_{R}}$ achieves 80.27 and 89.81 for 1- and 5-shot classifications, outperforming \textit{Rotation}, which only achieves 72.40 and 84.83.
Since OIM is the leading technique for person search, but it had not been adopted for few-shot classification, this result motivates the QGN proposition for a unified approach to both tasks.

Next, we add our proposed QSSE on top of this baseline. For ResNet10, the addition of QSSE brings an improvement of almost 1pp for both 1-shot and 5-shot. For ResNet18, it brings an improvement of 0.5pp for the 1-shot and of 1.5pp for the 5-shot case. Then we add QSimNet on top of $\mathrm{OIM_{R}}$. For ResNet10, it improves by almost 2.4pp and 1.2pp for the 1-shot and 5-shot  respectively. For ResNet18, it improves by almost 2pp and 1pp. QGN for few-shot fine grained classification is given by combining QSSE and QSimNet. For ResNet10, QGN achieves an accuracy of 80.83 and 89.39, for the 1-shot and 5-shot case respectively. For ResNet18, QGN achieves an accuracy of 83.82 and 91.22. A similar improvement can be seen for the deeper WRN. Overall, in most cases, the best performance is consistently achieved by combining the two components, showing that QSSE and QSimNet are complementary.

\noindent \textbf{QSSE Analysis.} In Table \ref{tab:params_speed}, we
compare the parameter and computational speed of $OIM_R$ and $OIM_R + QSSE$. The comparison shows that the inclusion of QSSE adds only marginal additional parameters $\sim 2$\%, however runtime complexity has increased by $\sim50$\%. This is due to the siamese design of QSSE architecture that processes pair of images together.

\begin{table*}[t]
\renewcommand{\tabcolsep}{10.5pt}
\renewcommand{\arraystretch}{0.8}
\begin{center}
\caption{Comparison of the number of parameters and runtime complexity between $\mathrm{OIM_{R}}$ and $\mathrm{OIM_{R}}$ + \textit{QSSE}. The TFLOPS have been measured on a Tesla K80 GPU.
\vspace{-0.1cm}
}
\label{tab:params_speed}

\resizebox{8cm}{!}{
\begin{tabular}{lcc}
\hline
& {Params (M)}   & {Runtime Complexity (TFLOPS)}  \\
%&  \CHANGES{(M)}  & \CHANGES{(TFLOPS)} \\
\hline
\hline
{$\mathrm{OIM_{R}}$} & {11.28}  & {229.91} \\
{$\mathrm{OIM_{R}}$ + QSSE} & {11.54} & {344.65} \\

\hline
\end{tabular}
}
\end{center}
\vspace{-0.5cm}

\end{table*}

%%%%%%%%%%%%%%%%%%%%%%%%%%%%%%%%%
% \bgroup
%\tabcolsep 1pt
%\renewcommand{\arraystretch}{0.5}

\begin{figure*}[t!]
\renewcommand{\arraystretch}{0.5}
\renewcommand{\tabcolsep}{1pt}
\begin{center}
%%%%%%%%%%%%%%%%%%%%%%%%%%%%%%%%%%%%%%%%%%%%%%%%%%%%%%%%%
\begin{tabular}{c|ccccc}

%&&&&&&\\

\includegraphics[scale=0.17, trim=0cm 0cm 0cm 0cm, clip=true]{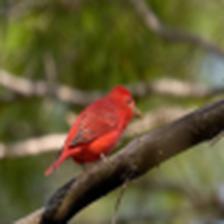}
&
\includegraphics[scale=0.17, trim=0cm 0cm 0cm 0cm, clip=true]{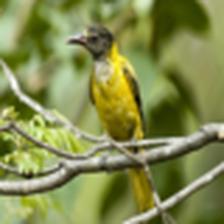} 
&

\includegraphics[scale=0.17,trim=0cm 0cm 0cm 0cm, clip=true]{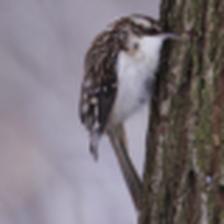}
&
\includegraphics[scale=0.17,trim=0cm 0cm 0cm 0cm, clip=true]{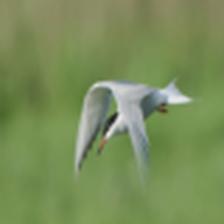}
&

\includegraphics[scale=0.17,trim=0cm 0cm 0cm 0cm, clip=true, cfbox=green 2pt 2pt]{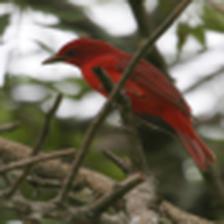}
&
\includegraphics[scale=0.17,trim=0cm 0cm 0cm 0cm, clip=true]{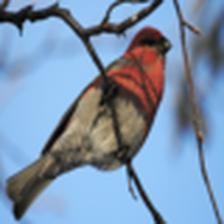}

\\

\includegraphics[scale=0.17, trim=0cm 0cm 0cm 0cm, clip=true]{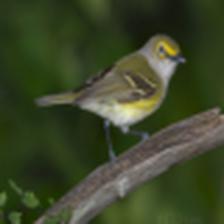}
&
\includegraphics[scale=0.17, trim=0cm 0cm 0cm 0cm, clip=true]{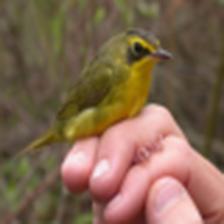} 
&
\includegraphics[scale=0.17,trim=0cm 0cm 0cm 0cm, clip=true, cfbox=green 2pt 2pt ]{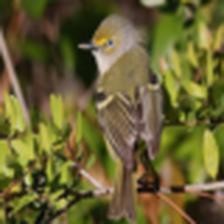}
&
\includegraphics[scale=0.17,trim=0cm 0cm 0cm 0cm, clip=true]{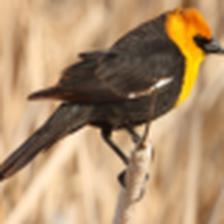}
&
\includegraphics[scale=0.17,trim=0cm 0cm 0cm 0cm, clip=true]{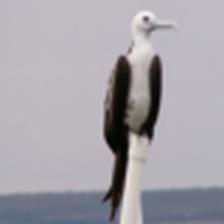}
&
\includegraphics[scale=0.17,trim=0cm 0cm 0cm 0cm, clip=true]{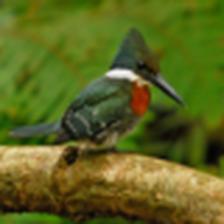}
\\

% \includegraphics[scale=0.173, trim=0cm 0cm 0cm 0cm, clip=true]{fig/cub_results/Episode_11_query_4_name_n0449338100000170.jpg}

% &

% \includegraphics[scale=0.173, trim=0cm 0cm 0cm 0cm, clip=true]{fig/cub_results/Episode_11_support_0_0_name_n0216715100000102.jpg} 
% &

% \includegraphics[scale=0.173,trim=0cm 0cm 0cm 0cm, clip=true]{fig/cub_results/Episode_11_support_1_0_name_n0756508300000628.jpg}
% &
% \includegraphics[scale=0.173,trim=0cm 0cm 0cm 0cm, clip=true]{fig/cub_results/Episode_11_support_2_0_name_n0210638200000003.jpg}
% &

% \includegraphics[scale=0.38,trim=0cm 0cm 0cm 0cm, clip=true, cfbox=red 2pt 2pt]{fig/cub_results/Episode_11_support_3_0_name_n0280844000000912_red.jpg}
% &
% \includegraphics[scale=0.173,trim=0cm 0cm 0cm 0cm, clip=true]{fig/cub_results/Episode_11_support_4_0_name_n0449338100000533.jpg}
% \\

\includegraphics[scale=0.17, trim=0cm 0cm 0cm 0cm, clip=true]{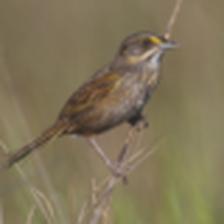} 
&
\includegraphics[scale=0.17,trim=0cm 0cm 0cm 0cm, clip=true]{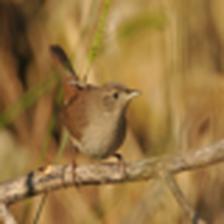}
&
\includegraphics[scale=0.17,trim=0cm 0cm 0cm 0cm, clip=true]{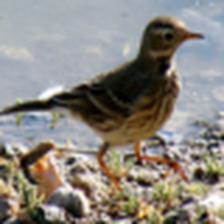}
&
\includegraphics[scale=0.17,trim=0cm 0cm 0cm 0cm, clip=true]{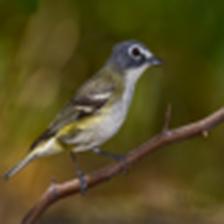}
&
\includegraphics[scale=0.17, trim=0cm 0cm 0cm 0cm, clip=true]{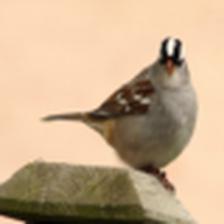}
&
\includegraphics[scale=0.17,trim=0cm 0cm 0cm 0cm, clip=true, cfbox=green 2pt 2pt]{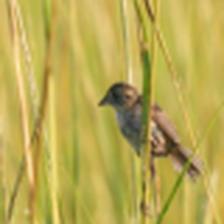}

\\

\includegraphics[scale=0.17, trim=0cm 0cm 0cm 0cm, clip=true]{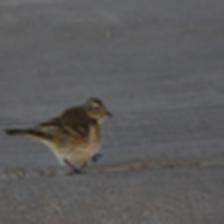} 
&
\includegraphics[scale=0.17,trim=0cm 0cm 0cm 0cm, clip=true, cfbox=green 2pt 2pt]{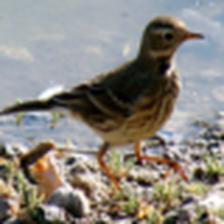}
&
\includegraphics[scale=0.17,trim=0cm 0cm 0cm 0cm, clip=true]{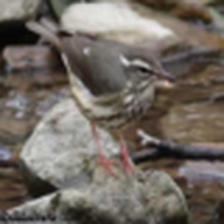}
&
\includegraphics[scale=0.17,trim=0cm 0cm 0cm 0cm, clip=true]{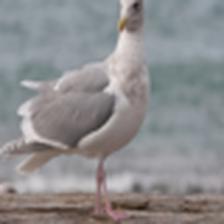}
&
\includegraphics[scale=0.17,trim=0cm 0cm 0cm 0cm, clip=true]{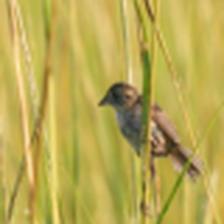}
&
\includegraphics[scale=0.17, trim=0cm 0cm 0cm 0cm, clip=true]{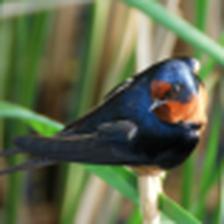}

\\

% \includegraphics[scale=0.173, trim=0cm 0cm 0cm 0cm, clip=true]{fig/cub_results/episode_ 10_class_ 0_0.jpg} 
% &
% \includegraphics[scale=0.173,trim=0cm 0cm 0cm 0cm, clip=true]{fig/cub_results/episode_ 10_query_ 0_0.jpg}
% &
% \includegraphics[scale=0.173,trim=0cm 0cm 0cm 0cm, clip=true, cfbox=red 2pt 2pt]{fig/cub_results/episode_ 10_query_ 0_1.jpg}
% &

% \includegraphics[scale=0.173,trim=0cm 0cm 0cm 0cm, clip=true]{fig/cub_results/episode_ 10_query_ 0_2.jpg}
% &
% \includegraphics[scale=0.173,trim=0cm 0cm 0cm 0cm, clip=true]{fig/cub_results/episode_ 10_query_ 0_3.jpg}
% &
% \includegraphics[scale=0.173, trim=0cm 0cm 0cm 0cm, clip=true]{fig/cub_results/episode_ 10_query_ 0_4.jpg}

% \\

\includegraphics[scale=0.17, trim=0cm 0cm 0cm 0cm, clip=true]{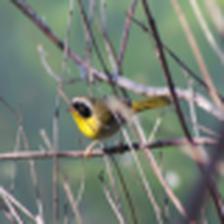}
&

\includegraphics[scale=0.17,trim=0cm 0cm 0cm 0cm, clip=true]{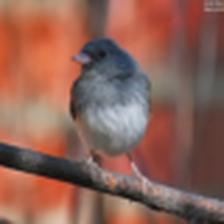}
&

\includegraphics[scale=0.17, trim=0cm 0cm 0cm 0cm, clip=true]{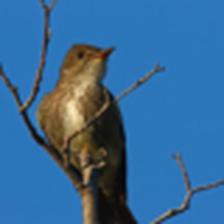} 
&
\includegraphics[scale=0.17,trim=0cm 0cm 0cm 0cm, clip=true, cfbox=red 2pt 2pt]{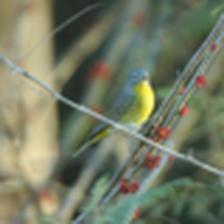}
&

\includegraphics[scale=0.17,trim=0cm 0cm 0cm 0cm, clip=true]{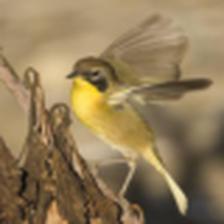}
&

\includegraphics[scale=0.17,trim=0cm 0cm 0cm 0cm, clip=true]{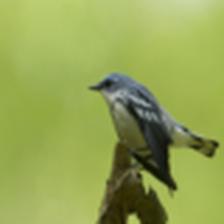}

\\
\includegraphics[scale=0.17, trim=0cm 0cm 0cm 0cm, clip=true]{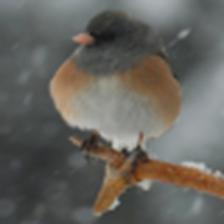}

&

\includegraphics[scale=0.17, trim=0cm 0cm 0cm 0cm, clip=true]{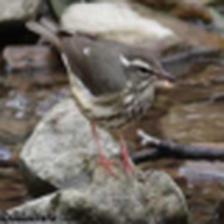} 
&

\includegraphics[scale=0.17,trim=0cm 0cm 0cm 0cm, clip=true, cfbox=red 2pt 2pt]{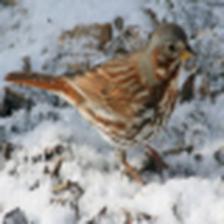}
&
\includegraphics[scale=0.17,trim=0cm 0cm 0cm 0cm, clip=true]{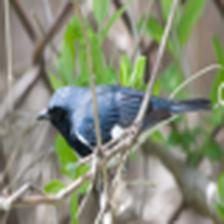}
&

\includegraphics[scale=0.17,trim=0cm 0cm 0cm 0cm, clip=true]{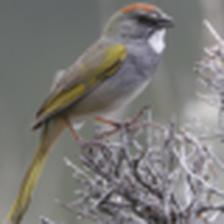}
&
\includegraphics[scale=0.17,trim=0cm 0cm 0cm 0cm, clip=true]{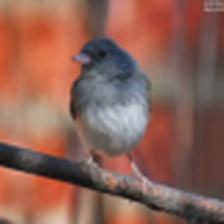}
\\

\footnotesize Gallery &  \multicolumn{5}{c}{\footnotesize {1 (-shot) query example from each of 5 (-way) classes}} \\

\end{tabular}

\end{center}
\vspace{-0.5cm}
\caption{Qualitative results on \textbf{CUB} for 5-way 1-shot classification using our proposed QGN. The first column shows the gallery image to be classified. The next five columns show 1 (-shot) query example from each of the 5 (-way) classes. For each gallery image, the query example with highest similarity score is marked. The correctly assigned class is marked with a green bounding box, while a red bounding box depicts wrong classification.}
\vspace{-0.4cm}
%Note that we follow the \textit{query-gallery} convention of person search.

\label{fig:cub-qualitative}
\end{figure*}

%%%%%%%%%%%%%%%%%%%%%%%%%%%%%%%%%
% \bgroup
%\renewcommand\tabcolsep{1pt}
%\tabcolsep 1.0pt
%\renewcommand{\arraystretch}{0.5}
% \setlength\arrayrulewidth{1.5pt}

\begin{figure*}[h]
\renewcommand\tabcolsep{1pt}
\renewcommand{\arraystretch}{0.5}
\begin{center}
%%%%%%%%%%%%%%%%%%%%%%%%%%%%%%%%%%%%%%%%%%%%%%%%%%%%%%%%%
\begin{tabular}{cc|cc|cc||cc|cc|cc}

%&\multicolumn{5}{c}{\textbf{Positive Pairs}} &\multicolumn{5}{c}{\textbf{Negative Pairs}}\\

\includegraphics[scale=0.12, trim=0cm 0cm 0cm 0cm, clip=true]{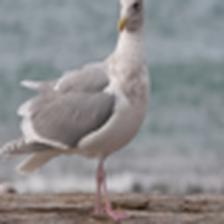}
&
\includegraphics[scale=0.12, trim=0cm 0cm 0cm 0cm, clip=true]{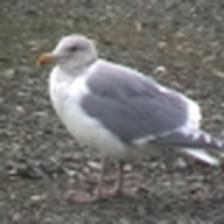}
&

\includegraphics[scale=0.12, trim=0cm 0cm 0cm 0cm, clip=true]{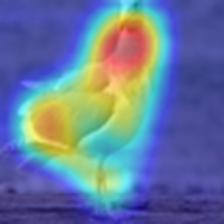}
&
\includegraphics[scale=0.12, trim=0cm 0cm 0cm 0cm, clip=true]{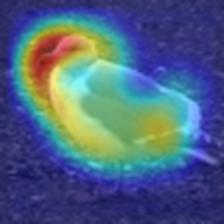} 
&
\includegraphics[scale=0.12,trim=0cm 0cm 0cm 0cm, clip=true]{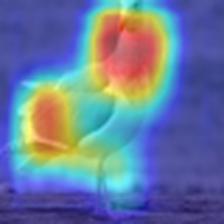}
&
\includegraphics[scale=0.12,trim=0cm 0cm 0cm 0cm, clip=true]{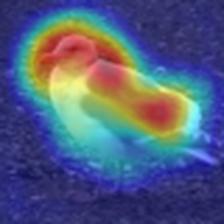} &
\includegraphics[scale=0.12, trim=0cm 0cm 0cm 0cm, clip=true]{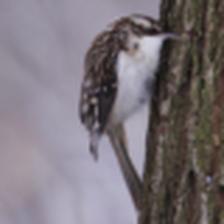}
&
\includegraphics[scale=0.12, trim=0cm 0cm 0cm 0cm, clip=true]{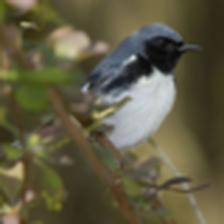} 
&

\includegraphics[scale=0.12, trim=0cm 0cm 0cm 0cm, clip=true]{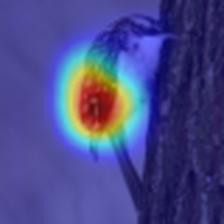}
&
\includegraphics[scale=0.12, trim=0cm 0cm 0cm 0cm, clip=true]{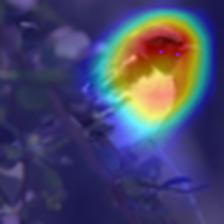} 
&
\includegraphics[scale=0.12,trim=0cm 0cm 0cm 0cm, clip=true]{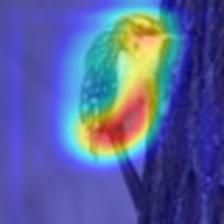}
&
\includegraphics[scale=0.12,trim=0cm 0cm 0cm 0cm, clip=true]{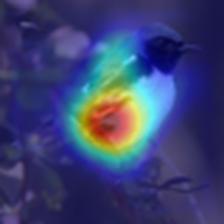}

\\
\includegraphics[scale=0.12, trim=0cm 0cm 0cm 0cm, clip=true]{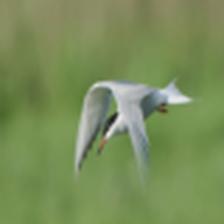}
&
\includegraphics[scale=0.12, trim=0cm 0cm 0cm 0cm, clip=true]{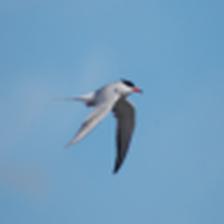}
&

\includegraphics[scale=0.12, trim=0cm 0cm 0cm 0cm, clip=true]{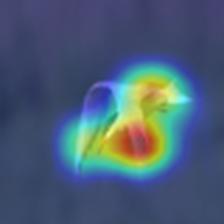}
&
\includegraphics[scale=0.12, trim=0cm 0cm 0cm 0cm, clip=true]{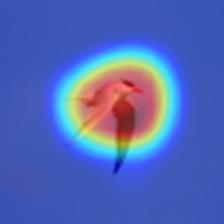} 
&
\includegraphics[scale=0.12,trim=0cm 0cm 0cm 0cm, clip=true]{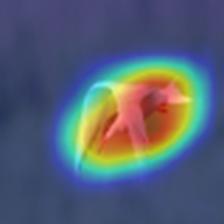}
&
\includegraphics[scale=0.12,trim=0cm 0cm 0cm 0cm, clip=true]{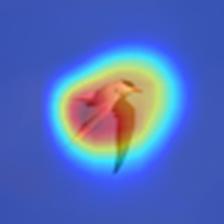} &

\includegraphics[scale=0.12, trim=0cm 0cm 0cm 0cm, clip=true]{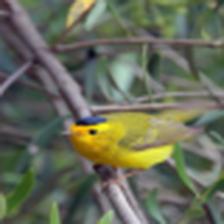}
&
\includegraphics[scale=0.12, trim=0cm 0cm 0cm 0cm, clip=true]{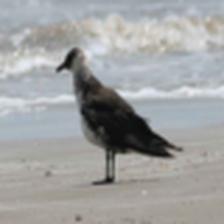} 
&

\includegraphics[scale=0.12, trim=0cm 0cm 0cm 0cm, clip=true]{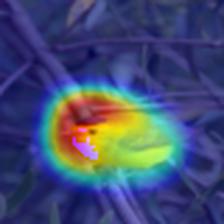}
&
\includegraphics[scale=0.12, trim=0cm 0cm 0cm 0cm, clip=true]{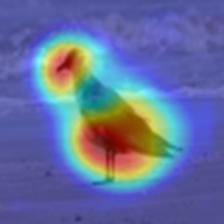} 
&
\includegraphics[scale=0.12,trim=0cm 0cm 0cm 0cm, clip=true]{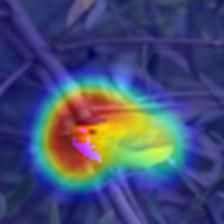}
&
\includegraphics[scale=0.12,trim=0cm 0cm 0cm 0cm, clip=true]{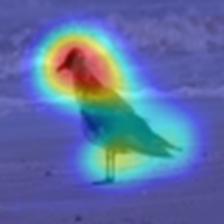} 

\\

% \includegraphics[scale=0.3, trim=0cm 0cm 0cm 0cm, clip=true]{fig/oimVSqsse/oim/grad_cam__episode_7__supp_20_flag_ 3_.jpg}
% &
% \includegraphics[scale=0.3, trim=0cm 0cm 0cm 0cm, clip=true]{fig/oimVSqsse/oim/grad_cam__episode_7__query_20_flag_ 3_.jpg} 
% &
% \includegraphics[scale=0.3,trim=0cm 0cm 0cm 0cm, clip=true]{fig/oimVSqsse/qsse/grad_cam__episode_7__supp_20_flag_ 3_.jpg}
% &
% \includegraphics[scale=0.3,trim=0cm 0cm 0cm 0cm, clip=true]{fig/oimVSqsse/qsse/grad_cam__episode_7__query_20_flag_ 3_.jpg}

% \\

\includegraphics[scale=0.12, trim=0cm 0cm 0cm 0cm, clip=true]{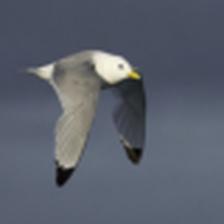}
&
\includegraphics[scale=0.12, trim=0cm 0cm 0cm 0cm, clip=true]{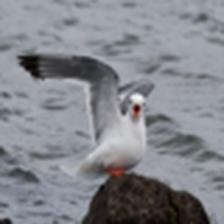}
&

\includegraphics[scale=0.12, trim=0cm 0cm 0cm 0cm, clip=true]{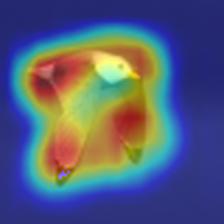}
&
\includegraphics[scale=0.12, trim=0cm 0cm 0cm 0cm, clip=true]{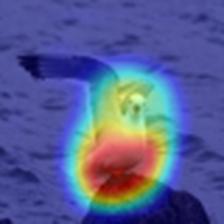} 
&
\includegraphics[scale=0.12,trim=0cm 0cm 0cm 0cm, clip=true]{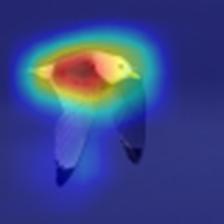}
&
\includegraphics[scale=0.12,trim=0cm 0cm 0cm 0cm, clip=true]{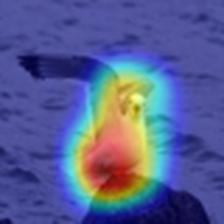} &

\includegraphics[scale=0.12, trim=0cm 0cm 0cm 0cm, clip=true]{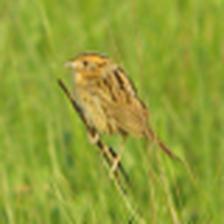}
&
\includegraphics[scale=0.12, trim=0cm 0cm 0cm 0cm, clip=true]{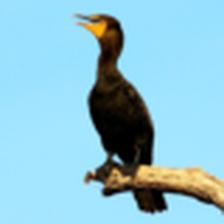} 
&

\includegraphics[scale=0.12, trim=0cm 0cm 0cm 0cm, clip=true]{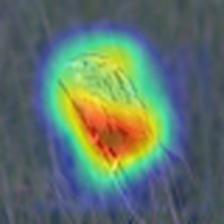}
&
\includegraphics[scale=0.12, trim=0cm 0cm 0cm 0cm, clip=true]{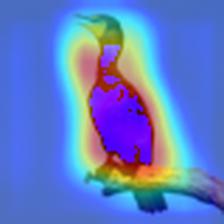} 
&
\includegraphics[scale=0.12,trim=0cm 0cm 0cm 0cm, clip=true]{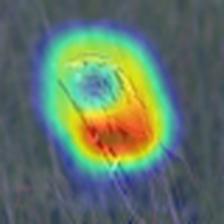}
&
\includegraphics[scale=0.12,trim=0cm 0cm 0cm 0cm, clip=true]{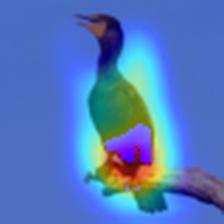}

\\
\tiny{Q} & \tiny{G} & \tiny{Q} & \tiny{G} & \tiny{Q} & \tiny{G} & \tiny{Q} & \tiny{G}  & \tiny{Q} & \tiny{G}  & \tiny{Q} & \tiny{G} \\
%&&&& &&&&&&\\

\multicolumn{2}{c}{{\scriptsize{{Positive Pairs}}}} & \multicolumn{2}{c}{{\scriptsize{ $\mathrm{OIM_{R}}$}}} & \multicolumn{2}{c}{{\scriptsize{ $\mathrm{OIM_{R}}$+QSSE}}} & \multicolumn{2}{c}{{\scriptsize{{Negative Pairs}}}}& \multicolumn{2}{c}{{\scriptsize{$\mathrm{OIM_{R}}$}}} & \multicolumn{2}{c}{{\scriptsize{$\mathrm{OIM_{R}}$ + QSSE}}} \\

\end{tabular}

\end{center}
\vspace{-0.5cm}

\caption{Class activation maps of $\mathrm{OIM_{R}}$ and $\mathrm{OIM_{R}+QSSE}$ using GradCam~\cite{conf/iccv/SelvarajuCDVPB17}. The left panel shows positive pairs of query (Q) and gallery (G) images from the same class; the right panel shows negative pairs. Red denotes a higher activation value while blue denotes lower. In most cases, both $\mathrm{OIM_{R}}$ and $\mathrm{OIM_{R}+QSSE}$ identify which image part to focus on (\textit{red}-er), but $\mathrm{OIM_{R}+QSSE}$ activations are in general more accurate.}

%The table is split into two parts: the upper part contains samples that OIM+QSSE has correctly classified and OIM not; in the lower part the ones OIM has correctly classified and OIM+QSSE not.

\label{fig:oimVSqsse}
\end{figure*}

%\egroup

%%%%%%%%%%%%%%%%%%%%%%%%%%%%%%%%%

%%%%%%%%%%%%%%%%%%%%%%%%%%%%%%%%%
% \bgroup
%\tabcolsep 1.0pt

%\renewcommand\tabcolsep{1pt}
%\renewcommand{\arraystretch}{0.5}
% \setlength\arrayrulewidth{1.5pt}

\begin{figure*}[h]
\renewcommand\tabcolsep{1pt}
\renewcommand{\arraystretch}{0.5}
\begin{center}
%%%%%%%%%%%%%%%%%%%%%%%%%%%%%%%%%%%%%%%%%%%%%%%%%%%%%%%%%
\begin{tabular}{cccc||cccc}

\includegraphics[scale=0.18, trim=0cm 0cm 0cm 0cm, clip=true]{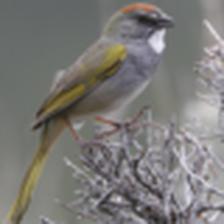}
&
\includegraphics[scale=0.18, trim=0cm 0cm 0cm 0cm, clip=true]{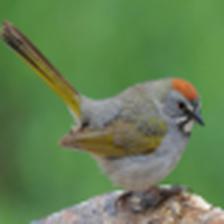} 
&

\includegraphics[scale=0.18,trim=0cm 0cm 0cm 0cm, clip=true]{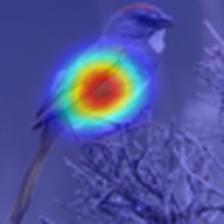}
&
\includegraphics[scale=0.18,trim=0cm 0cm 0cm 0cm, clip=true]{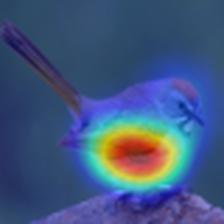} &

\includegraphics[scale=0.18, trim=0cm 0cm 0cm 0cm, clip=true]{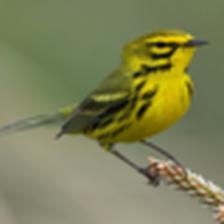}
&
\includegraphics[scale=0.18, trim=0cm 0cm 0cm 0cm, clip=true]{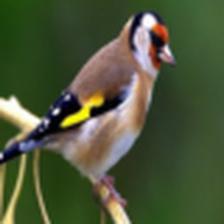} 

&
\includegraphics[scale=0.18,trim=0cm 0cm 0cm 0cm, clip=true]{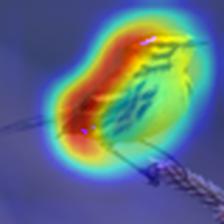}
&
\includegraphics[scale=0.18,trim=0cm 0cm 0cm 0cm, clip=true]{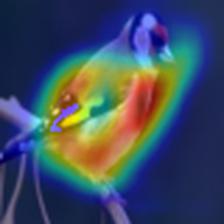} \\
\tiny{Q} & \tiny{G}  & \tiny{Q} & \tiny{G}  &\tiny{Q} & \tiny{G} & \tiny{Q} & \tiny{G} \\
% &&&&&&\\

\multicolumn{4}{c}{{ {\footnotesize{(i) Non-discriminant corresponding pairs}}}} &  
\multicolumn{4}{c}{{\footnotesize{(ii) Focus on background}}}
 \\
 
%\multicolumn{4}{c}{{ {\footnotesize{(i) Non-discriminant corresponding pairs}}}}  &  \multicolumn{4}{c}{{\footnotesize{(ii) Missing correct correspondences}}} &  \multicolumn{4}{c}{{\footnotesize{(iii) Focus on background}}}
 \\

%å&&&&\\

%\\&&& &&&&&&&\\

\end{tabular}

\end{center}

\vspace{-0.8cm}
\caption{Class activation maps of some failure cases, where $\mathrm{OIM_{R}+QSSE}$ could not recognize the correct bird class. Q stands for query and G stands for gallery. Red denotes higher activation value while blue denotes lower. See Sec.~\ref{sec:qualitative_personsearch} for a discussion.
}
\vspace{-0.3cm}
%As shown, failure happens mainly because of two reasons (See discussion in Sec.~\ref{sec:qualitative_personsearch}).
%Left section shows positive pairs where the support and query belong to same class, while pairs on the right are negative ones. 

%The table is split into two parts: the upper part contains samples that OIM+QSSE has correctly classified and OIM not; in the lower part the ones OIM has correctly classified and OIM+QSSE not.

\label{fig:oimVSqsse_failure}
\end{figure*}

%\egroup

%%%%%%%%%%%%%%%%%%%%%%%%%%%%%%%%%

\subsubsection{{Qualitative Results}}
\label{sec:qualitative_personsearch}

In Figure~\ref{fig:cub-qualitative}, we illustrate some sample results of QGN for the 5-way 1-shot case on the CUB dataset. Given a gallery in the first column, we show 5 query examples from each of the 5 (-way) classes in the next 5 columns. In the first four rows, some challenging examples are given where QGN correctly classifies (green box) the gallery image. In the last two rows, there are examples where QGN assigns the gallery to wrong classes (red box). Note that failure cases are also challenging for human observers, as they mainly correspond to matching front to back views of the birds.

Next, in Figure~\ref{fig:oimVSqsse}, we delve into the QSSE component. Using GradCam~\cite{conf/iccv/SelvarajuCDVPB17}, we visualize some class activation maps for the $\mathrm{OIM_{R}}$ and $\mathrm{OIM_{R}}$+QSSE models. With reference to the left panel, reporting positive query (Q) and gallery (G) pairs,
note how the $\mathrm{OIM_{R}}$+QSSE model focuses on corresponding body regions that are mostly discriminative. For example, in the first row / left panel, $\mathrm{OIM_{R}}$+QSSE looks at the discriminant grey wing and yellow beak of the bird in both query and gallery, while $\mathrm{OIM_{R}}$ fails to focus on the wings. In the third row / left panel, high activations spread over the query example for $\mathrm{OIM_{R}}$, while for $\mathrm{OIM_{R}}$+QSSE high activations appear on a region which looks similar to the gallery. With reference to the right panel, reporting negative pairs, note that the head part of the query (yellow bird) is blue in color, while that of the gallery is black, and that $\mathrm{OIM_{R}}$+QSSE focuses only on the discriminant head part. 

In Figure~\ref{fig:oimVSqsse_failure}, we demonstrate some examples where $\mathrm{OIM_{R}}$+QSSE could not recognize the correct bird. The failure happens mainly for two reasons: \textbf{i.}\ when the corresponding pairs attended by QSSE are not discriminative enough;  and \textbf{ii.}\ when $\mathrm{OIM_{R}}$+QSSE focuses on background. In general, the proposed $\mathrm{OIM_{R}}$+QSSE finds the correct discriminative corresponding parts, better than when not using QSSE.

\subsection{Experiments on Person Search}

Here we evaluate QGN on the CUHK-SYSU~\cite{munjal2019cvpr} and PRW~\cite{8099840} datasets; we analyse quantitatively the influence of backbone architectures, input image sizes and the ROI-Pool Vs.\ -Align; and we illustrate the effect of QRPN.
%in the generation of proposals.

%It has rich variations in background, lightning and occlusion conditions, with multiple person scales and poses. 

\subsubsection{Benchmarks and Implementation details } 
\noindent{\textbf{CUHK-SYSU}}~\cite{xiao2017joint} is the most used dataset for evaluating person search. It comprises 18,184 images annotated with 96,143 person bounding boxes of 8,432 identities. The training set contains 11,206 images of 5,532 identities. The test set consists of 6,978 images and 2900 queries.

\noindent 
{\textbf{PRW}}~\cite{8099840} is a dataset acquired by 6 stationary cameras in a university campus. The dataset comprises 11,816 images annotated with 43,110 bounding boxes. The training set includes 5,134 images with 482 identities, while the test set has 6,112 images with 450 identities and 2057 queries.

\noindent{\textbf{Evaluation metrics:}} Following previous works~\cite{xiao2017joint}, we report the performance using two metrics: Common Matching Characteristic (CMC top-K) and mean Average Precision (mAP). CMC top-K is measured as the probability of retrieving at least one match in top-K predictions. Average Precision (AP) is measured for each query by calculating the area under precision-recall curve. mAP is then calculated using the mean of APs for all queries. 

%For both metrics, a box is considered as a match if IoU is $\geq0.5$. 

\noindent \textbf{Implementation Details:}\label{sec:impldetails} We use OIM~\cite{xiao2017joint} as baseline and extend it with the three proposed query-guided components. The images are re-scaled such that their shorter side is 600 pixels, unless mentioned explicitly. All models are trained using SGD for 4 epochs over pre-trained OIM model. The learning rate is set to 0.001, then dropped by a factor of 10 after 2 epochs. CUHK-SYSU considers as query-gallery pairs all combinations for each ID. The training set is further augmented by flipping both query and gallery images. 
For PRW, we sample only three gallery images for each possible query image of an ID, since the number of boxes per ID are already very large. 

% The anchor scales are set to \{2, 4, 8, 16, 32\} and anchor ratios to  \{1, 2, 3\}.  Both RPN and QRPN use the same anchor settings, to allow the scores to be combined later. 

\subsubsection{{Comparison to the state of the art}}
\label{sec:sota}

In Table~\ref{tab:sota_cuhk}, we compare QGN to the state-of-the-art. In the top section, we report joint end-to-end methods, in the bottom section we list cascaded approaches. In each section the approaches are chronologically ordered. 

As visible from the table regarding the CUHK-SYSU dataset, QGN achieves an accuracy of 91.5 mAP and 92.1 top-1, surpassing APNet~\cite{Zhong2020} by 2.6pp mAP and 2.8pp top-1, BINet~\cite{dong2020} by 1.5pp mAP and 1.4pp top-1. Following recent approaches~\cite{Yan_2021_CVPR,DBLP:conf/aaai/HanZGSY21}, we further report the performance of QGN leveraging the better FPN~\cite{Lin_2017_CVPR} backbone. As shown in the last row of the table, FPN+QGN achieves an accuracy of 93.7 mAP and 94.4 top-1, surpassing the most recent joint approaches including DMRNet~\cite{DBLP:conf/aaai/HanZGSY21} by 0.5pp mAP and 0.2pp top-1, DKD~\cite{DBLP:conf/aaai/ZhangWBSY21} by 0.6pp mAP and 0.2pp top-1. Note that FPN+QGN also performs competitive with AlignPS~\cite{Yan_2021_CVPR}, only 0.3pp away in terms of mAP. 

On PRW, QGN achieves an accuracy of 42.9 mAP and 81.9 top-1, surpassing APNet by 1pp mAP and .5pp top-1, NAE by .8pp top-1. Adopting the better FPN backbone further improves the performance. Particularly, FPN+QGN achieves an accuracy of 46.7 mAP and 82.9 top-1, surpassing NAE by 2.7pp mAP and 1.8pp top-1, PGA by 2.5pp mAP, AlignPS by 0.6pp mAP and 0.8pp top-1. Also note that FPN+QGN performs competitive to DMRNet.

\begin{table}[t!]
\setlength{\tabcolsep}{4.7pt}
\renewcommand{\arraystretch}{0.8}
\caption{Comparison with the state-of-the-art on the CUHK-SYSU and PRW datasets. For CUHK-SYSU, gallery size of 100 is used and for PRW the whole test set is used. Methods in the top section are joint models (\textit{Joint}), those in the bottom are cascaded approaches (\textit{Seq}).
\vspace{-0.4cm}
}

\label{tab:sota_cuhk}

\begin{center}
\resizebox{10cm}{!}{
\begin{tabular}{clccccc}
\hline
&&\multicolumn{2}{c}{\textbf{CUHK}}   &\multicolumn{2}{c}{{\textbf{PRW}}} &  \\
 \cmidrule(l){3-4} \cmidrule(l){5-6}
 
& \bf{Method} & \bf{mAP} & \bf{top-1} & {\bf{mAP}} & {\bf{top-1}} & \bf{Publication}\\
\hline
\hline
&OIM~\cite{xiao2017joint}  & 75.5 & 78.7 & {21.3}& {49.9}& CVPR17  \\
%&NPSM~\cite{Liu2017NPSM}  & 77.9 & 81.2& {24.2}& {53.1}& ICCV17\\

&Context~\cite{yan2019}   & 84.1 &  86.5 & {33.4} & {73.6}& CVPR19\\
&APNet~\cite{Zhong2020} & {88.9} &  {89.3} & {41.9}& {81.4}& CVPR20\\
&BINet~\cite{dong2020} & {90.0} &  {90.7} & {45.3}& {81.7}& CVPR20\\
\textit{Joint}&NAE~\cite{Chen_2020_CVPR} & {92.1} &  {92.9}  & {44.0}& {81.1}& CVPR20\\
& {PGA~\cite{Kim_2021_CVPR}} & {92.3}  &  \textbf{94.7} & {44.2}& {85.2}& {CVPR21}\\
& {FPN + AlignPS~\cite{Yan_2021_CVPR}} &  {\textbf{94.0}} &  {{94.5}} & {46.1}& {82.1}& {CVPR21}\\

& {FPN + DMRNet~\cite{DBLP:conf/aaai/HanZGSY21}} & {93.2}  &  {94.2} & {46.9}& {83.3}& {AAAI21}\\
&{DKD~\cite{DBLP:conf/aaai/ZhangWBSY21}} & {93.1} &  {94.2} & {\bf{50.5}} & {\bf{87.1}}& {AAAI21}\\

& \textbf{QGN} & 91.5  &  92.1 & {42.9} & {81.9} & Proposed\\

& \textbf{FPN + QGN} & {93.7}  &  {94.4} & {46.7} & {82.9} & Proposed\\

\hline

%&MGTS~\cite{Chen_2018_ECCV} & 83.0 & 83.7 & {32.6}& {72.1}& ECCV18\\
%&CLSA~\cite{lan_2018_ECCV} & {87.2} & {88.5}& {38.7}& {65.0}& ECCV18 \\
\textit{Seq.} &FPN+RDLR~\cite{Chuchu2019}& {93.0} & {94.2} & {42.9}& {70.2}& ICCV19 \\
&{IGPN}~\cite{Dong_2020_CVPR}&90.3& 91.4& \bf{47.2}& {87.0}& CVPR20 \\
&{TCTS}~\cite{cheng_2020_CVPR} & \bf{93.9} & \bf{95.1}& {{46.9}}& {\bf{87.5}}& CVPR20\\

\hline
\end{tabular}
}
\end{center}
\vspace{-0.4cm}

\end{table}

\begin{table}[t]
\renewcommand{\tabcolsep}{5.5pt}
\renewcommand{\arraystretch}{0.8}

\caption{
Evaluation of our proposed query-guided components on CUHK-SYSU~\cite{xiao2017joint} dataset. We present results for gallery size 100 using Resnet50 and Resnet18 architectures. All models in this table use an image size of 600. The OIM~\cite{xiao2017joint} results in the first row are taken from the original paper. OIM in the second row is our own implementation. In the last row, we report the results for our final model, OIM + QSSE + QRPN + QSimNet, which we dub as QGN. 
\vspace{-0.3cm}
}
\begin{center}
\resizebox{10cm}{!}{
\begin{tabular}{lcccc}
%\hline

\hline
 & \multicolumn{2}{c}{\textbf{ ResNet50}} & \multicolumn{2}{c}{\textbf{ ResNet18}}   \\
 \cmidrule(l){2-3} \cmidrule(l){4-5}
\bf{Model} & \bf{mAP} & \bf{top-1}  & \bf{mAP} & \bf{top-1}   \\
\hline
\hline
OIM \cite{xiao2017joint}   & 75.5 & 78.7 & - & -\\

{OIM} (Baseline)  & 77.2 & 77.6 & 70.0 & 69.7 \\
+ \textit{QSSE} & 80.1 & 80.6 & 73.7 & 73.9\\ 
+ \textit{QRPN} & 79.6 &80.4 & 73.9 & 73.5\\
+ \textit{QSimNet}  & 82.6 & 83.0 &75.3& 75.3\\ 
{+ \textit{QSSE} + \textit{QRPN}} & {82.4} & {82.8} & {74.7} & {74.4}\\ 
{+ \textit{QSSE} + \textit{QSimNet}} & {83.3} & {83.4} & {76.1}& {75.9} \\ 
+ \textit{QRPN} + \textit{QSimNet}  &83.1  & 83.3 & {75.9} & {75.5}\\ 
+ \textit{QSSE} + \textit{QRPN} + \textit{QSimNet} (= QGN)  & \bf{84.4} & \bf{84.4} & \bf{78.4} & \bf{77.7}\\
\hline

\end{tabular}
}
\end{center}
\vspace{-0.5cm}

\label{tab:ablation}
\end{table}

\subsubsection{{Ablation Studies}}
\label{ps_ablation}

First we evaluate the impact of QGN components, then the effect of model hyper-parameters on both OIM and QGN.

\noindent\textbf{QGN components.}
In Table~\ref{tab:ablation}, we quantify the improvements of the QGN components when integrated into OIM~\cite{xiao2017joint}, considering two network architectures (ResNet50, ResNet18) and gallery size 100.
We re-implement OIM, named \textit{Baseline} in the table, yielding slightly better performance than \cite{xiao2017joint}. As illustrated, each QGN component improves over OIM. Also, improvements are consistent for each component across different backbone architectures. Taking the representative case of ResNet50, the baseline OIM (77.2 mAP) is improved by 2.9pp with QSSE (80.1 mAP), it is improved by 2.4pp with QRPN (79.6 mAP), and by 5.4pp with QSimNet (82.6 mAP), which is the strongest single component.

QGN components are also complementary. In Table~\ref{tab:ablation}, considering ResNet50, {QSSE+QRPN gives 82.4 mAP, QSSE+QSimNet gives 83.3 mAP,} QRPN+QSimNet gives 83.1 mAP, and the full QGN set (QSSE+QRPN+QSimNet) reaches 84.4 mAP. This means an overall improvement \textit{wrt} the baseline OIM of 7.2pp.

\begin{table}[t]
\renewcommand{\tabcolsep}{11.5pt}
\renewcommand{\arraystretch}{0.75}
\caption{{Person search accuracy and parameter size of OIM+QRPN ResNet18 model at different reduction ratios. We evaluate on CUHK-SYSU dataset using gallery size 100.} 
\vspace{-0.3cm}
}
\begin{center}
\resizebox{7cm}{!}{
\begin{tabular}{cccc}
\hline
{\textbf{Ratio $r$}} & {\textbf{mAP}}& {\textbf{top-1}}  & {\textbf{Params (M)}}   \\
\hline
\hline
{2} & {74.0} & {73.7} &  {15.3}\\
{4} & {73.7} & {73.6} & {14.5}\\
{8} & {73.9} & {73.6} & {14.1}\\
{16} & {73.9} & {73.5} & {13.9}\\
{32} & {72.9} & {72.6} & {13.8}\\

\hline

\end{tabular}
}
\end{center}
\vspace{-0.5cm}
\label{tab:reduction_ratio}
\end{table}

\noindent{\textbf{Reduction Ratio ${r}$ of QRPN.} For QRPN we choose reduction ratio $r$ to be 16 as in~\cite{Hu_2018_CVPR}. Our experiments (cf. Table~\ref{tab:reduction_ratio}) also confirm this to be a reasonable choice as it maintains a good balance between mAP and parameter size.}

\begin{table}
\setlength{\tabcolsep}{4.5pt}
\renewcommand{\arraystretch}{0.8}
\caption{Person search accuracy on CUHK-SYSU and PRW datasets, using different design choices. For CUHK-SYSU, the standard gallery size of 100 is used and for PRW the whole test set is used. Second column gives the image size. $Pool(n)$ refers to ROI pool operation and $Align(n)$ refers to ROI align operation with output size $n\times n$. gCat refers to the concatenation of globally pooled ROI align feature with ClsIdenNet output feature (Fig.~\ref{fig:network}).
\vspace{-0.3cm}
} 

\begin{center}
\resizebox{10cm}{!}{
\begin{tabular}{ccccccccc}
\hline
& & & &  &\multicolumn{2}{c}{\textbf{CUHK}}   &\multicolumn{2}{c}{\textbf{PRW}} \\
\cmidrule(l){6-7} \cmidrule(l){8-9}
 \bf{Model} & \bf{imSize} & \bf{ROI}  & \bf{bSize} & \bf{gCat}  & \bf{mAP}  & \bf{top-1}& \bf{mAP}  & \bf{top-1}\\
\hline
\hline
OIM & 600 & $Pool(7)$ &  1 & &77.2 & 77.6 & 29.2&  65.0\\
OIM & 600 & $Pool(14)$ &  1 &  & 80.8 & 80.9 & 32.8 & 71.3\\
OIM & 600 & $Align(14)$ & 1 & & 81.2 & 81.7 & 33.6 & 71.4\\
OIM & 900 & $Align(14)$ & 1& & 83.9 & 84.2 & 36.9 & 75.7\\
OIM & 900 & $Align(14)$ & 2& &86.1  & 87.8 & 38.7 & 78.4\\
{OIM} & {900} & {$Align(14)$} & {2}& {\checkmark} & {88.6}  &  {88.8} & {40.4} & {79.2}\\
\textbf{QGN} & 900 & $Align(14)$  & 2 & \checkmark  &\textbf{91.5}  &  \textbf{92.1} & \textbf{42.9} & \textbf{80.9} \\
\hline

\end{tabular}
}
\end{center}

\vspace{-0.5cm}
\label{tab:cuhk_analysis}
\end{table}

\noindent{\textbf{Hyper-parameters of OIM and QGN.}}
\label{sec:stronger_baseline}
In Table~\ref{tab:cuhk_analysis}, we evaluate different design choices for OIM and QGN using the ResNet50 backbone.
%on CUHK-SYSU and PRW datasets.

\noindent \textbf{CUHK-SYSU}: As shown in the first few rows, the OIM baseline (77.2 mAP) improves by 3.6pp (80.8 mAP) when adopting the larger ROI pooling size $14\times14$ (Vs.\ the standard $7\times7$). It further improves slightly by 0.4pp (81.2 mAP) when switching to the more complex pooling method, ROI-Align. It improves by 2.7pp (83.9 mAP) when considering larger input images (smaller size re-scaled to 900 Vs. 600). Also, a larger batch size gives additional improvement taking the accuracy to 86.1 mAP (row 5). Following NAE~\footnote{\label{nae}\url{https://github.com/DeanChan/NAE4PS}}, OIM may be further improved by concatenating globally pooled 1024-d features after ROI align with 2048-d feature from ClsIdenNet, bringing the OIM accuracy to 88.6 mAP. We treat this particular OIM as our baseline. Adding QGN components on top of this baseline gives our proposed model QGN, with a performance of 91.5 mAP.

\noindent {\textbf{PRW}: Similarly on PRW dataset, largest improvements are due to increasing the pool size (32.8 Vs. 29.2 mAP), image size (36.9 Vs. 33.6 mAP), batch size (38.7 Vs. 36.9 mAP) and using finer features with gCat (40.4 Vs. 38.7 mAP). As shown in the last row, our proposed QGN gives an accuracy of 42.0 mAP.}

\noindent {\textbf{Discussion on Runtime.} Our method jointly processes each query-gallery pair. This means, for a test set of M queries and N galleries, an exhaustive search of $M \times N$ combinations is required, which makes it computationally expensive. However, note that in practical person search scenarios M is usually a small number (typically 1, i.e only one query person is being searched).}

%%%%%%%%%%%%%%%%%%%%%%%%%%%%%%%%%
% \bgroup
\tabcolsep 1.0pt
\renewcommand{\arraystretch}{0.5}

\newlength{\cuhkfigheight}
\setlength{\cuhkfigheight}{1.1cm}

\begin{figure}[t]
\begin{center}
%%%%%%%%%%%%%%%%%%%%%%%%%%%%%%%%%%%%%%%%%%%%%%%%%%%%%%%%%
\begin{tabular}{ccc|ccc}
%probe 

\includegraphics[scale=0.038]{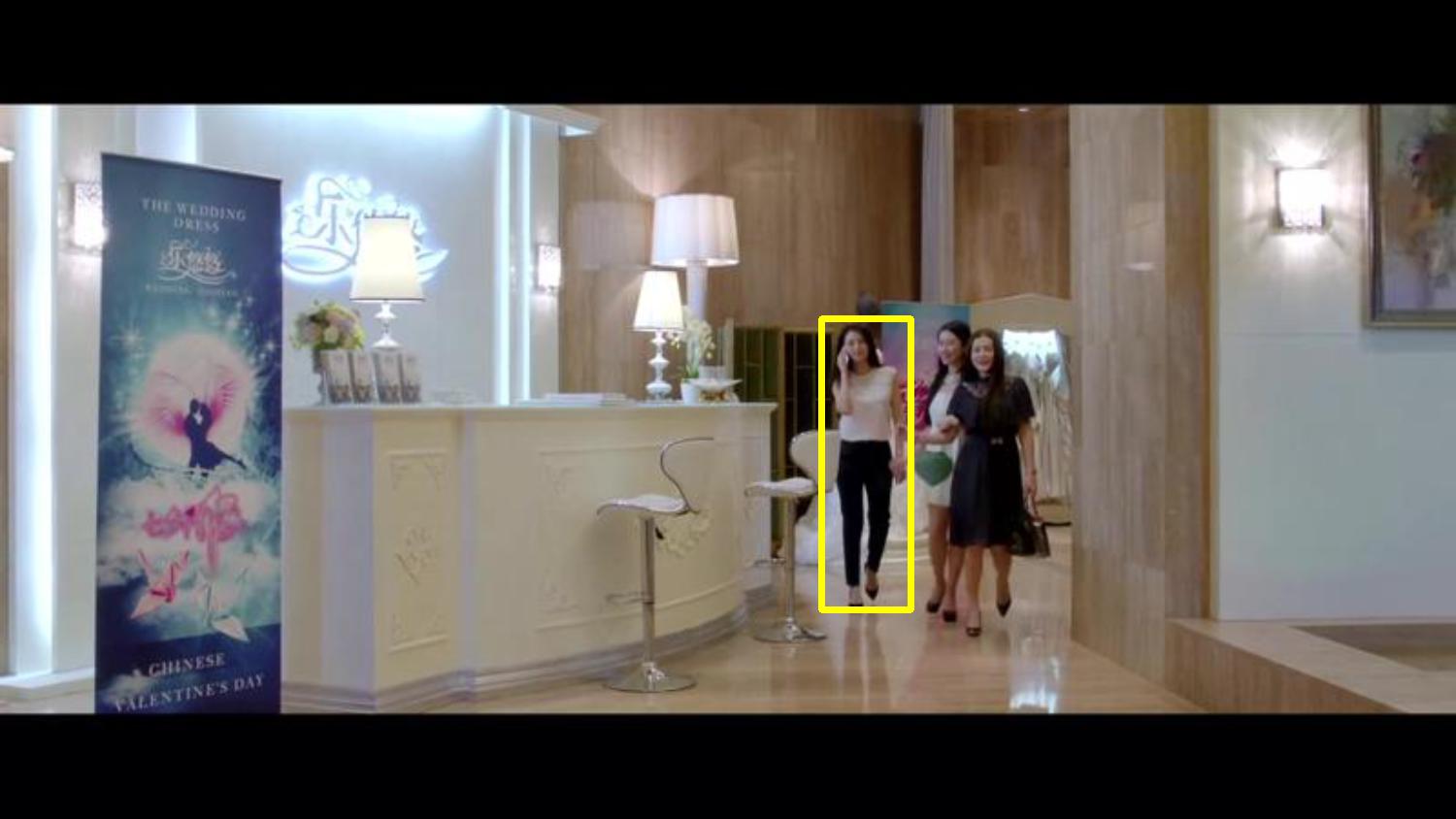}&
\includegraphics[ scale=0.038]{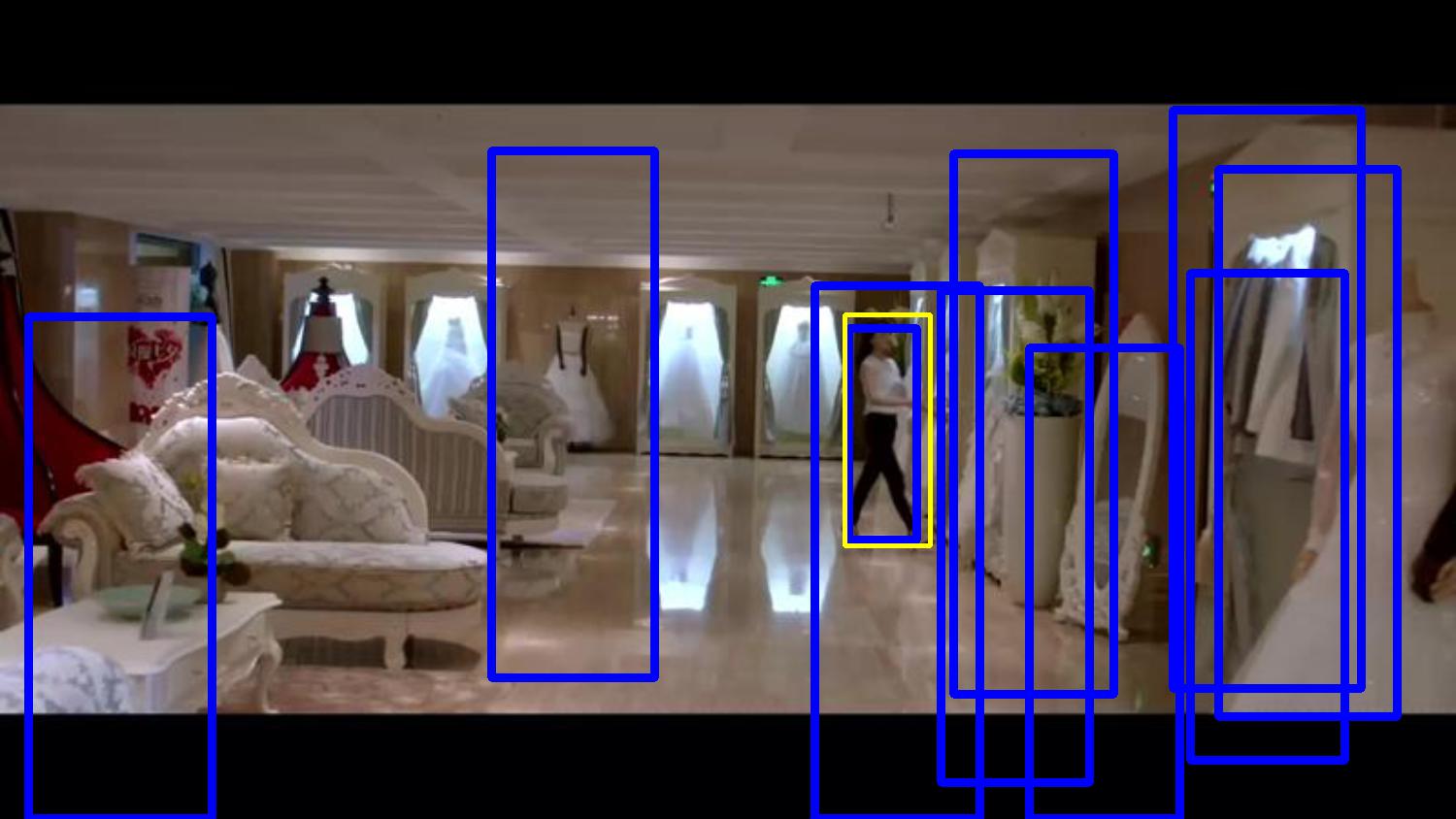}&
\includegraphics[scale=0.038,,]{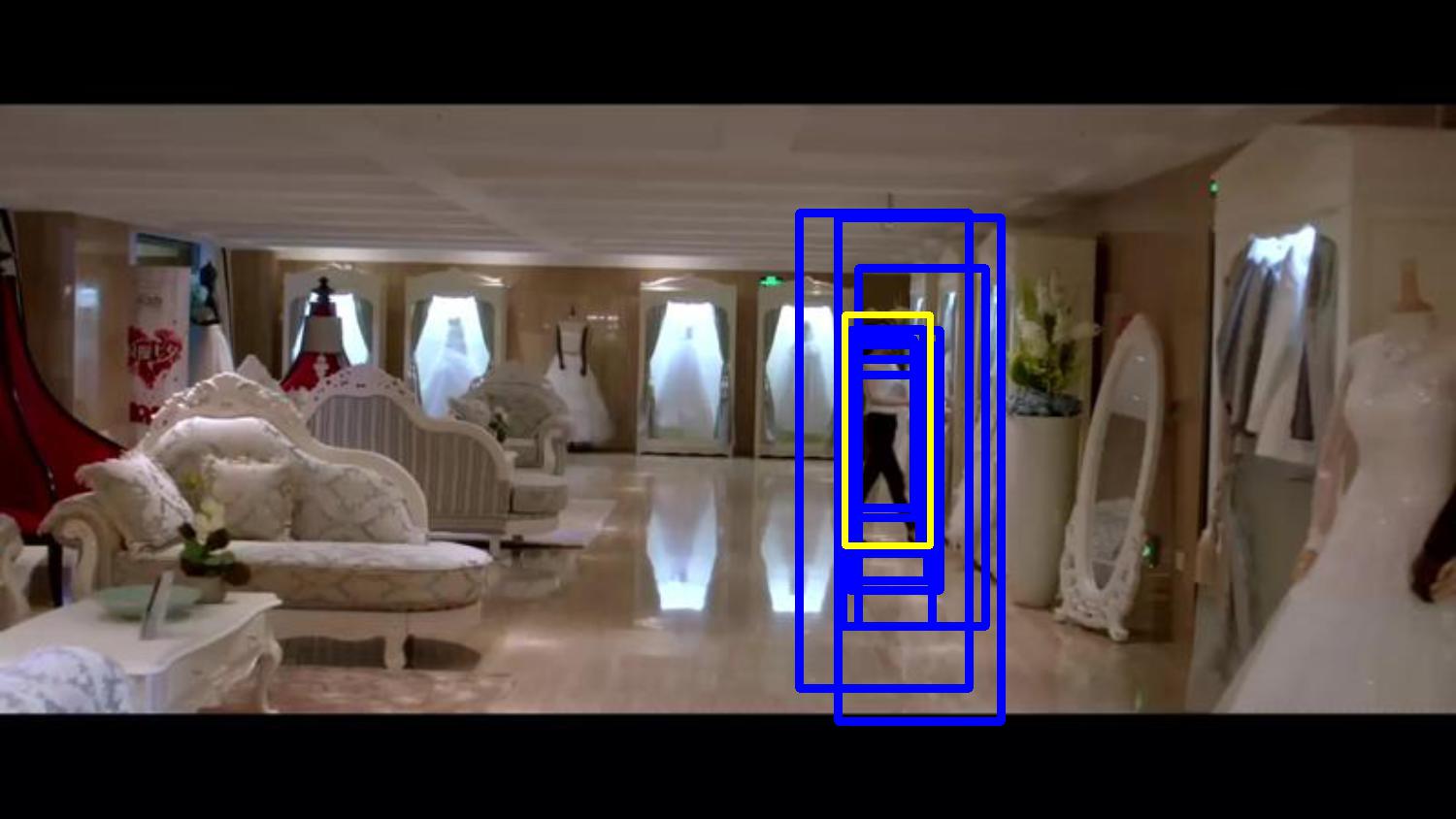} & 

\includegraphics[scale=0.038]{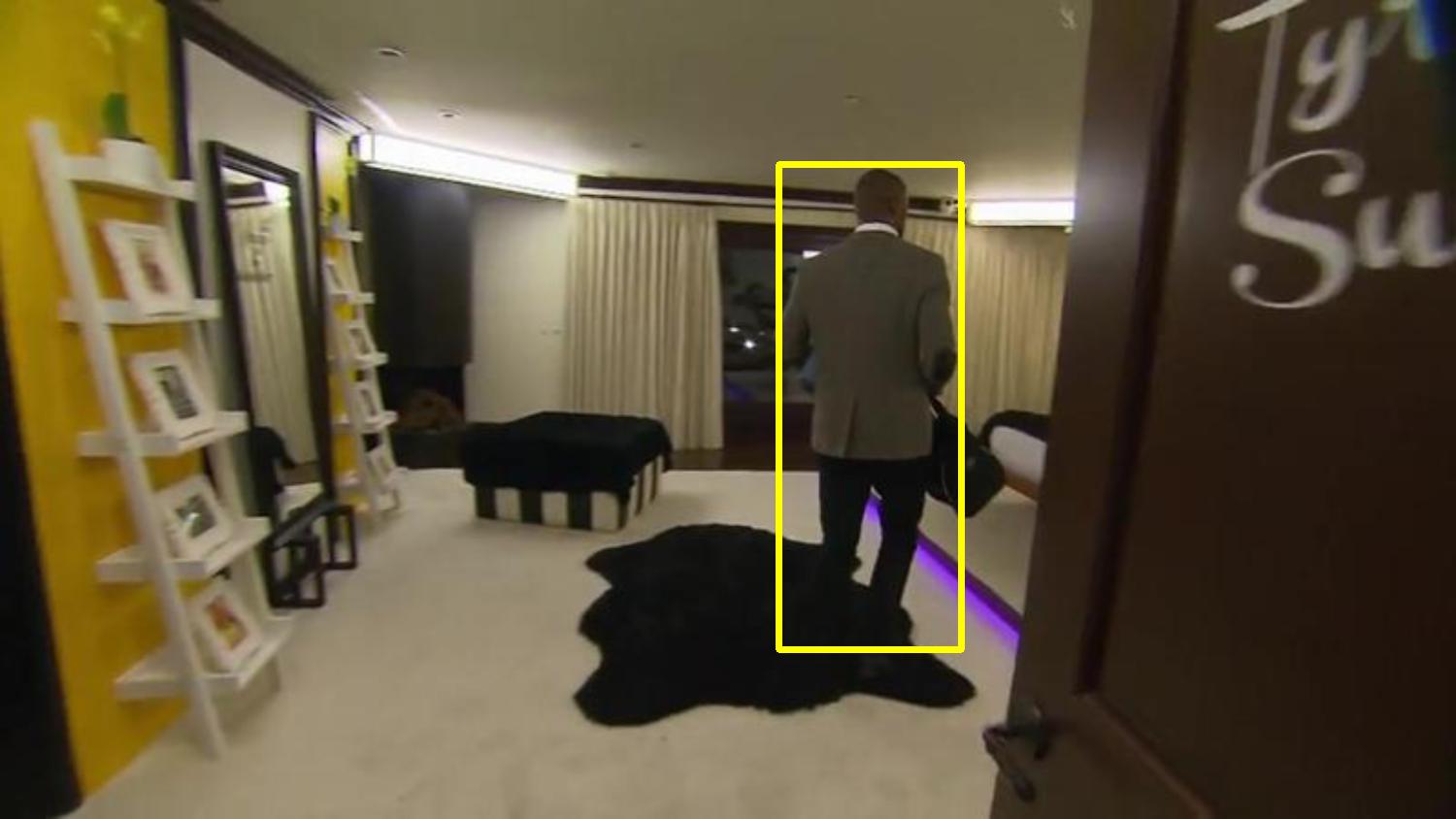}&
\includegraphics[scale=0.038]{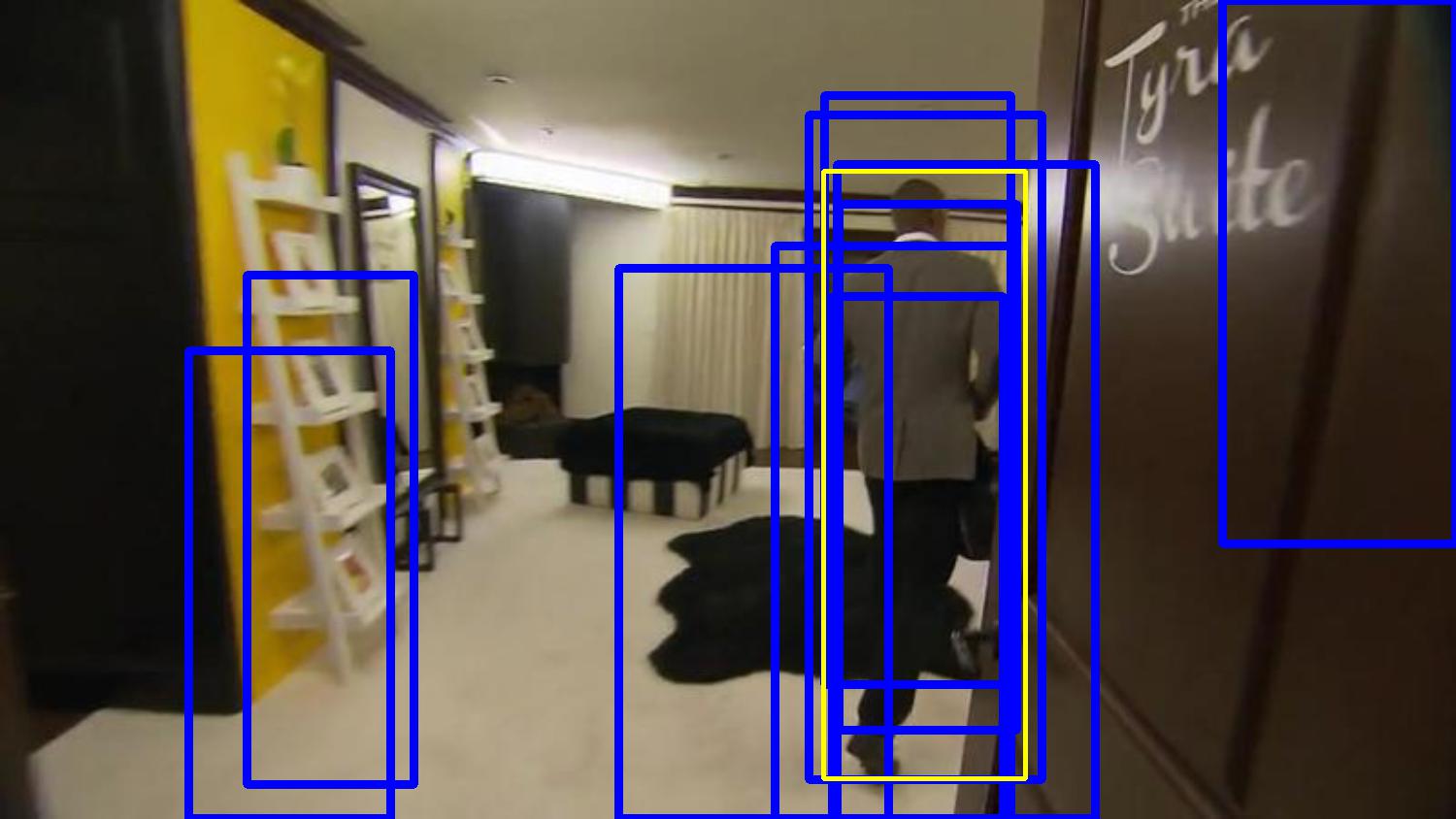}&
\includegraphics[scale=0.038]{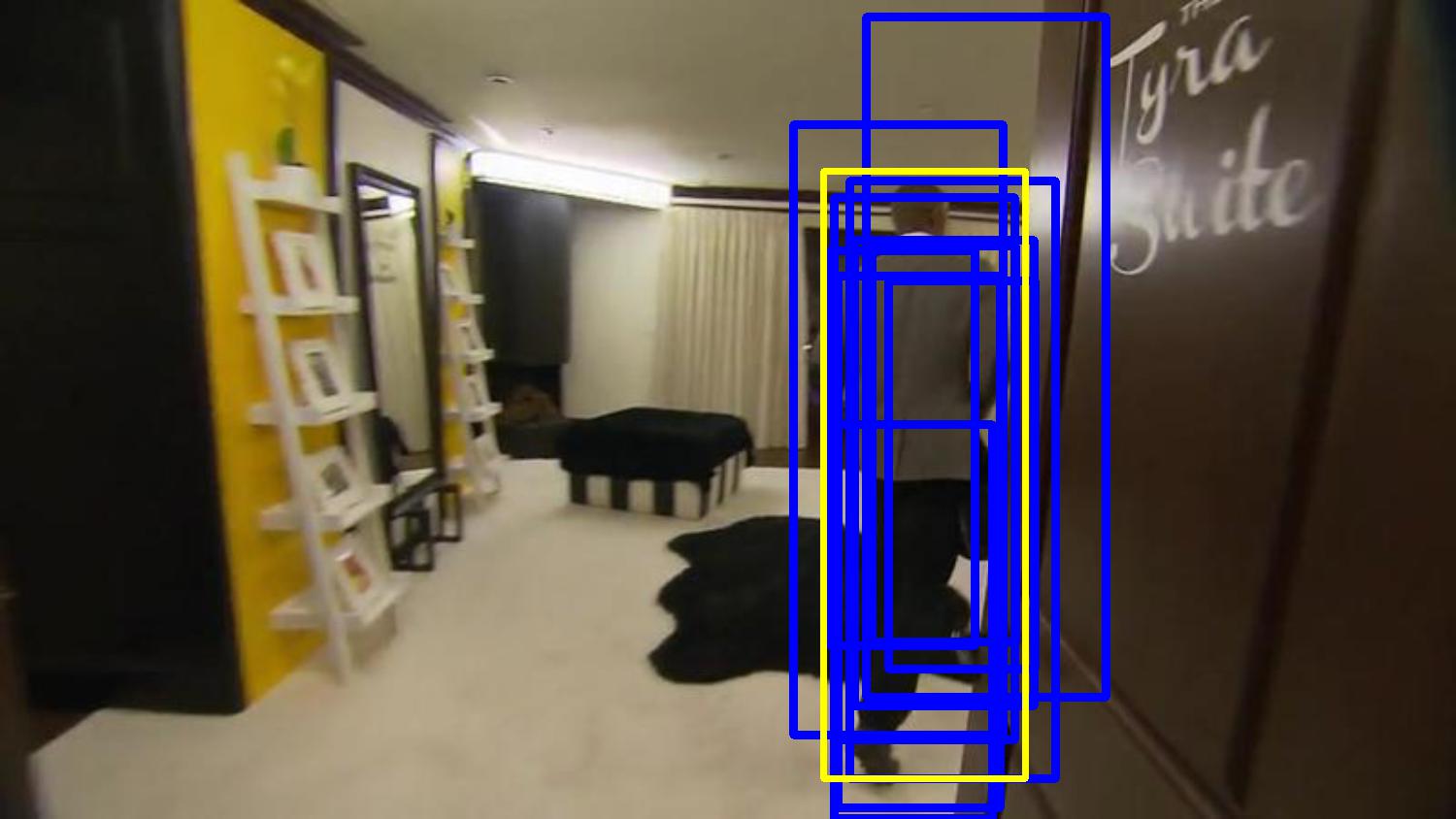}\\

\includegraphics[scale=0.038, ]{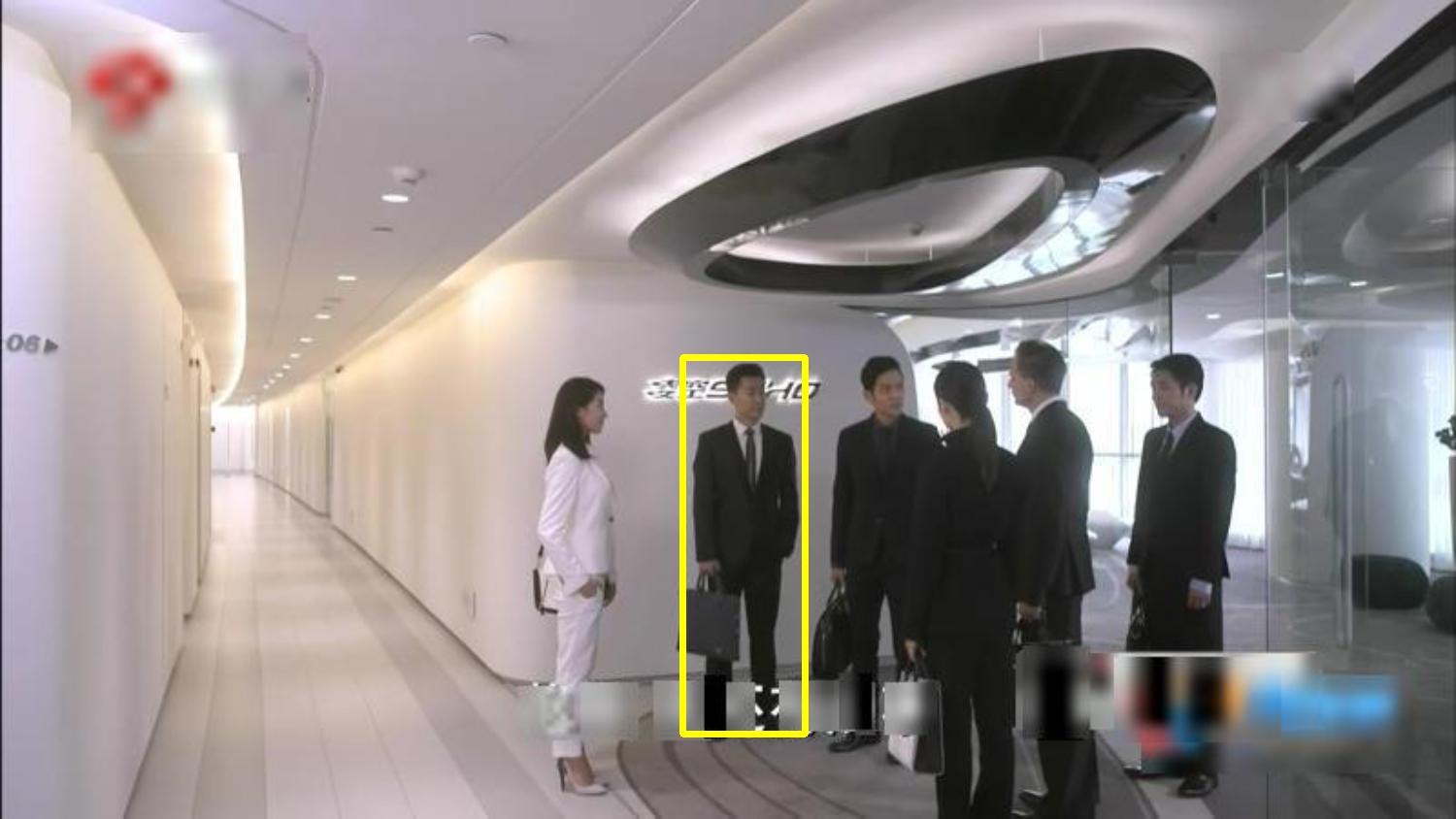}&
\includegraphics[ scale=0.038,]{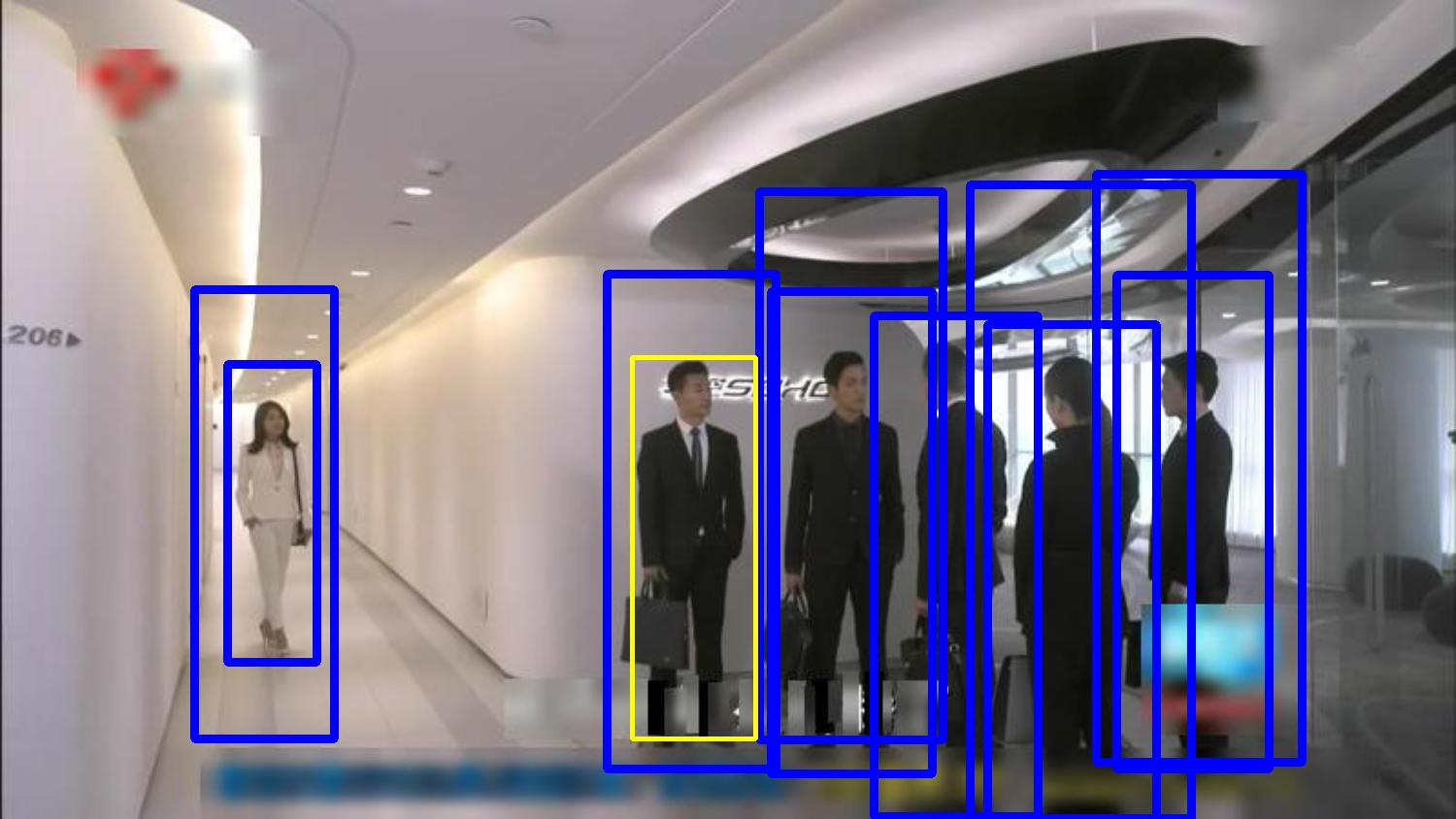}&
\includegraphics[ scale=0.038,]{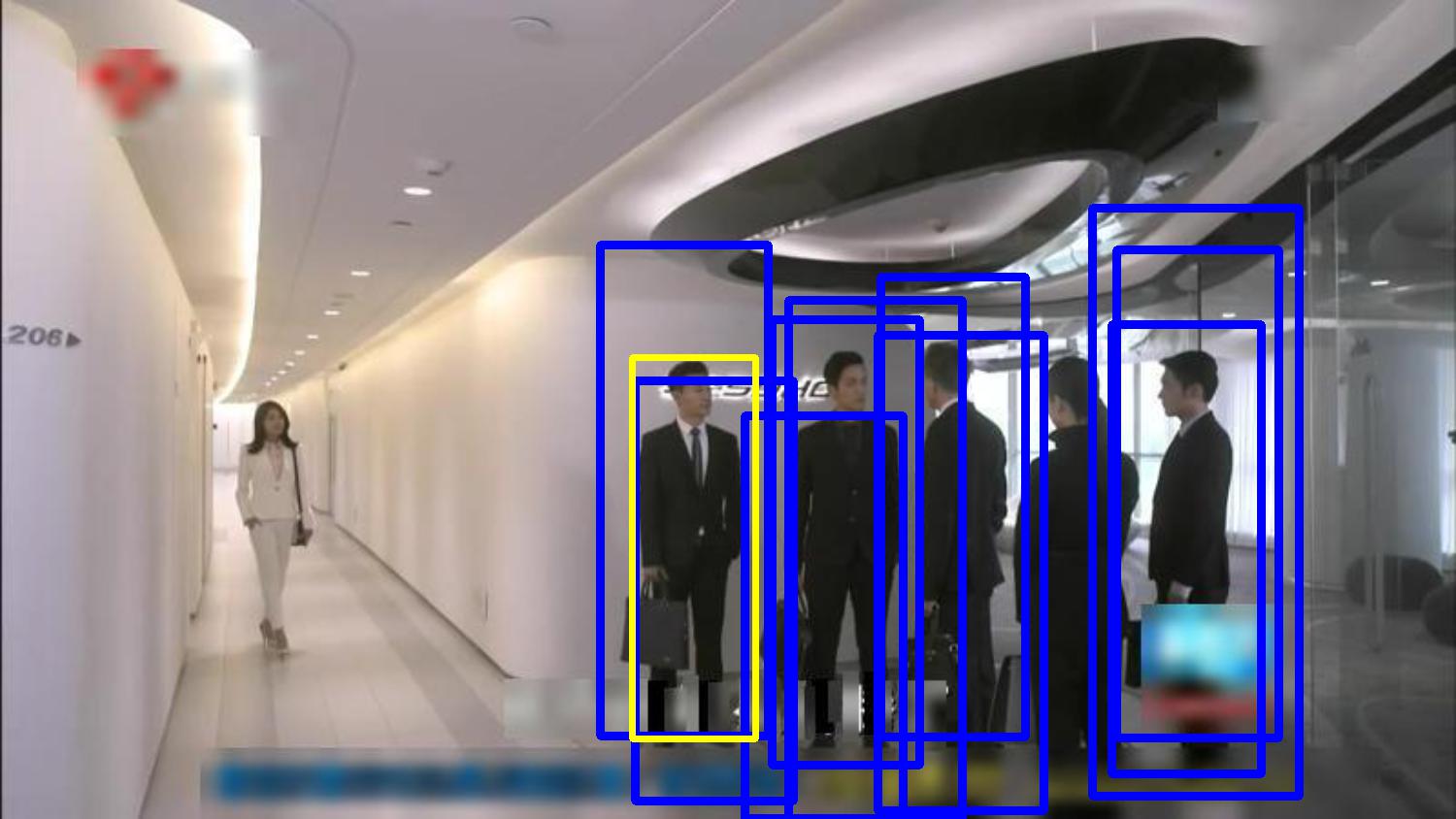}&

 \includegraphics[scale=0.038,]{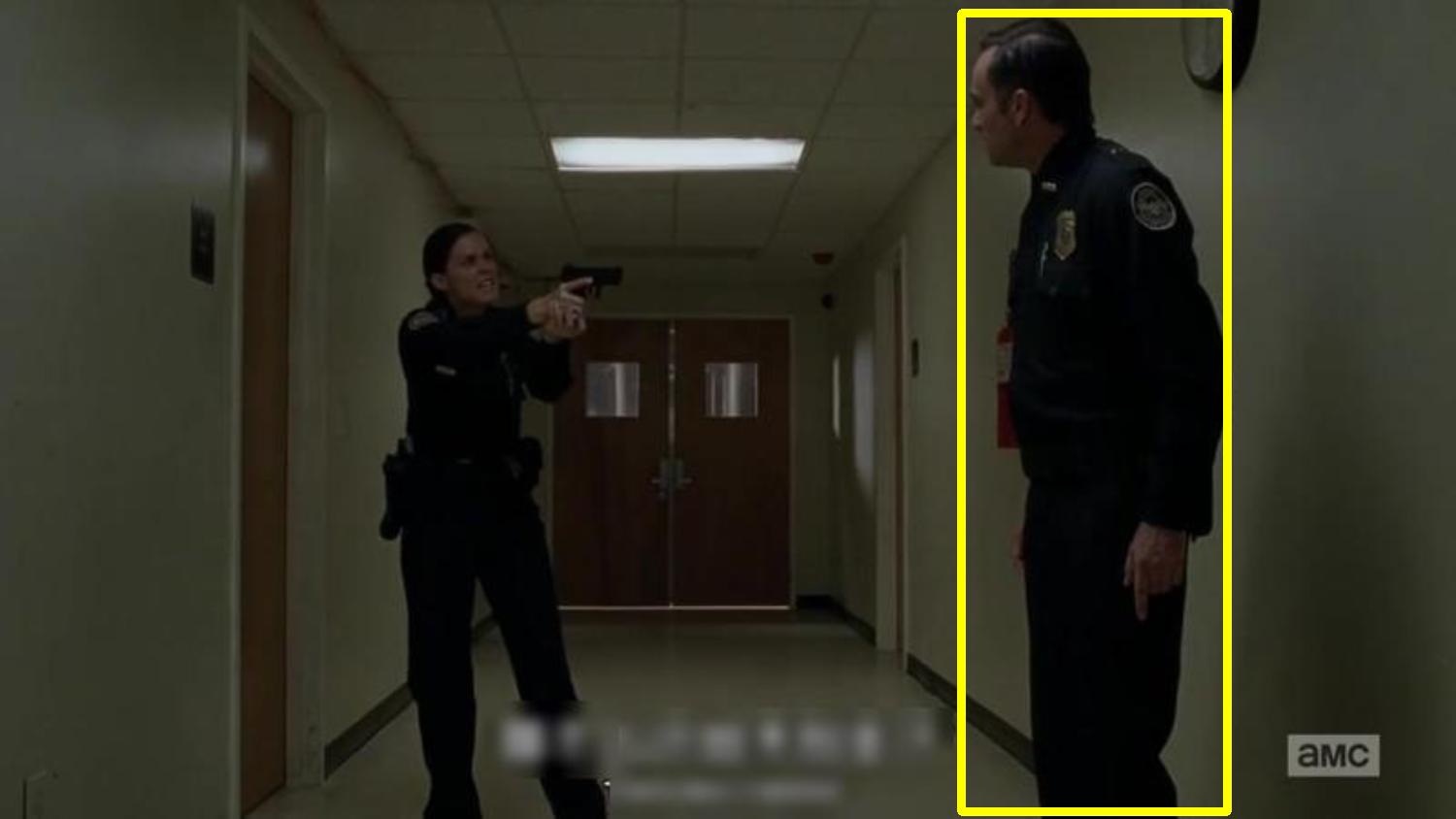}&
 \includegraphics[scale=0.038,]{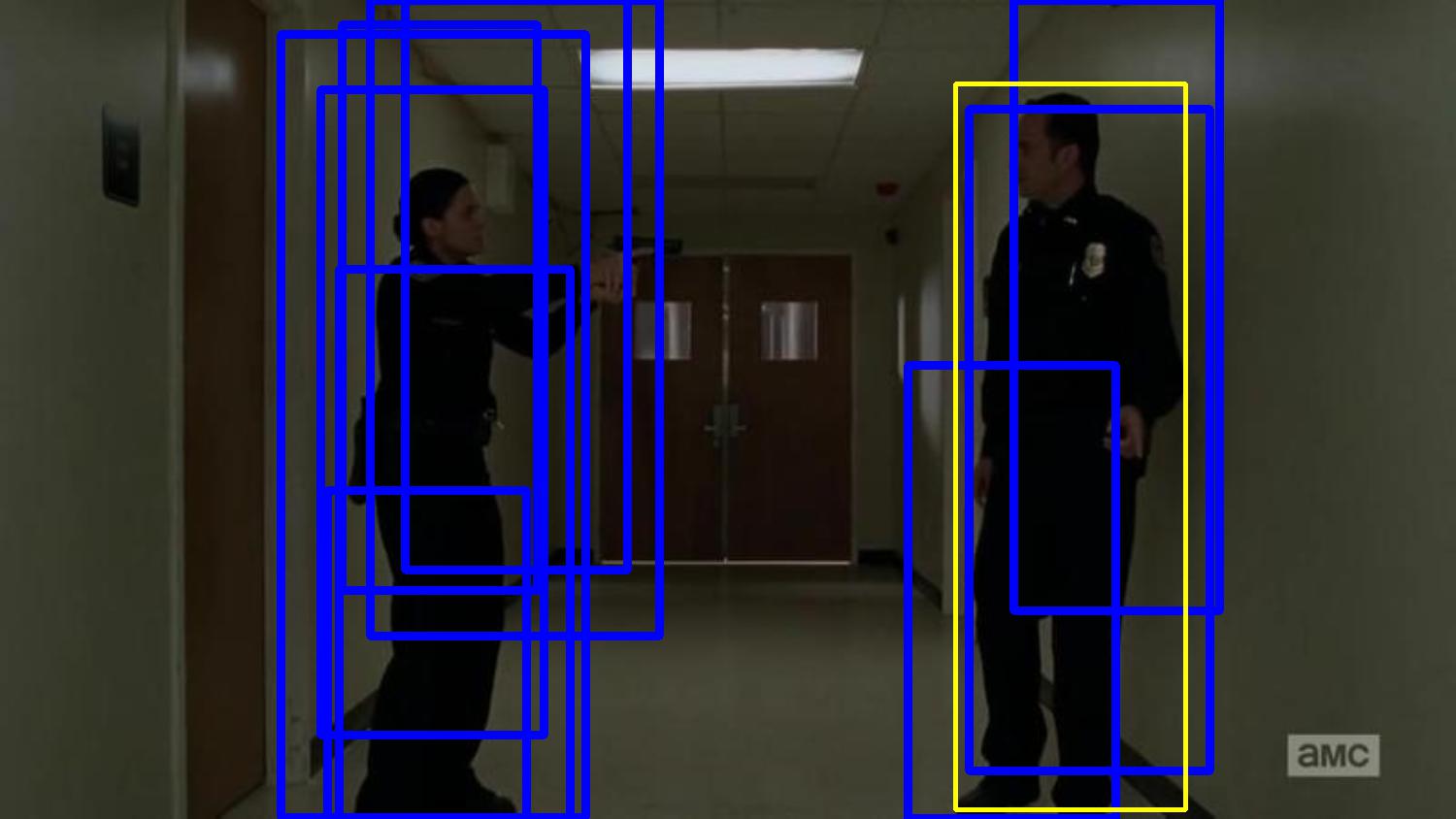}&
 \includegraphics[scale=0.038,]{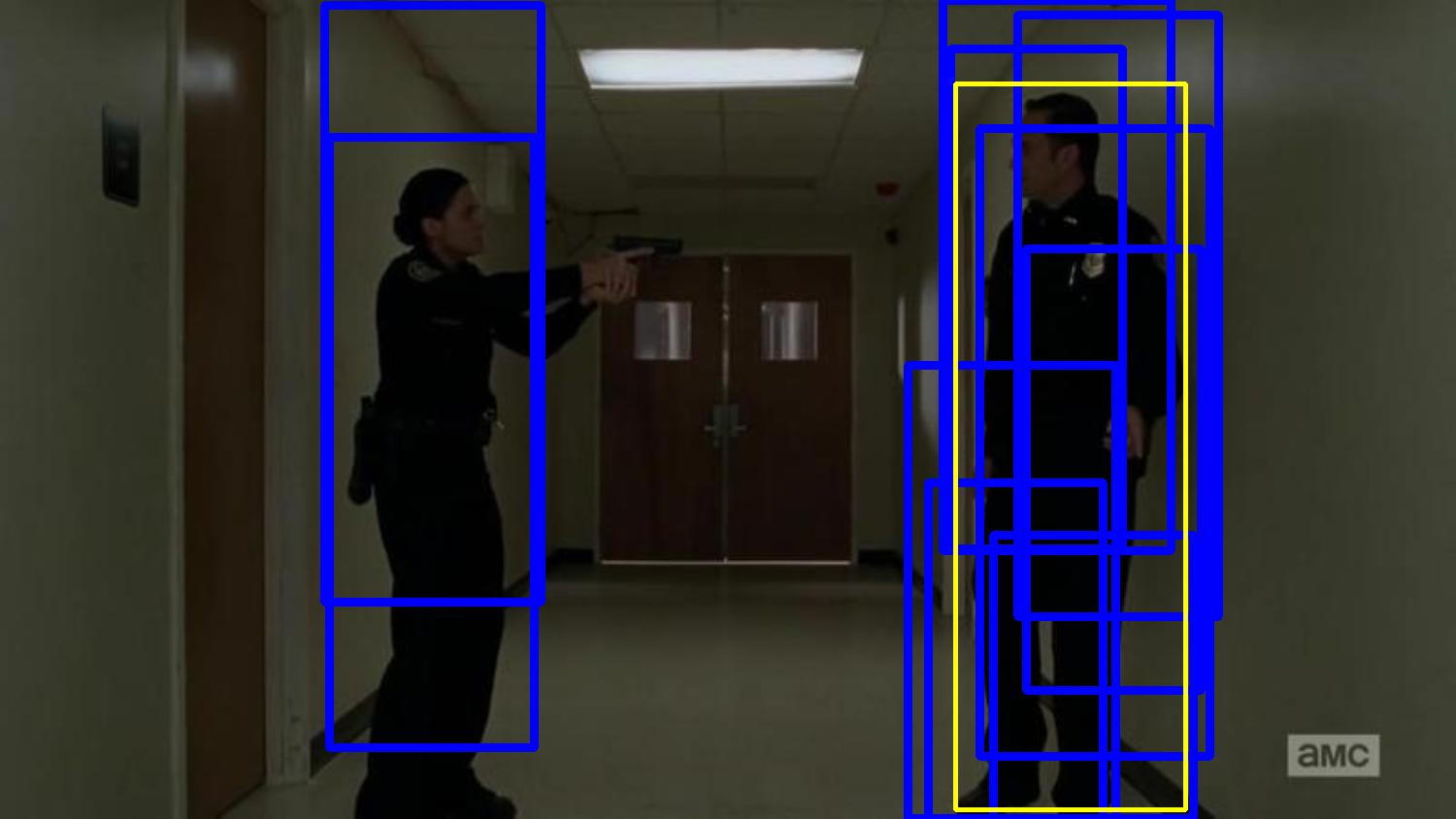}\\

\includegraphics[scale=0.038, ]{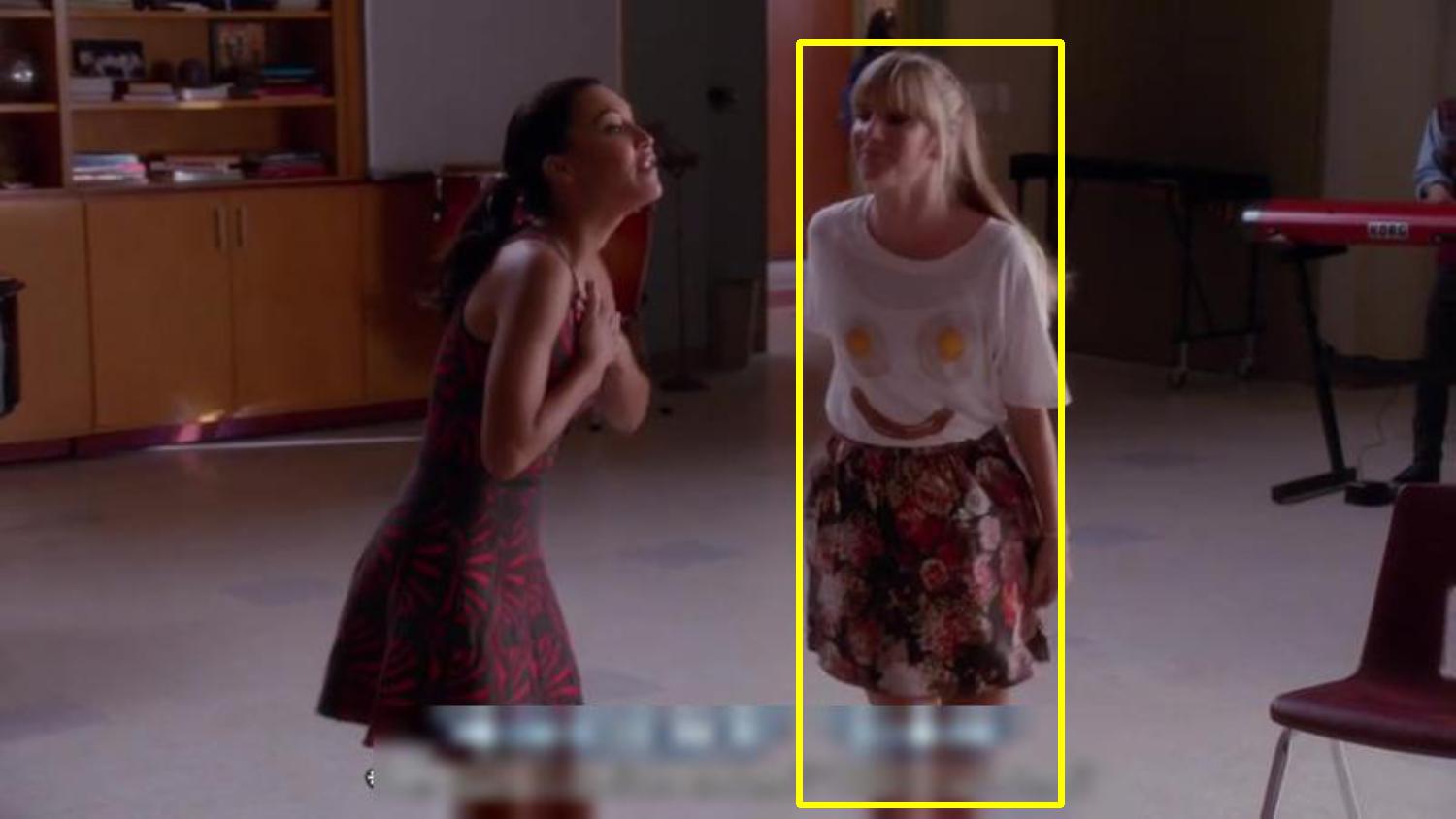}&
\includegraphics[ scale=0.038,]{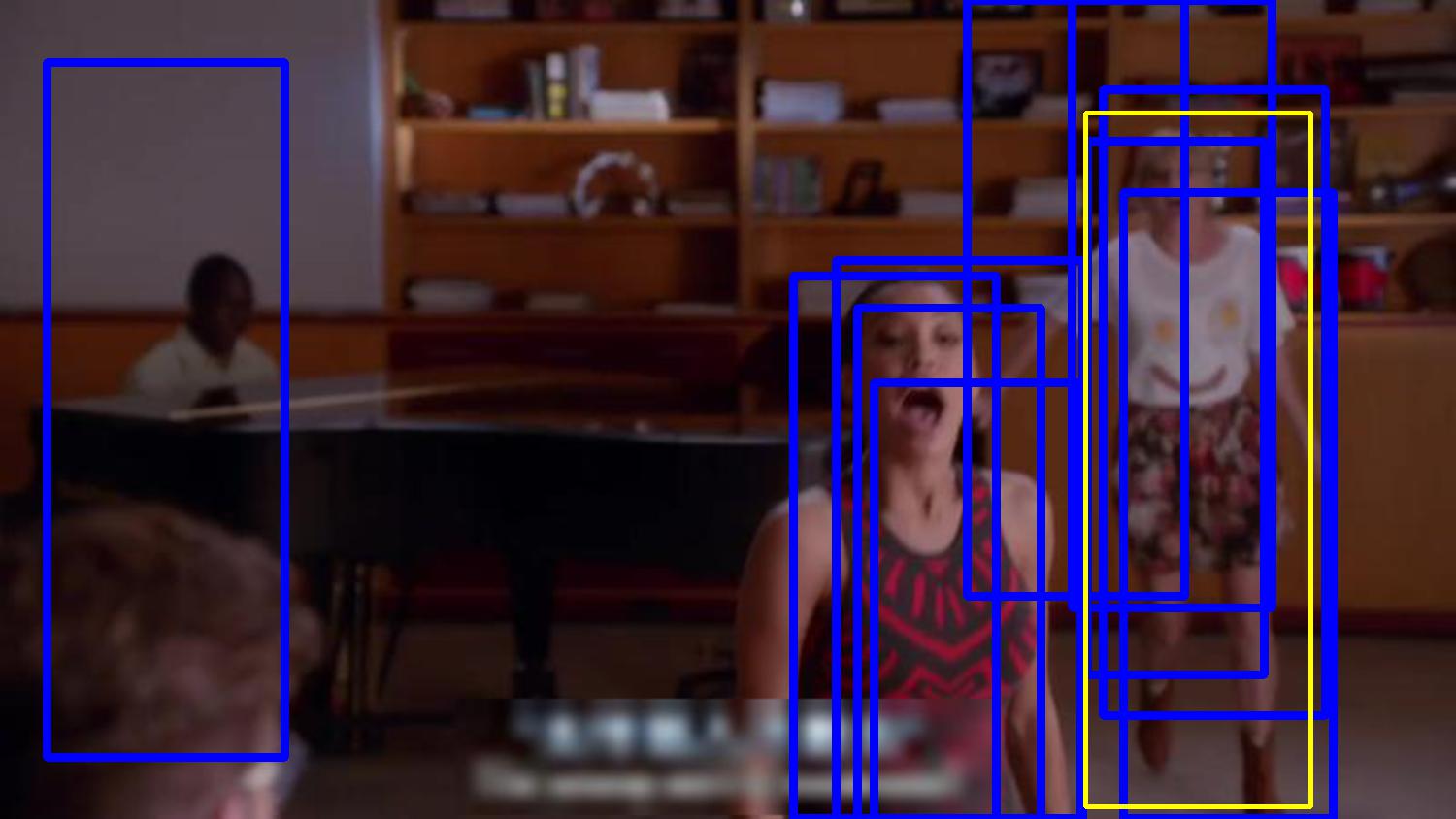}&
\includegraphics[scale=0.038, ]{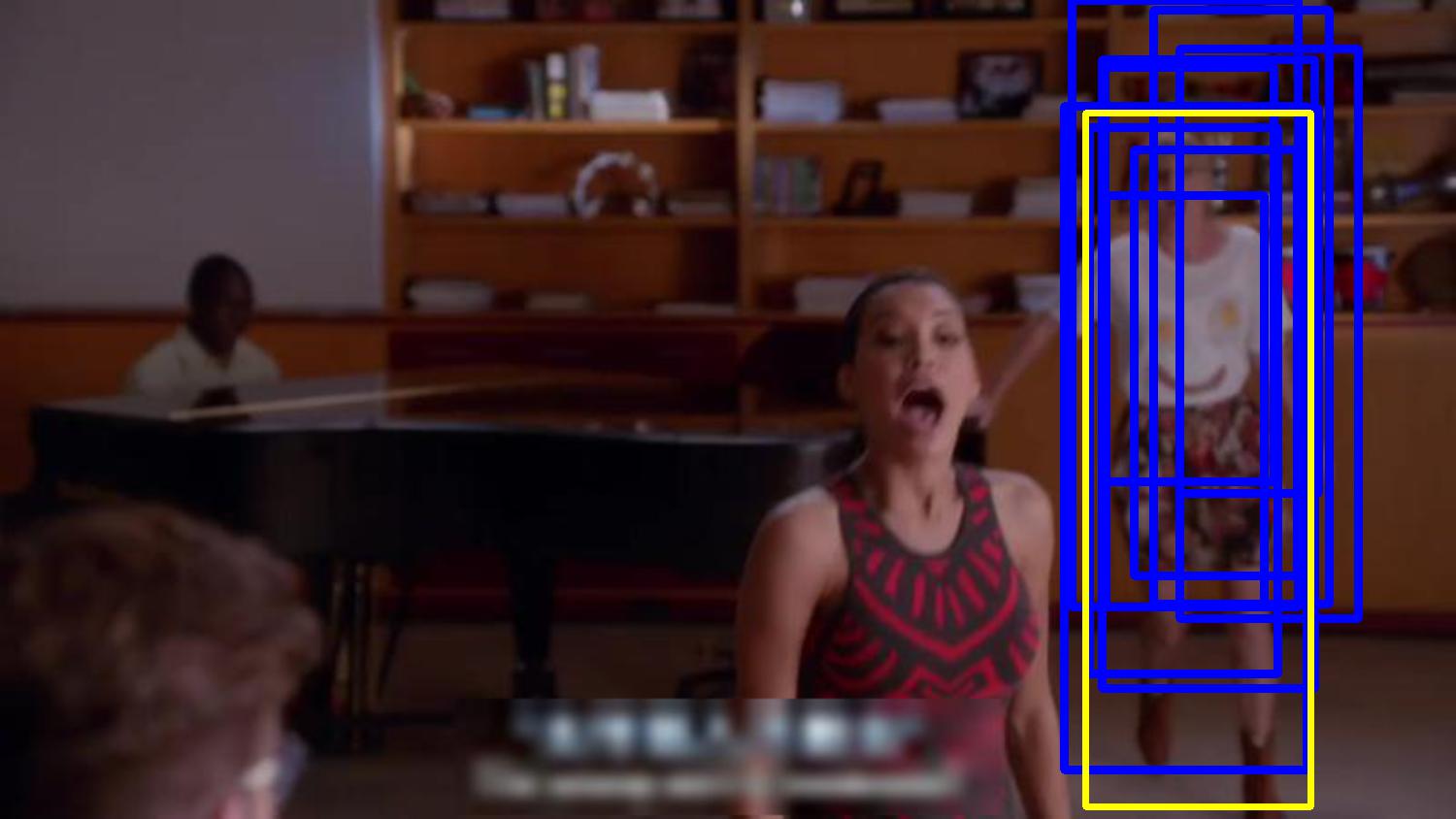}&

\includegraphics[ scale=0.038,]{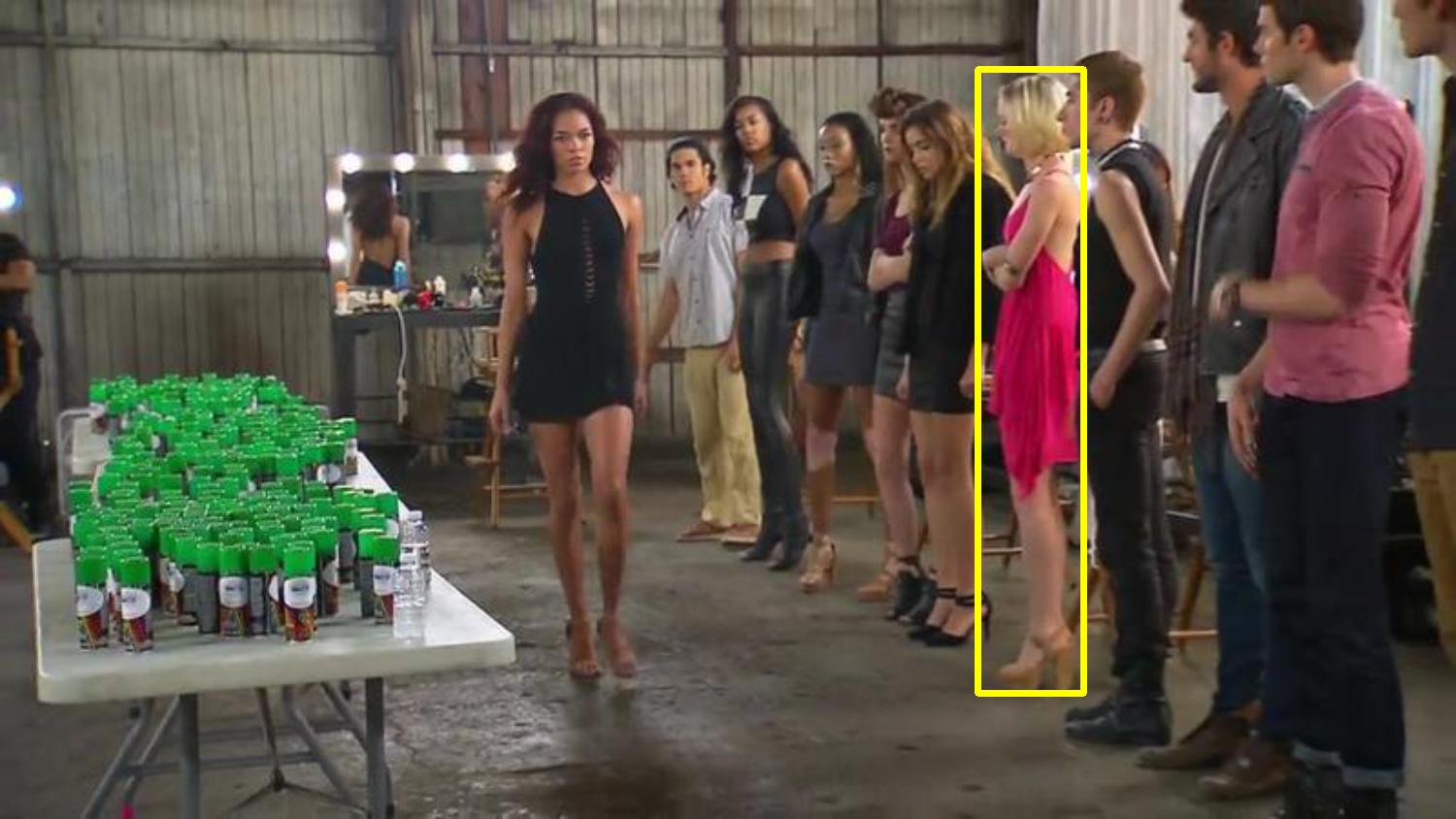}&
\includegraphics[ scale=0.038,]{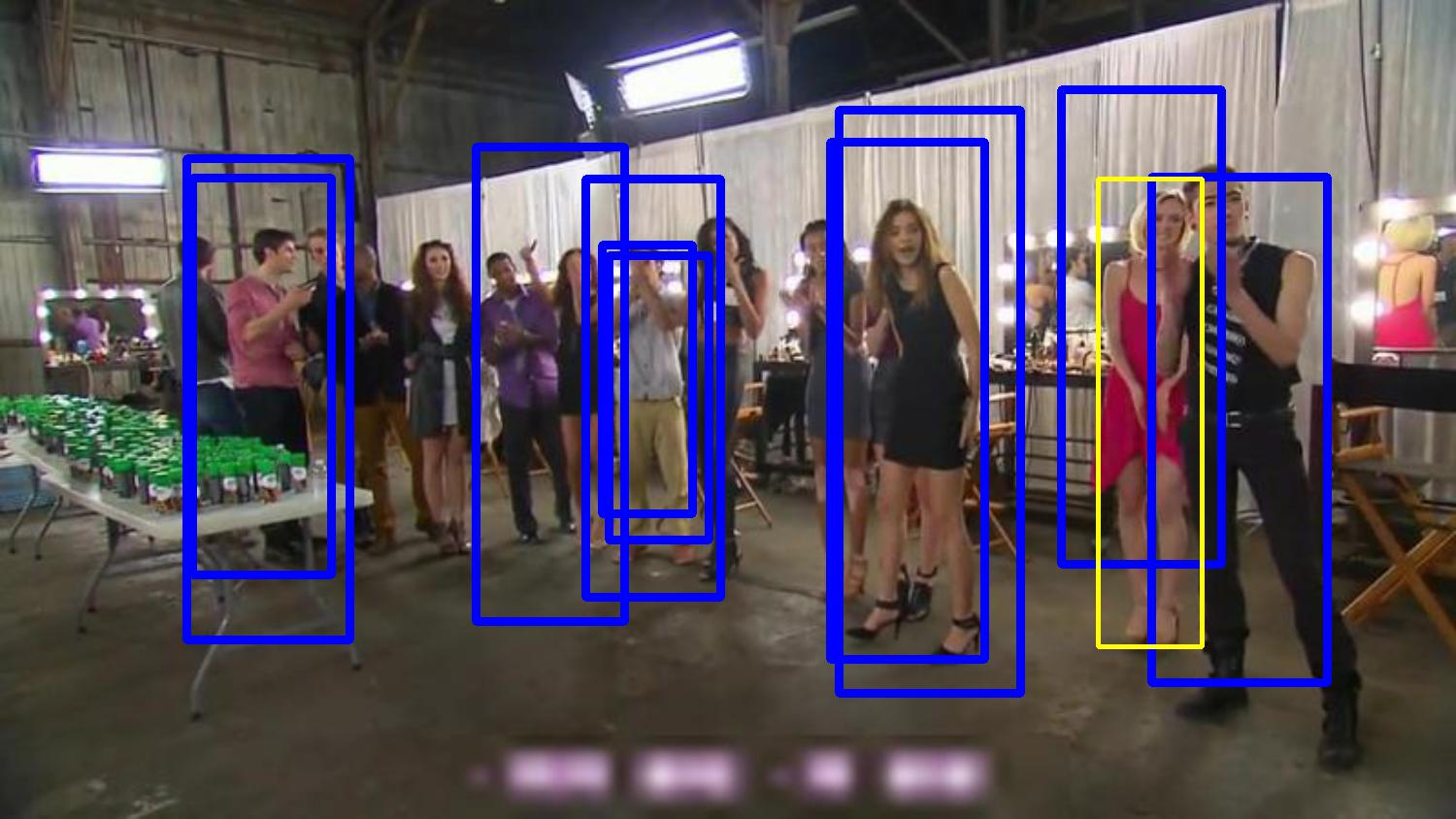}&
\includegraphics[scale=0.038, ]{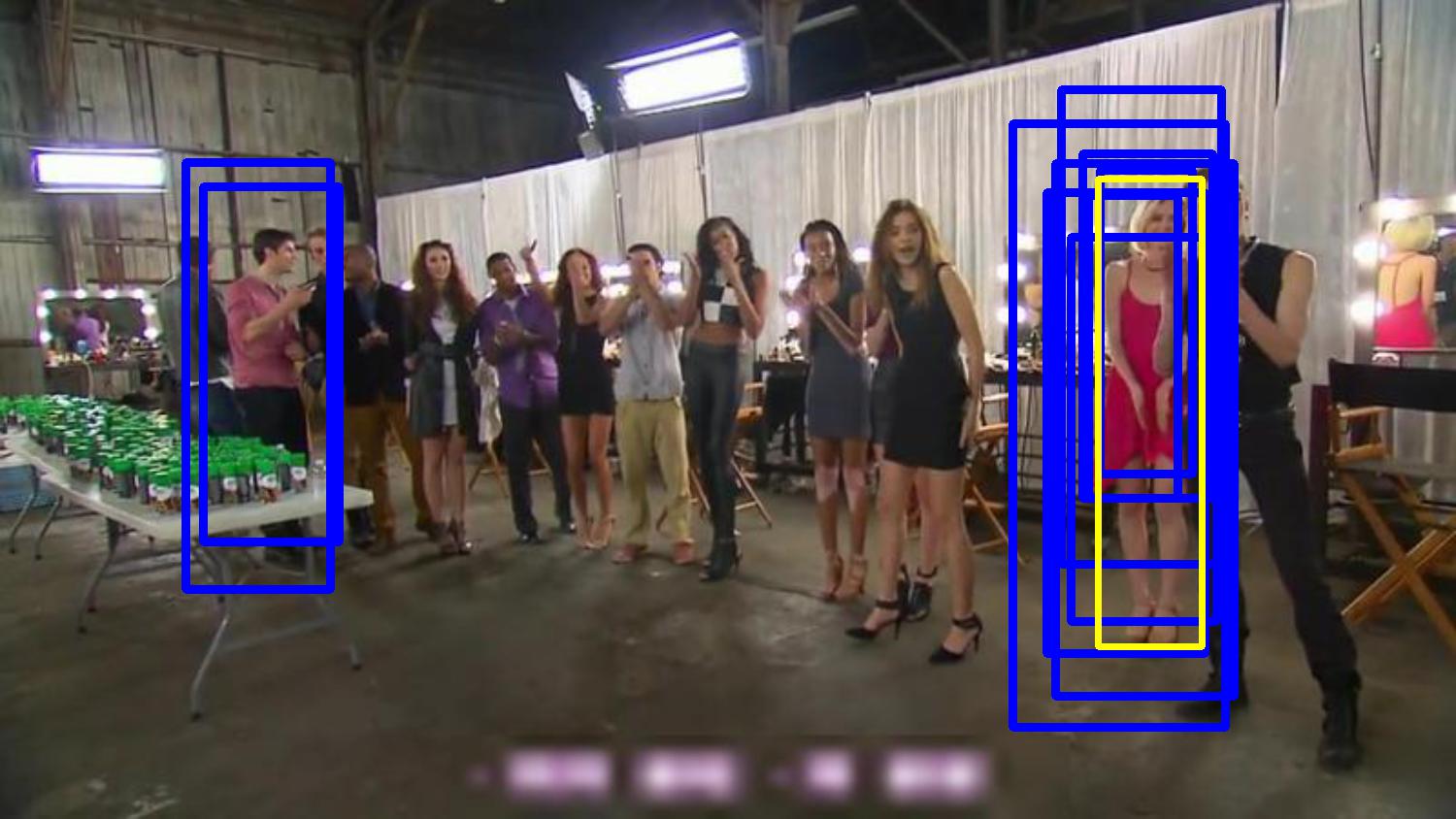}\\

\includegraphics[ scale=0.038,]{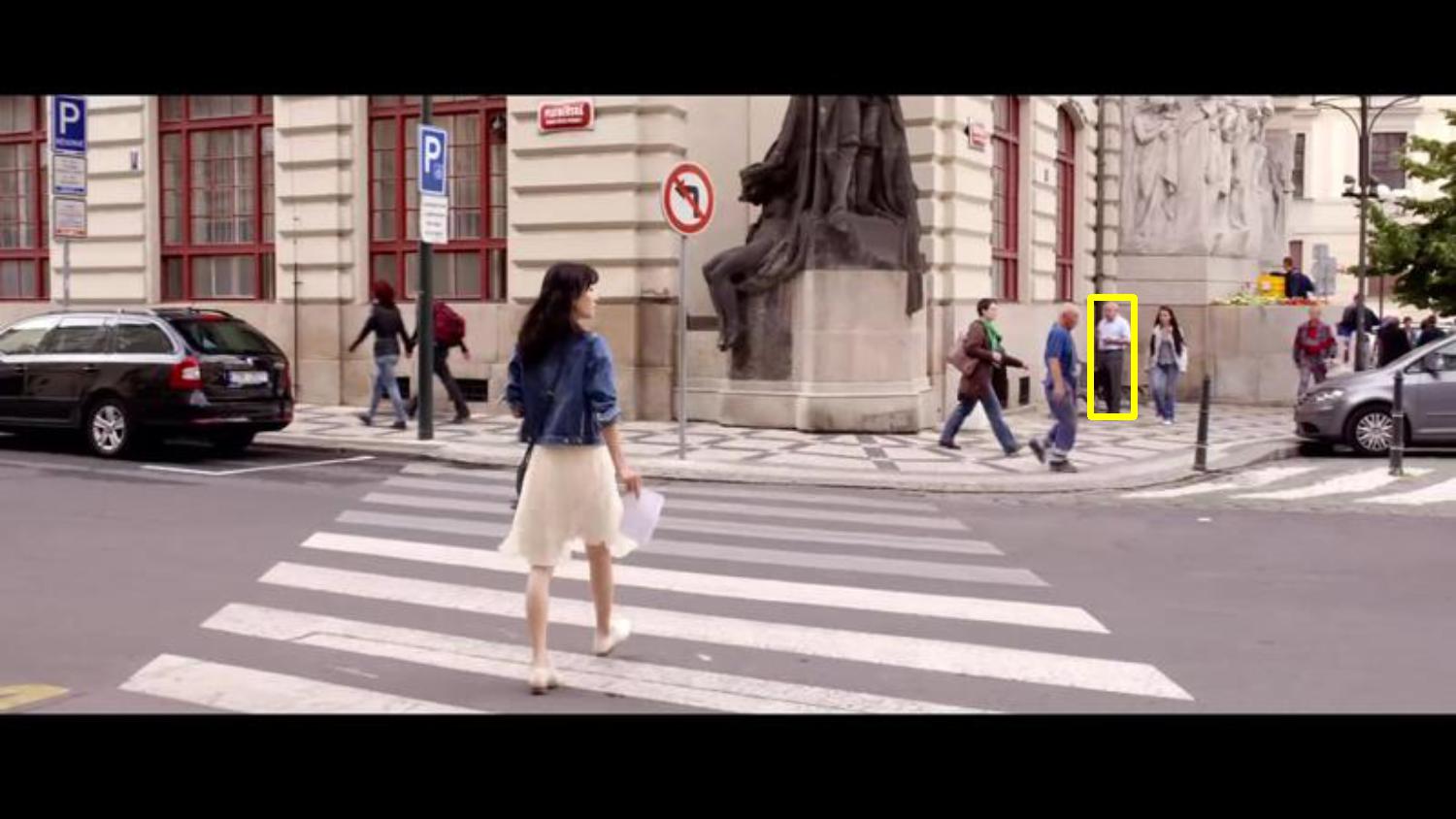}&
\includegraphics[ scale=0.038,]{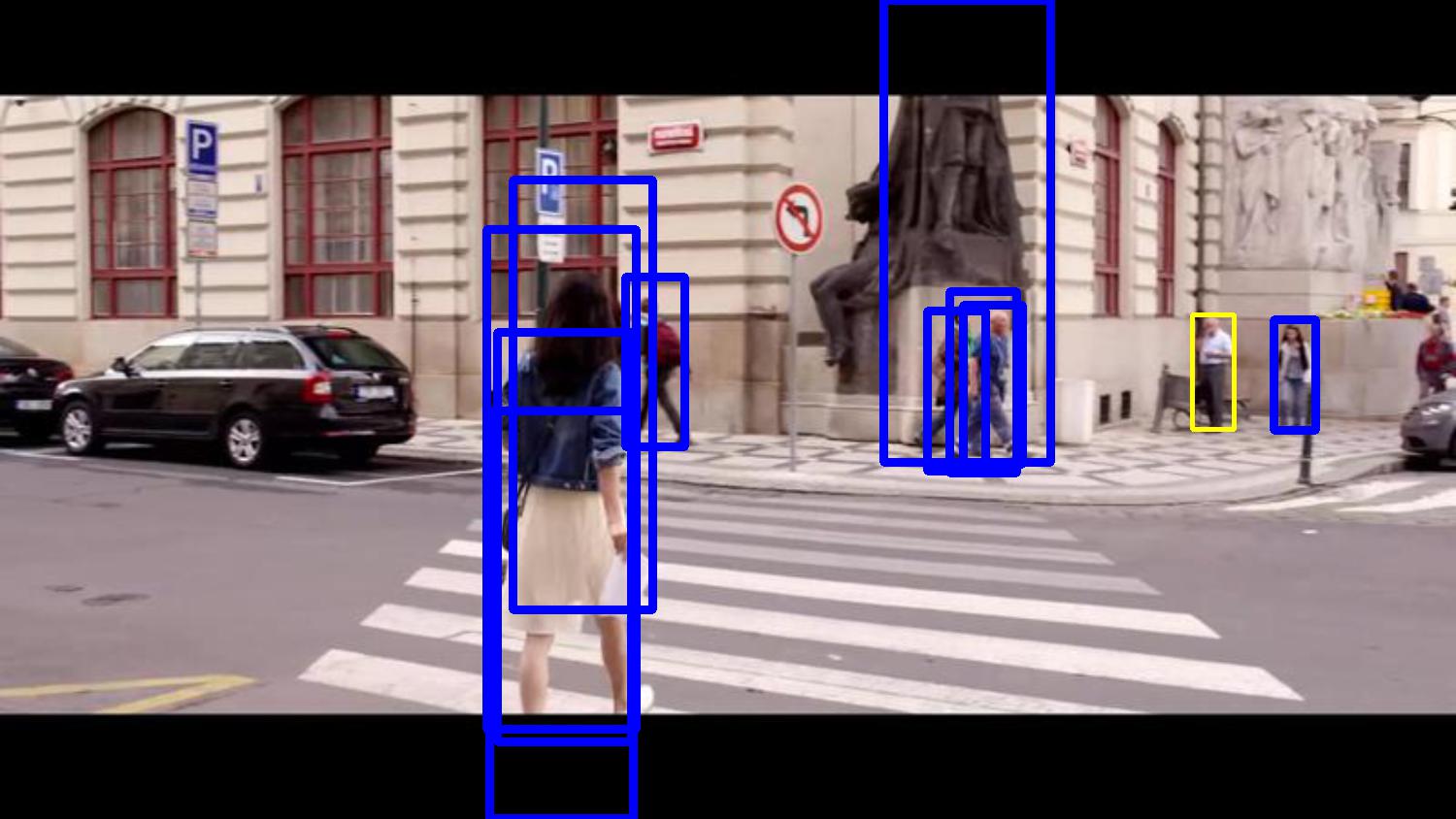}&
\includegraphics[ scale=0.038,]{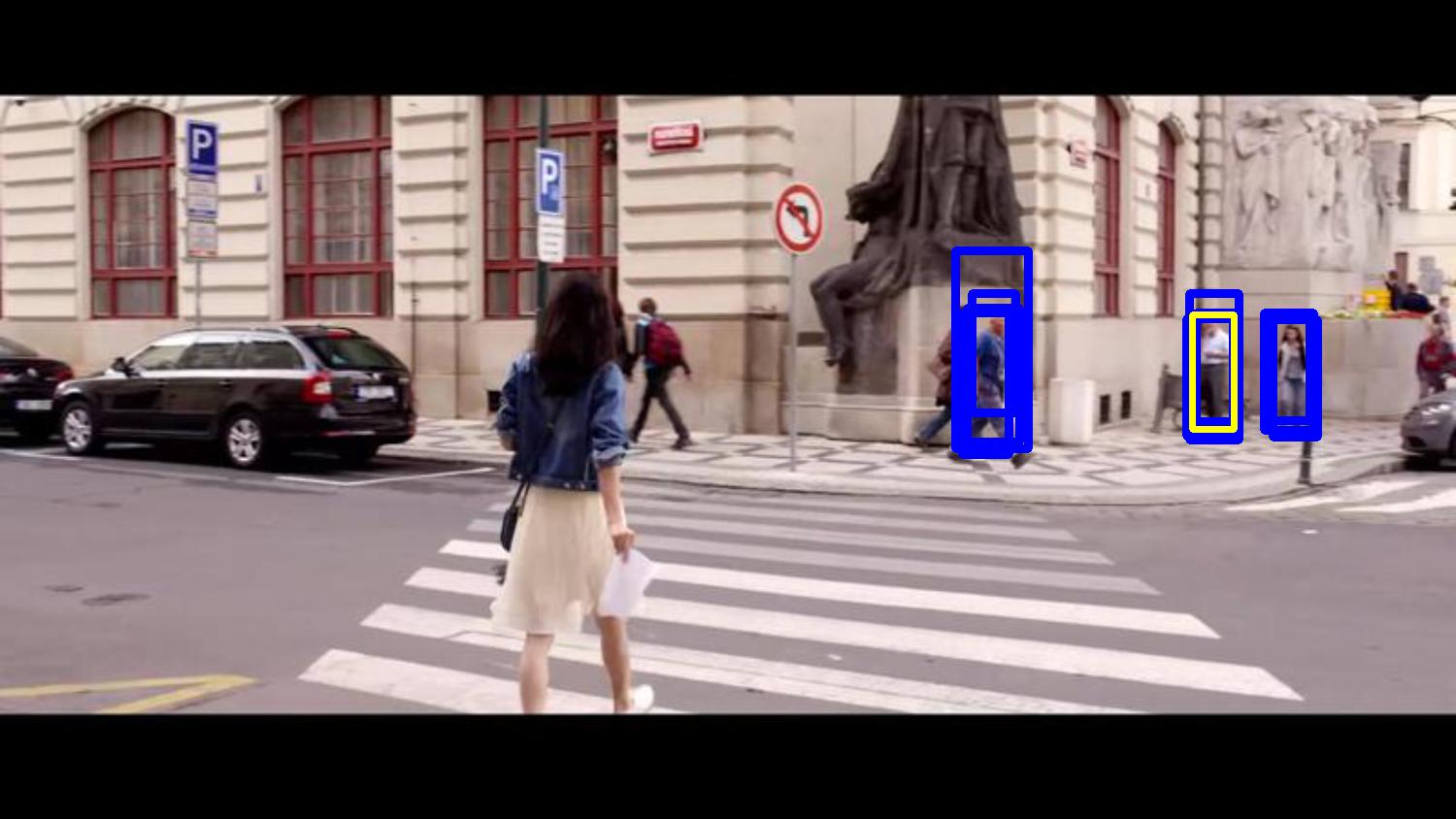} & 

\includegraphics[ scale=0.038]{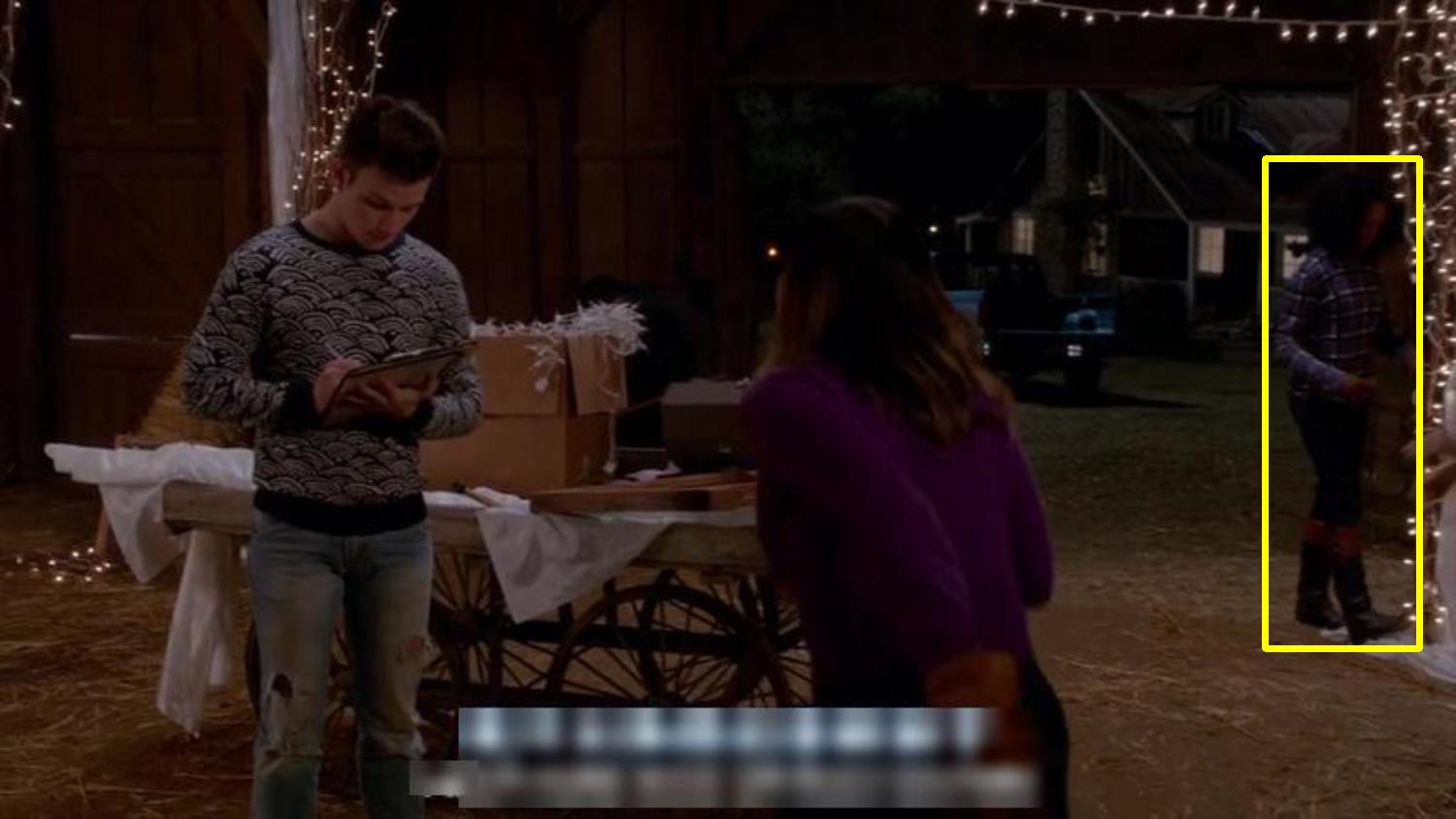}&
\includegraphics[scale=0.038, ]{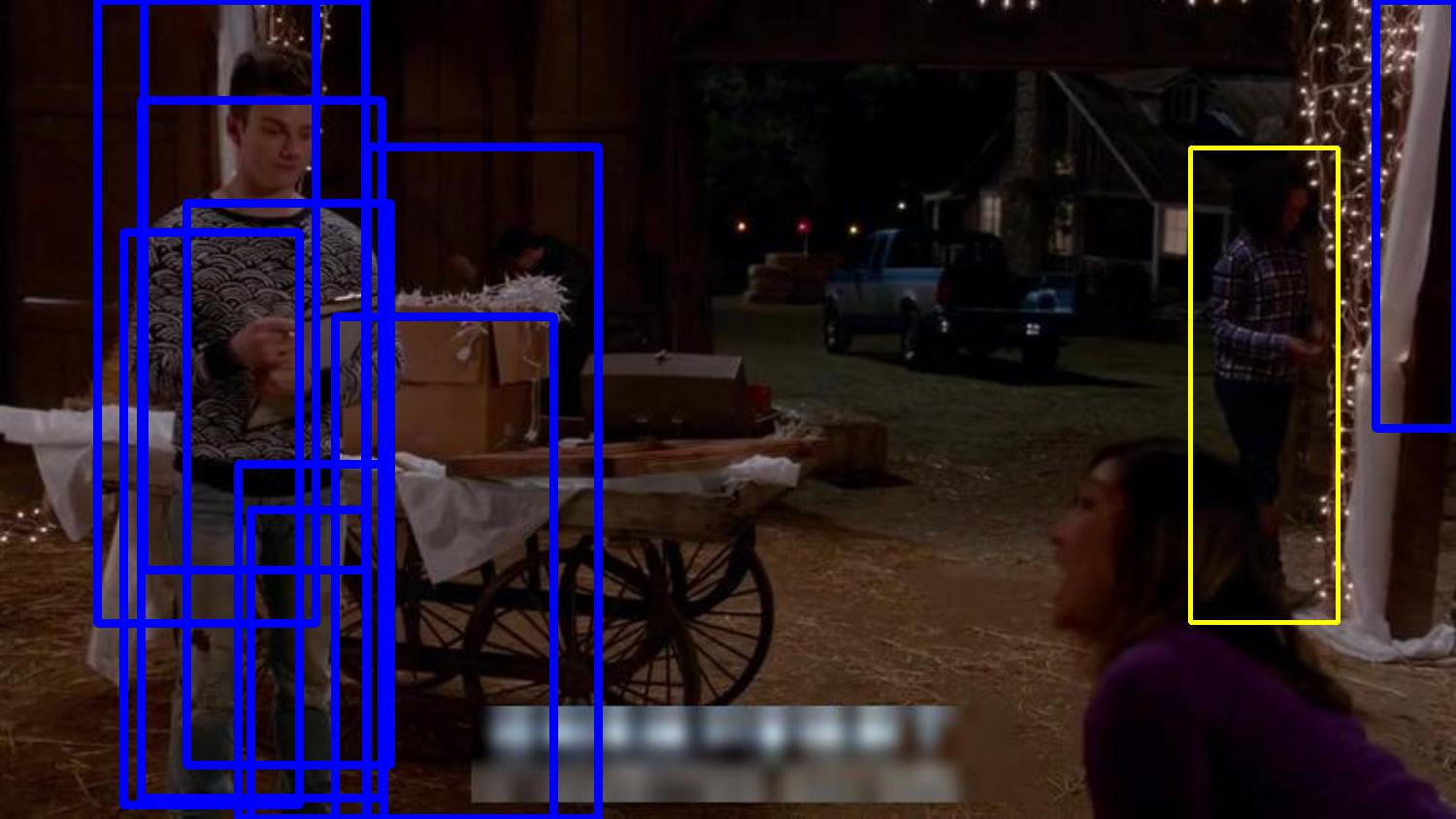}&
\includegraphics[ scale=0.038,]{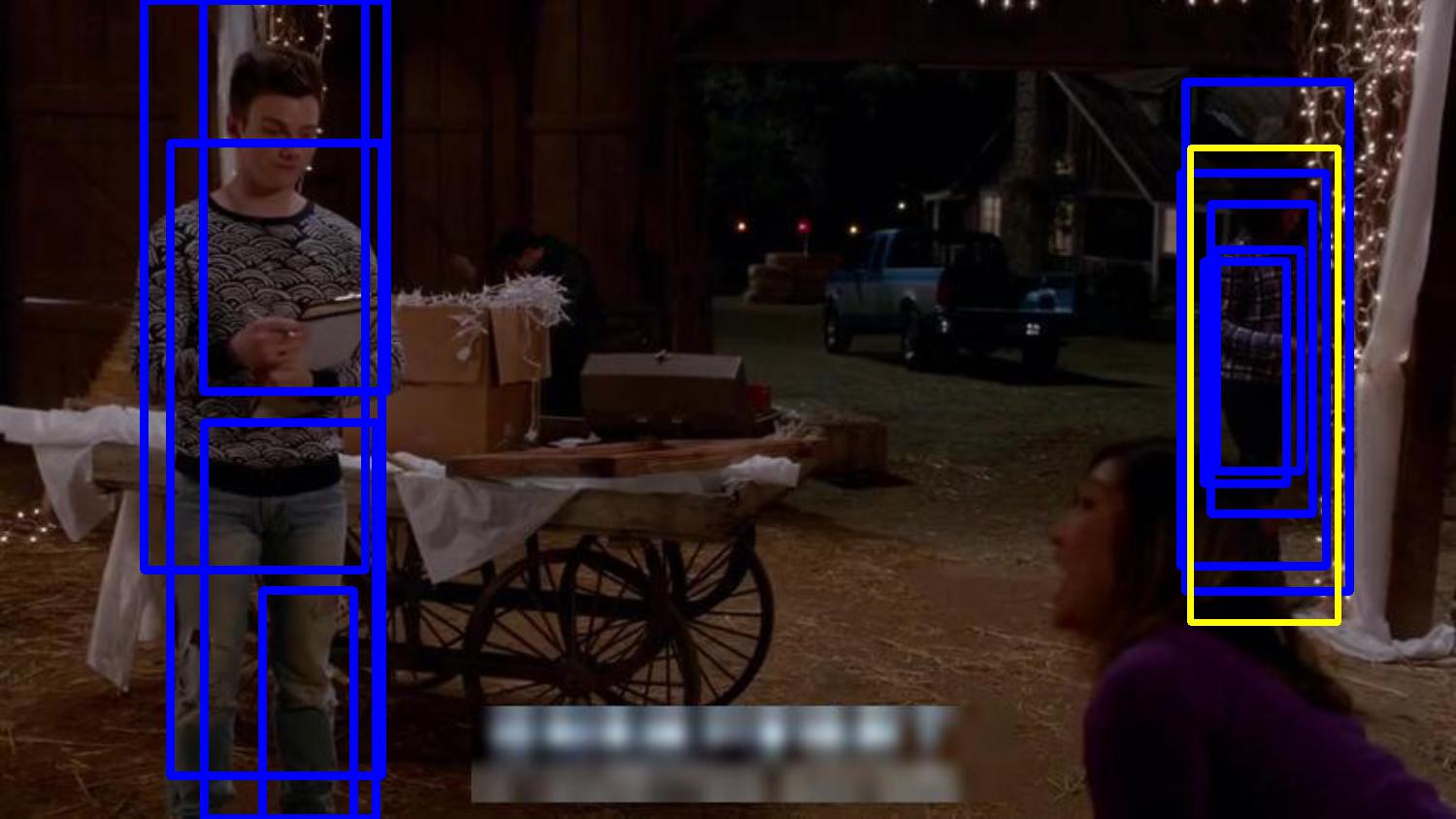}\\

%\cmidrule(r){1-3} %\cmidrule(l){4-6}
 \scriptsize{(a) Query} & \scriptsize{(b) RPN} & \scriptsize{(c) QRPN} & \scriptsize{(a) Query} & \scriptsize{(b) RPN} & \scriptsize{(c) QRPN}

\end{tabular}
%%%%%%%%%%%%%%%%%%%%%%%%%%%%%%%%%%%%%%%%%%%%%%%%%%%%%%%%%
\vspace{-0.2cm}
\caption{\textbf{\emph{Top-10} region proposals} given by RPN and QRPN. Ground-truth boxes are in yellow, output region proposals are in blue. (a) Query images with the queried person ground-truth box, (b) Gallery images with RPN proposals (c) Gallery images with QRPN proposals.
}
\vspace{-0.5cm}
\label{fig:results_proposals}
\end{center}
\end{figure}
%\egroup

%%%%%%%%%%%%%%%%%%%%%%%%%%%%%%%%%

\begin{figure}[t!]
\begin{center}
   
	\includegraphics[trim=0cm 0.8cm 0cm 1.4cm, clip=true, width=0.5\linewidth]{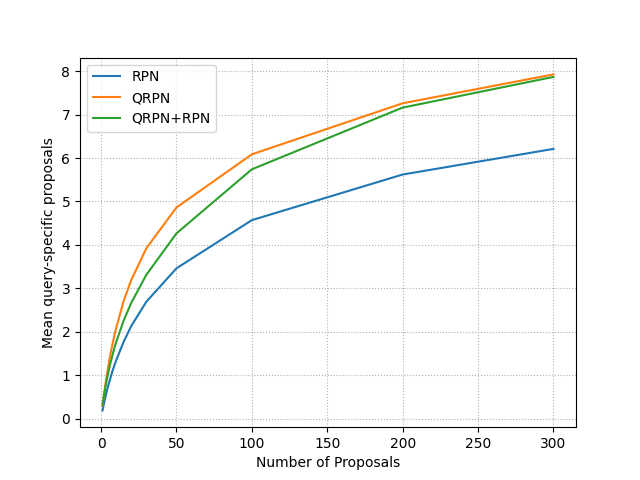}
	  \caption{ 
	  Average query-specific proposals in top-N proposals for the RPN, QRPN and RPN+QRPN sub-networks.}
	\label{fig:qrpn_proposals}
    
\end{center}
\end{figure}
%%%%%%%%%%%%%%%%%%%%%%%%%%%%%%%%%
% \bgroup
\tabcolsep 1.0pt
\renewcommand{\arraystretch}{0.5}

\newlength{\cuhkfigh}
\setlength{\cuhkfigh}{1.1cm}

\begin{figure*}[t]
\begin{center}
%%%%%%%%%%%%%%%%%%%%%%%%%%%%%%%%%%%%%%%%%%%%%%%%%%%%%%%%%
\begin{tabular}{ccc|ccc}
%probe 

\includegraphics[trim=0cm 0cm 0cm 0cm, clip=true, height=\cuhkfigh]{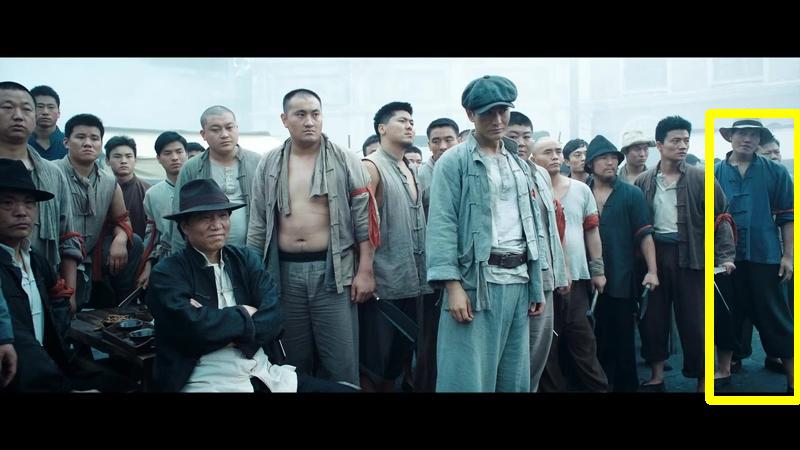}&
\includegraphics[trim=0cm 0cm 0cm 0cm, clip=true, height=\cuhkfigh]{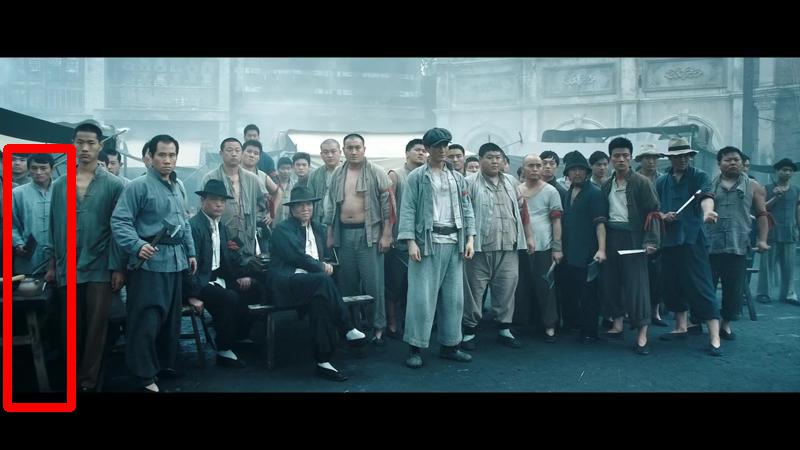}&
\includegraphics[trim=0cm 0cm 0cm 0cm, clip=true, height=\cuhkfigh]{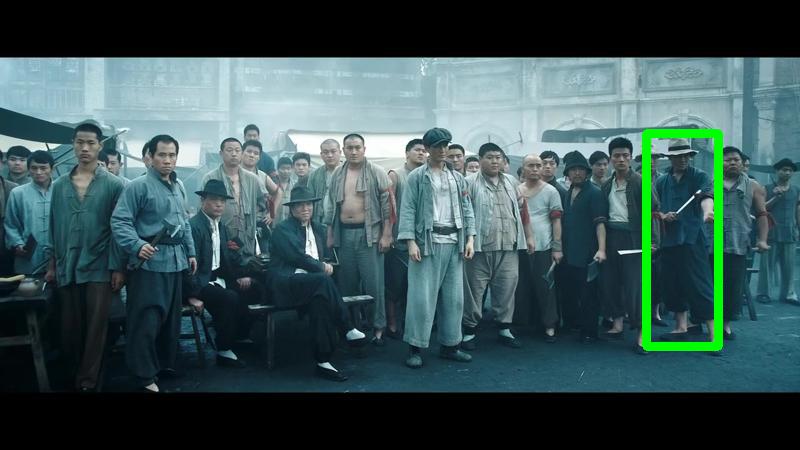} &

\includegraphics[trim=0cm 0cm 0cm 0cm, clip=true, height=\cuhkfigh]{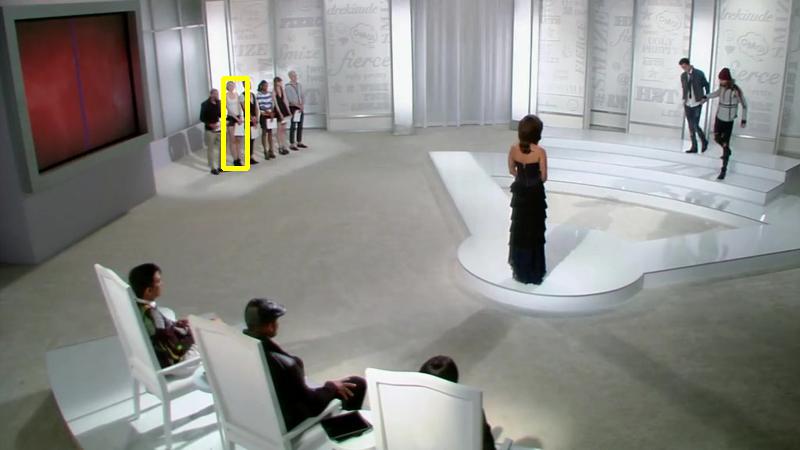}&
\includegraphics[trim=0cm 0cm 0cm 0cm, clip=true, height=\cuhkfigh]{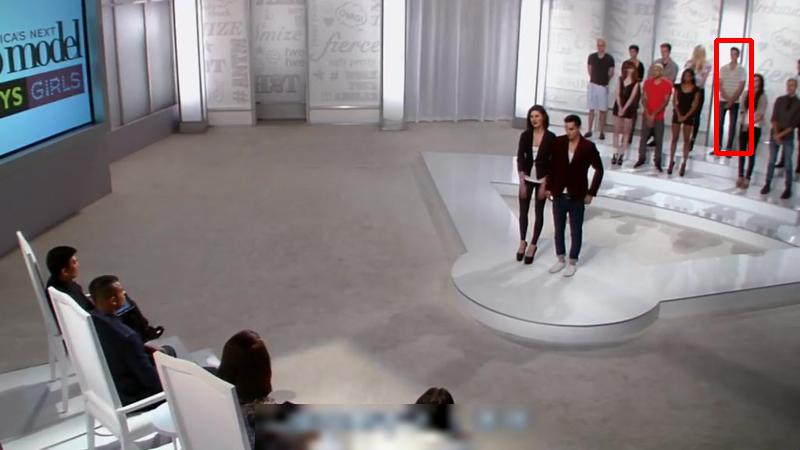}&
\includegraphics[trim=0cm 0cm 0cm 0cm, clip=true, height=\cuhkfigh]{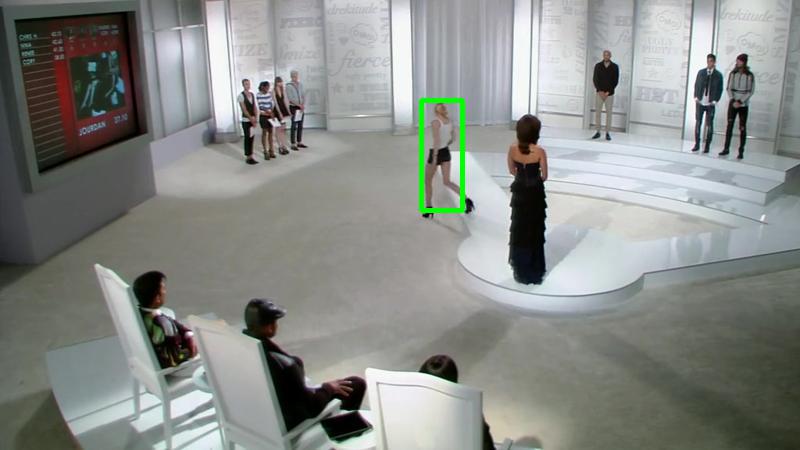}\\

% %probe no 688
\includegraphics[trim=3cm 9cm 0cm 9cm, clip=true, height=\cuhkfigh]{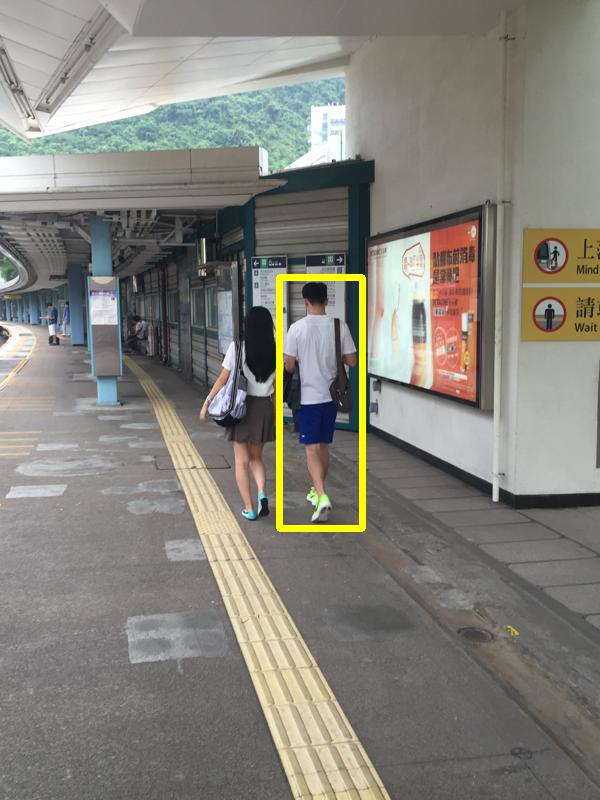}&
\includegraphics[trim=0cm 12.3cm 0cm 4cm, clip=true, height=\cuhkfigh]{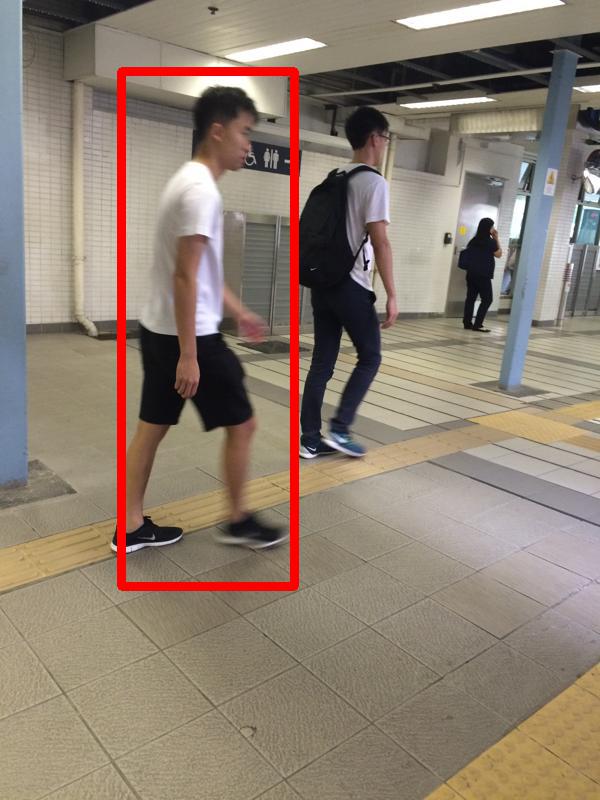}&
\includegraphics[trim=0cm 4.3cm 0cm 12cm, clip=true, height=\cuhkfigh]{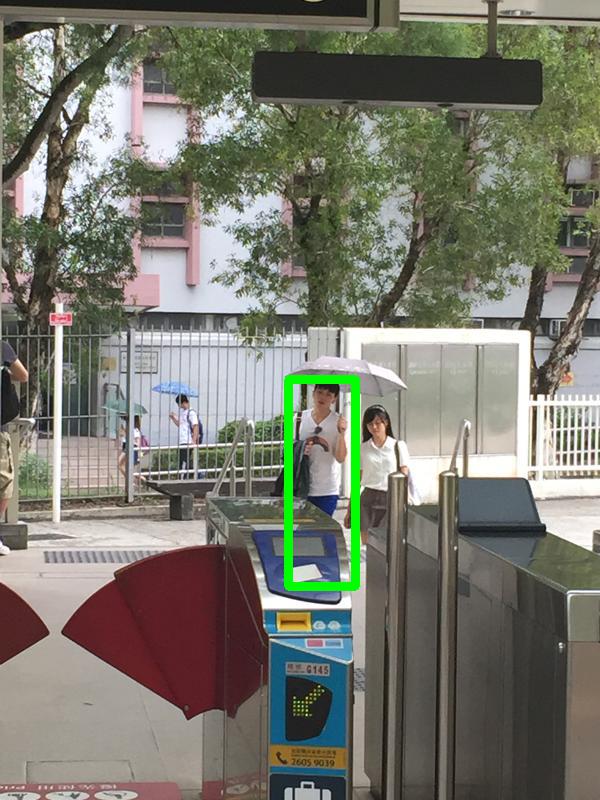}
&

\includegraphics[trim=0cm 0cm 0cm 0cm, clip=true, height=\cuhkfigh]{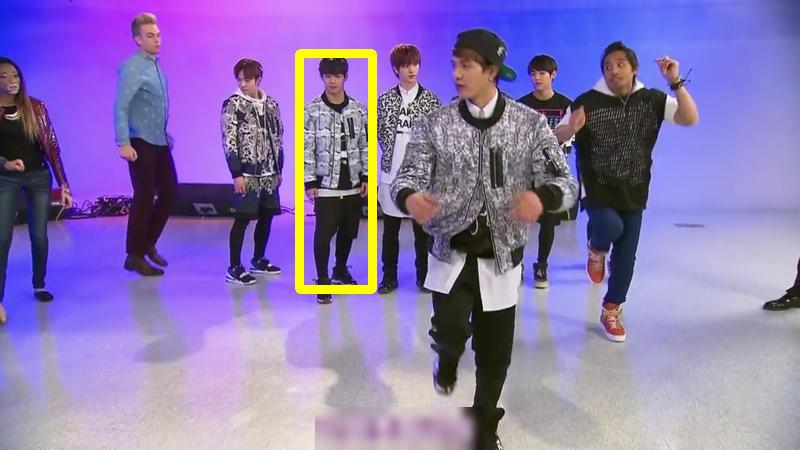}&
\includegraphics[trim=0cm 0cm 0cm 0cm, clip=true, height=\cuhkfigh]{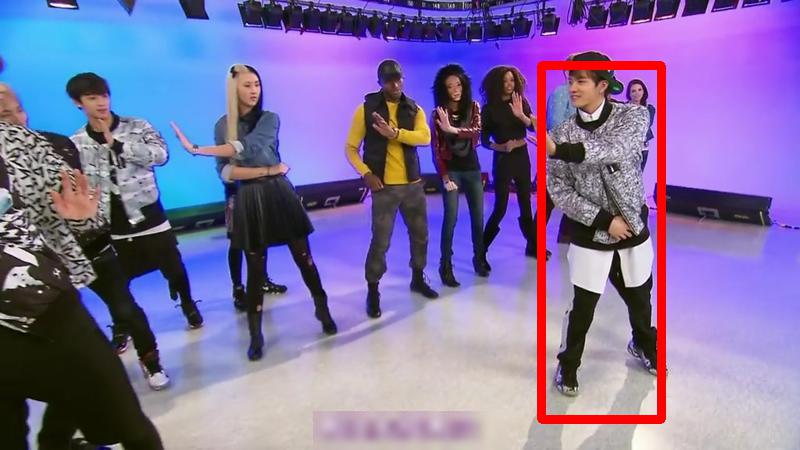}&
\includegraphics[trim=0cm 0cm 0cm 0cm, clip=true, height=\cuhkfigh]{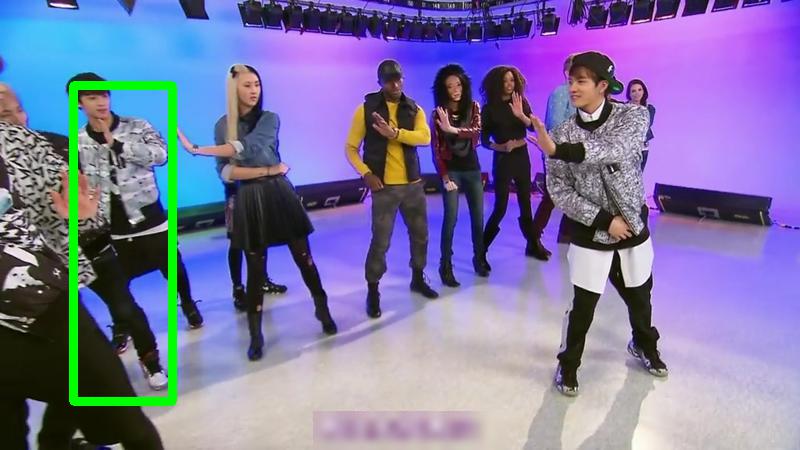}\\

\includegraphics[trim=0cm 0cm 0cm 0cm, clip=true, height=\cuhkfigh]{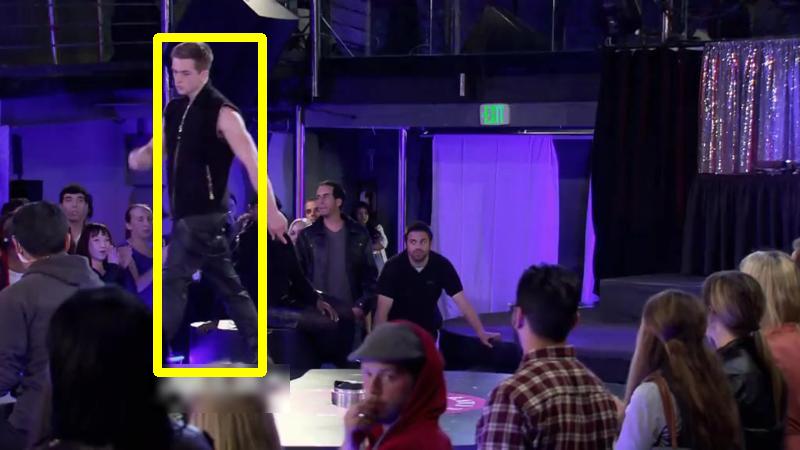}&
\includegraphics[trim=0cm 0cm 0cm 0cm, clip=true, height=\cuhkfigh]{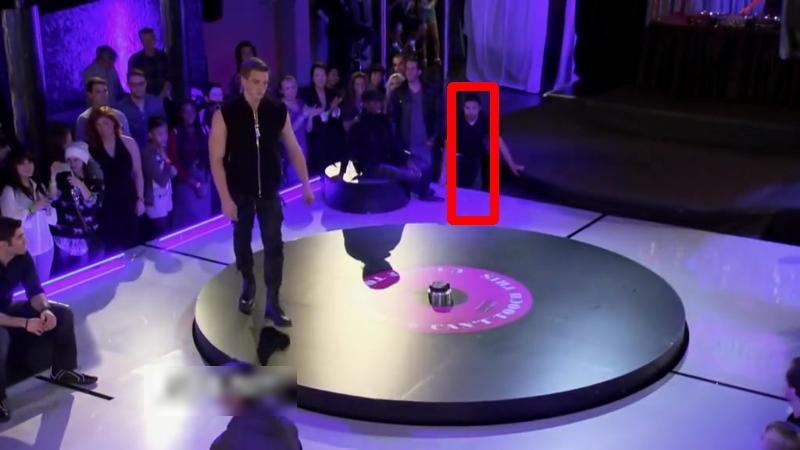}&
\includegraphics[trim=0cm 0cm 0cm 0cm, clip=true, height=\cuhkfigh]{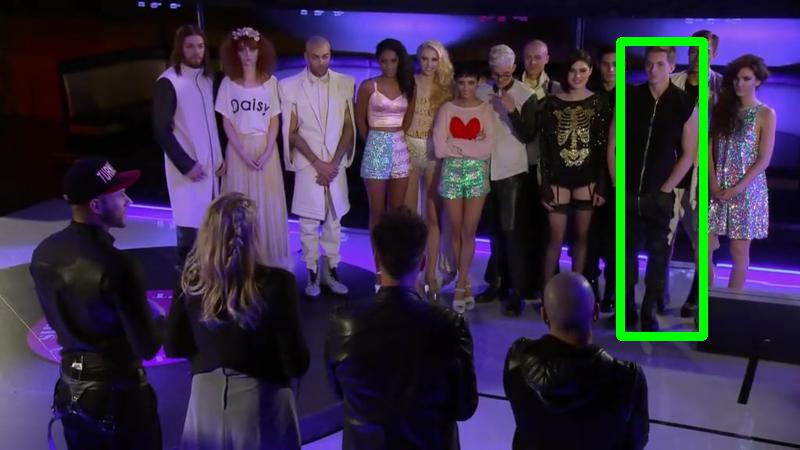} &

\includegraphics[trim=0cm 0cm 0cm 0cm, clip=true, height=\cuhkfigh]{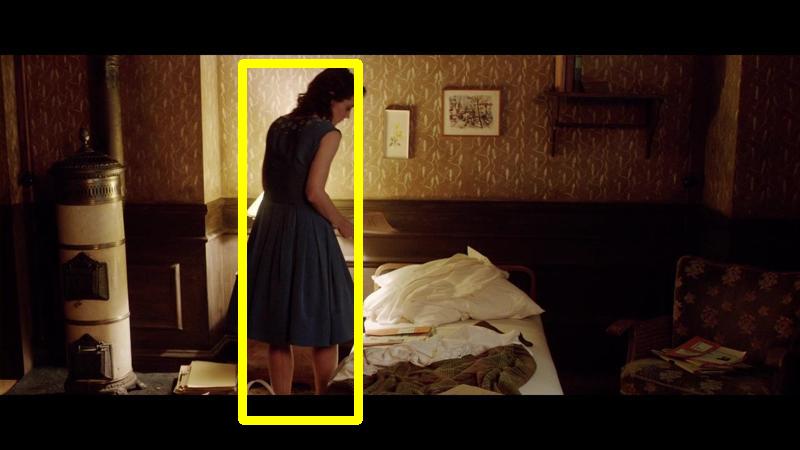}&
\includegraphics[trim=0cm 0cm 0cm 0cm, clip=true, height=\cuhkfigh]{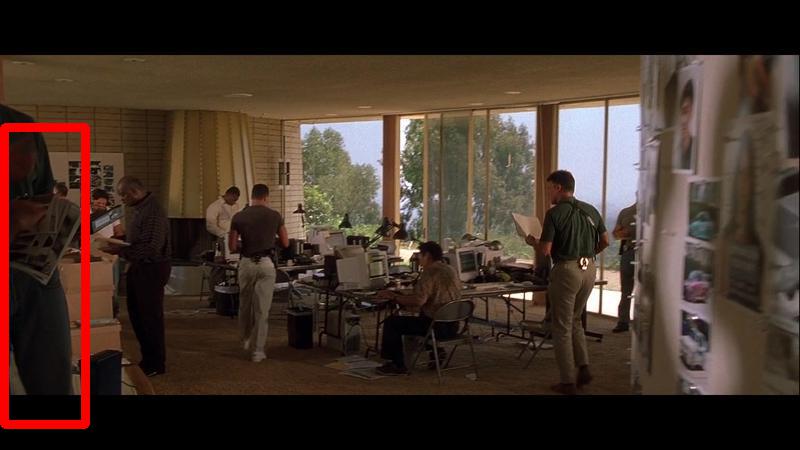}&
\includegraphics[trim=0cm 0cm 0cm 0cm, clip=true, height=\cuhkfigh]{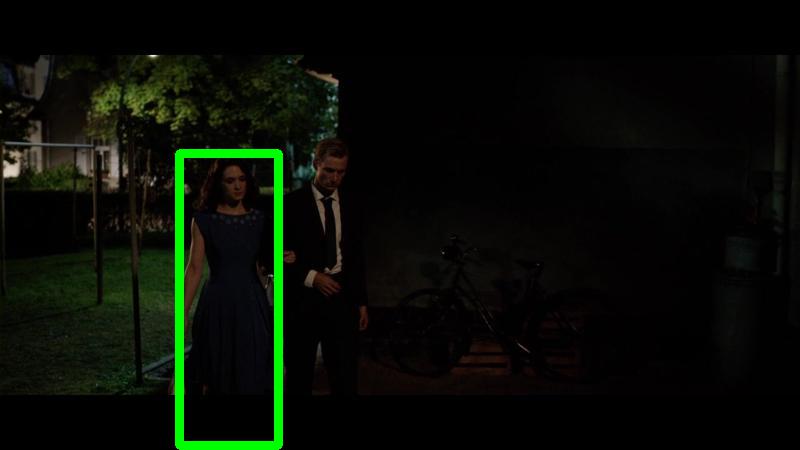}\\

\scriptsize {(a) Query }& \scriptsize {(b) OIM\ddag\ Top-1 }& \scriptsize {(c) QGN Top-1 }&\scriptsize { (a) Query} & \scriptsize {(b) OIM\ddag\ Top-1} & \scriptsize {(c) QGN Top-1}

\end{tabular}
%%%%%%%%%%%%%%%%%%%%%%%%%%%%%%%%%%%%%%%%%%%%%%%%%%%%%%%%%
\vspace{-0.2cm}
\caption{
Qualitative \emph{Top-1} person search results for a number of challenging examples.
For each example, we show (a) the query images with the bounding box of the query-person, in yellow, (b) their corresponding output matches given by the baseline OIM,
and (c) results of our proposed QGN. Red bounding boxes are failures, green ones represent correct matches.
\vspace{-0.5cm}
}

\label{fig:results}
\end{center}
\end{figure*}
%\egroup

%%%%%%%%%%%%%%%%%%%%%%%%%%%%%%%%%

%%%%%%%%%%%%%%%%%%%%%%%%%%%%%%%%%
% \bgroup
\tabcolsep 1.0pt
\renewcommand{\arraystretch}{0.5}

\newlength{\cuhkfigheightfail}
\setlength{\cuhkfigheightfail}{1.21cm}

\begin{figure*}[t]
\begin{center}
%%%%%%%%%%%%%%%%%%%%%%%%%%%%%%%%%%%%%%%%%%%%%%%%%%%%%%%%%
\begin{tabular}{cc|cc|cc}
\includegraphics[trim=0cm 8.3cm 0cm 8cm, clip=true, height=\cuhkfigheightfail]{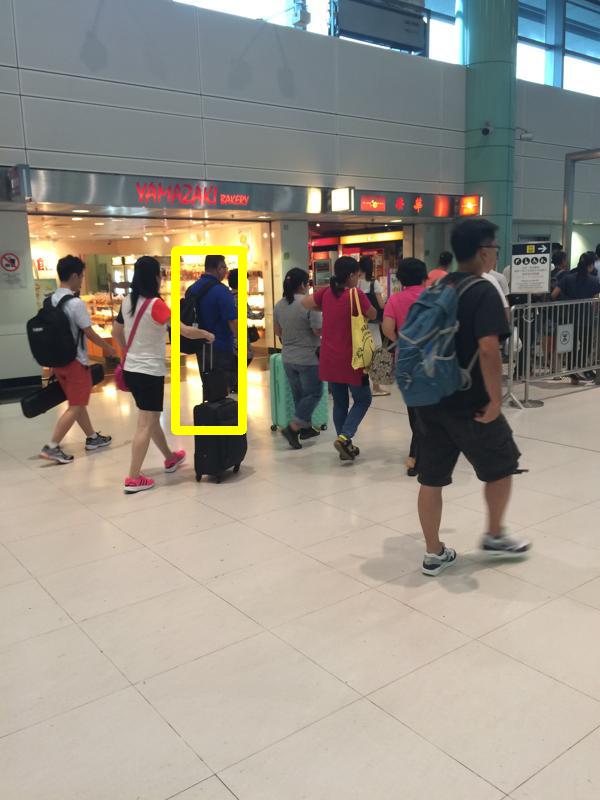}&
\includegraphics[trim=0cm 8.3cm 0cm 8cm, clip=true, height=\cuhkfigheightfail]{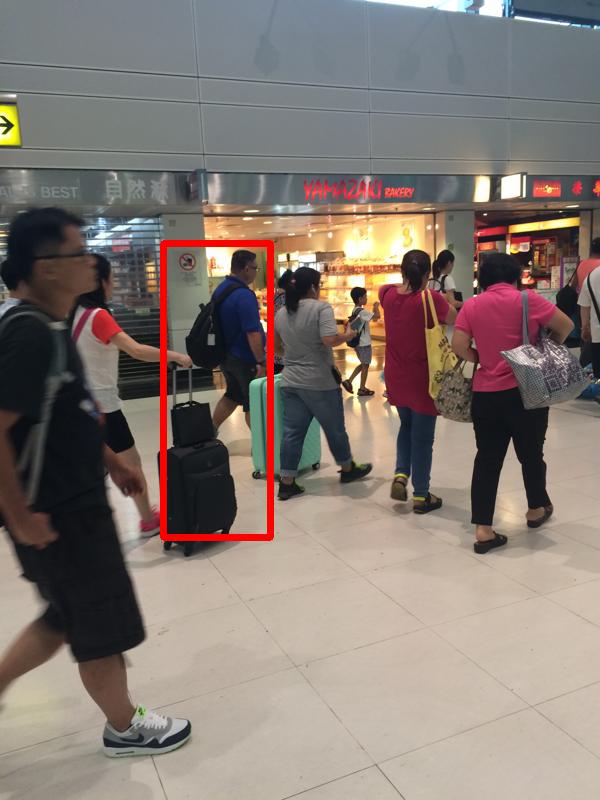}&
\includegraphics[trim=0cm 14cm 0cm 1cm, clip=true, height=\cuhkfigheightfail]{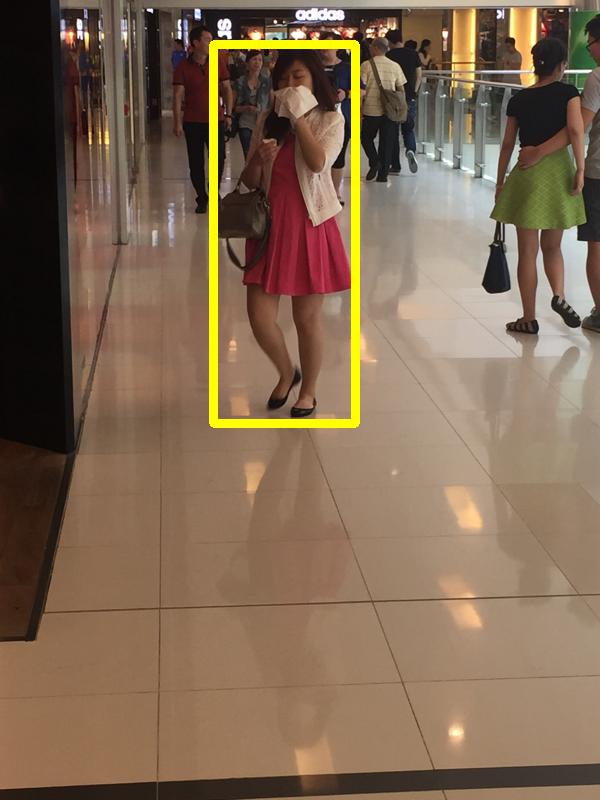}&
\includegraphics[trim=0cm 12cm 0cm 0cm, clip=true, height=\cuhkfigheightfail]{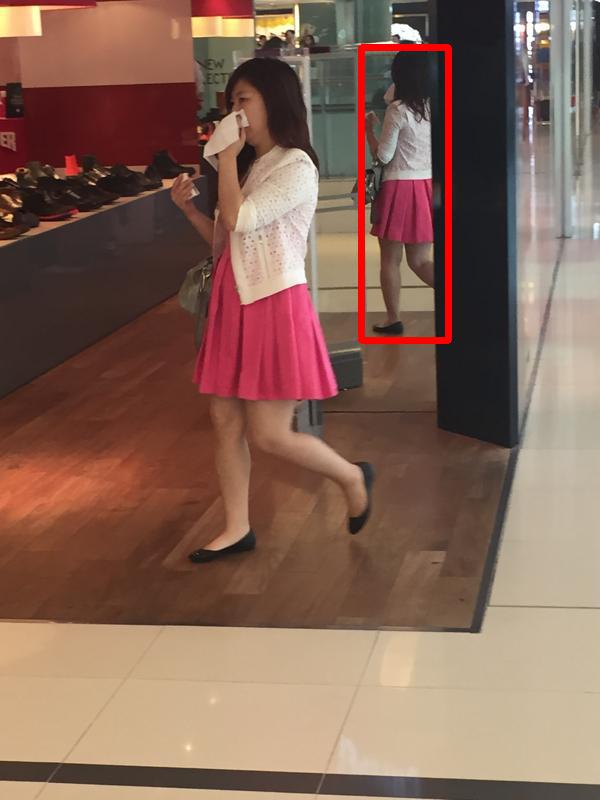}&
\includegraphics[trim=0cm 0cm 0cm 0cm, clip=true, height=\cuhkfigheightfail]{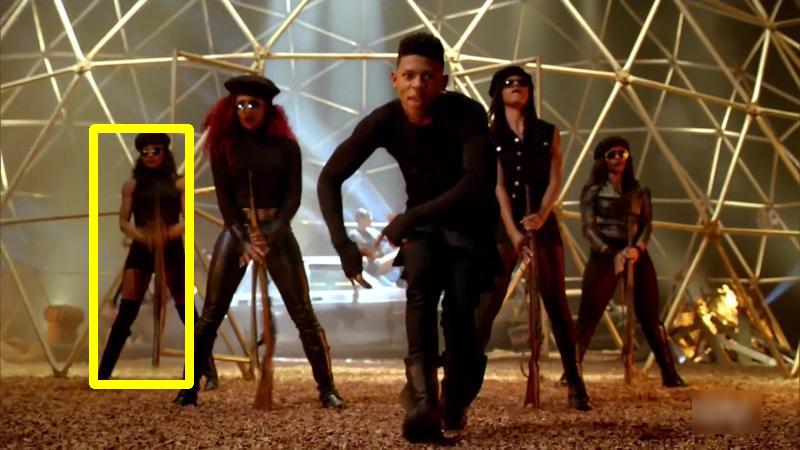}&
\includegraphics[trim=0cm 0cm 0cm 0cm, clip=true, height=\cuhkfigheightfail]{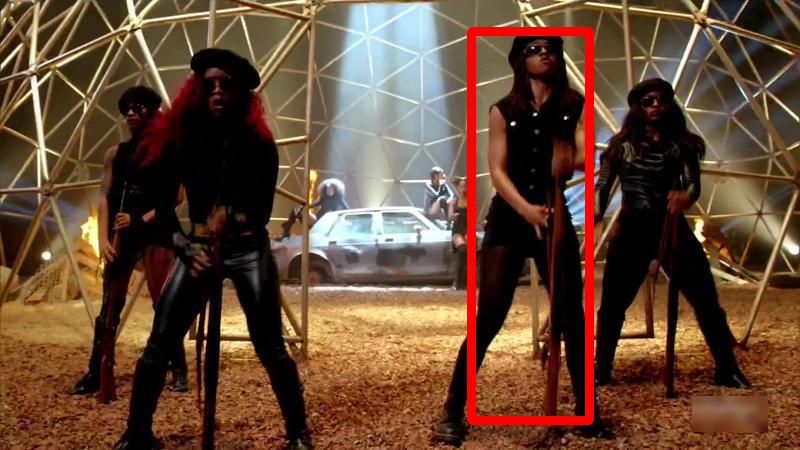}

\\

\scriptsize {Query} & \scriptsize {QGN Top-1} & \scriptsize {Query} & \scriptsize {QGN Top-1} & \scriptsize {Query} & \scriptsize {QGN Top-1}\\
\multicolumn{2}{c}{\scriptsize {(a)}} & \multicolumn{2}{c}{\footnotesize {(b)}} & \multicolumn{2}{c}{\scriptsize {(c)}}
\end{tabular}
%%%%%%%%%%%%%%%%%%%%%%%%%%%%%%%%%%%%%%%%%%%%%%%%%%%%%%%%%
\vspace{-0.3cm}
\caption{
Typical failures:
(a) localization error,
(b) missing annotation,
and (c) a challenging example with similarly-looking people.
%\vspace{-0.5cm}
%or low visibility.
}
\vspace{-0.6cm}
\label{fig:results_fail}
\end{center}
\end{figure*}
%\egroup

%%%%%%%%%%%%%%%%%%%%%%%%%%%%%%%%%

\subsubsection{{Qualitative results}}

First we compare the standard RPN~\cite{ren2015faster} Vs.\ the proposed QRPN, then we compare OIM and QGN results.

\noindent \textbf{RPN Vs.\ QRPN Proposals.}
Fig.~\ref{fig:results_proposals} illustrates region proposals by the RPN Vs.\ the proposed QRPN. Given a query-gallery image pair, in column \textit{(a)} we show the query images with the person bounding boxes (in yellow). In columns \textit{(b)} and \textit{(c)} we illustrate the top 10 region proposals in the gallery by RPN and QRPN, respectively. Note that the proposals by the RPN are on any person in the image, as it is trained for generic person detection. By contrast, the QRPN proposals in column \textit{(c)} are query-guided and are focused on those people which mostly resemble the queried person. Specific examples are the second row/left panel and the third row/right panel, where QRPN specifically proposes people wearing clothes of the same color, and the last row/right panel where RPN fails due to contrast challenges while QRPN leverages the query person pattern and successfully estimates regions over it. 

We support the qualitative result with Fig.~\ref{fig:qrpn_proposals}, i.e.\ a plot of the number of query-specific proposals (y-axis) among the top-N proposals (x-axis). A query-specific proposal is one that has IoU $>= 0.5$ with the target, one which serves to detect the queried person.
Note how QRPN and QRPN+RPN consistently provide more query-specific proposals than the standard RPN. Additionally, training with both QRPN and RPN sub-networks results in better performances.

\noindent \textbf{OIM Vs.\ QGN.}
Fig.~\ref{fig:results} illustrates some challenging queries (column \textit{(a)}) and gallery images, where these are searched for, either with OIM (column \textit{(b)}) or QGN (column \textit{(c)}). Top-1 search results are reported. Note how QGN retrieves a query person from a crowd (first row / left panel), distinguishes a query person from similarly dressed ones (second row / right panel), and also re-identifies the query in low contrast and illumination conditions (third row / right panel). 

In Fig.~\ref{fig:results_fail}, we illustrate typical failure cases of QGN.
In \textit{(a)}, QGN  successfully retrieves the 
correct person, but the bounding box is poorly aligned (IoU $< 0.5$). \textit{(b)} is an interesting case of missing annotation for the target person, i.e.\ QGN detects the reflection of the girl in the mirror, which is considered false positive. \textit{(c)} is challenging due to the similar appearance and low visibility of the people.

\section{Conclusion and Future Work}
This work has addressed, for the first time, few-shot fine-grained classification and person search with a unified Query-Guided Network (QGN). Uniting best practices from the two tasks has allowed QGN to define a novel state-of-the-art in few-shot fine-grained classification and to be on par with it for person search. A second contribution has been to propose query guidance via three components, which may be plugged-in at various stages of classification and detection models. Query guidance is novel for few-shot fine-grained classification, and it has been shown effective both quantitatively and qualitatively. In person search, query-guidance had been the novel introduction of our work~\cite{munjal2019cvpr}, now adopted by various state-of-the-art techniques, which we re-state here as effective. A drawback of our approach is its computational complexity which is due to the interaction of a pair of images at all levels in the network, notably in the Siamese QSSE network. 
In future work, following the spirit of a unified query-guided framework, we plan to research few-shot fine-grained detection, for which the query-guided proposal network module of QGN may also be relevant. 

\section{Acknowledgments}
This work is partially supported by Sapienza (Bandi d'Ateneo) and by  the  project of the Italian  Ministry  of  Education, Universities  and  Research (MIUR) ``Dipartimenti di Eccellenza 2018-2022''.

\bibliography{bibfile}

\begin{thebibliography}{10}
\expandafter\ifx\csname url\endcsname\relax
  \def\url#1{\texttt{#1}}\fi
\expandafter\ifx\csname urlprefix\endcsname\relax\def\urlprefix{URL }\fi
\expandafter\ifx\csname href\endcsname\relax
  \def\href#1#2{#2} \def\path#1{#1}\fi

\bibitem{Kim_2021_CVPR}
H.~Kim, S.~Joung, I.-J. Kim, K.~Sohn, Prototype-guided saliency feature
  learning for person search, in: Proceedings of the IEEE/CVF Conference on
  Computer Vision and Pattern Recognition (CVPR), 2021, pp. 4865--4874.

\bibitem{Wang_2020_CVPR}
Y.~Wang, C.~Xu, C.~Liu, L.~Zhang, Y.~Fu, Instance credibility inference for
  few-shot learning, in: Proceedings of the IEEE/CVF Conference on Computer
  Vision and Pattern Recognition (CVPR), 2020, pp. 12836--12845.

\bibitem{xiao2017joint}
T.~Xiao, S.~Li, B.~Wang, L.~Lin, X.~Wang, Joint detection and identification
  feature learning for person search, in: Proceedings of the IEEE Conference on
  Computer Vision and Pattern Recognition (CVPR), 2017, pp. 3415--3424.

\bibitem{tang2020revisiting}
D.~W. L.~Tang, B.~Hariharan, Revisiting pose-normalization for fine-grained
  few-shot recognition, in: Proceedings of the IEEE/CVF Conference on Computer
  Vision and Pattern Recognition (CVPR), 2020, pp. 14352--14361.

\bibitem{Chen_2020_CVPR}
D.~Chen, S.~Zhang, J.~Yang, B.~Schiele, Norm-aware embedding for efficient
  person search, in: Proceedings of the IEEE/CVF Conference on Computer Vision
  and Pattern Recognition (CVPR), 2020, pp. 12615--12624.

\bibitem{Dong_2020_CVPR}
W.~Dong, Z.~Zhang, C.~Song, T.~Tan, Instance guided proposal network for person
  search, in: Proceedings of the IEEE/CVF Conference on Computer Vision and
  Pattern Recognition (CVPR), 2020, pp. 2585--2594.

\bibitem{snell2017nips}
J.~Snell, K.~Swersky, R.~Zemel, Prototypical networks for few-shot learning,
  in: Proceedings of the 31st International Conference on Neural Information
  Processing Systems, 2017, pp. 4077--4087.

\bibitem{Mangla2020ChartingTR}
P.~Mangla, N.~Kumari, A.~Sinha, M.~Singh, B.~Krishnamurthy, V.~N.
  Balasubramanian, Charting the right manifold: Manifold mixup for few-shot
  learning, in: Proceedings of the IEEE/CVF Winter Conference on Applications
  of Computer Vision (WACV), 2020, pp. 2218--2227.

\bibitem{chen2019closerlook}
W.-Y. Chen, Y.-C. Liu, Z.~Kira, Y.-C. Wang, J.-B. Huang, A closer look at
  few-shot classification, in: International Conference on Learning
  Representations (ICLR), URL \url{https://openreview.net/forum?id=HkxLXnAcFQ},
  2019.

\bibitem{WelinderEtal2010}
C.~Wah, S.~Branson, P.~Welinder, P.~Perona, S.~Belongie, The caltech-ucsd
  birds-200-2011 dataset, Tech. Rep. CNS-TR-2011-001, California Institute of
  Technology (2011).

\bibitem{KrauseStarkDengFei-Fei_3DRR2013}
J.~Krause, M.~Stark, J.~Deng, L.~Fei-Fei, 3d object representations for
  fine-grained categorization, in: IEEE International Conference on Computer
  Vision Workshops (ICCVW), 2013, pp. 554--561.

\bibitem{aircraftdataset}
S.~Maji, J.~Kannala, E.~Rahtu, M.~Blaschko, A.~Vedaldi, Fine-grained visual
  classification of aircraft, Tech. rep. (2013).
\newblock \href {http://arxiv.org/abs/1306.5151} {\path{arXiv:1306.5151}}.

\bibitem{Khosla2012NovelDF}
A.~Khosla, N.~Jayadevaprakash, B.~Yao, L.~Fei-Fei, Novel dataset for
  fine-grained image categorization, in: Workshop on Fine-Grained Visual
  Categorization (FGVC), IEEE Conference on Computer Vision and Pattern
  Recognition (CVPR), 2011.

\bibitem{10.1109/CVPR.2006.42}
M.~E. Nilsback, A.~Zisserman, A visual vocabulary for flower classification,
  in: IEEE Computer Society Conference on Computer Vision and Pattern
  Recognition (CVPR), 2006, pp. 1447--1454.

\bibitem{8099840}
L.~Zheng, H.~Zhang, S.~Sun, M.~Chandraker, Y.~Yang, Q.~Tian, Person
  re-identification in the wild, in: Proceedings of the IEEE Conference on
  Computer Vision and Pattern Recognition (CVPR), 2017, pp. 1367--1376.

\bibitem{munjal2019cvpr}
B.~Munjal, S.~Amin, F.~Tombari, F.~Galasso, Query-guided end-to-end person
  search, in: Proceedings of the IEEE/CVF Conference on Computer Vision and
  Pattern Recognition (CVPR), 2019, pp. 811--820.

\bibitem{sung2018cvpr}
F.~Sung, Y.~Yang, L.~Zhang, T.~Xiang, P.~H. Torr, T.~M. Hospedales, Learning to
  compare: Relation network for few-shot learning, in: Proceedings of the IEEE
  Conference on Computer Vision and Pattern Recognition (CVPR), 2018, pp.
  1199--1208.

\bibitem{ZHAO2022108880}
Z.~Zhao, Q.~Liu, W.~Cao, D.~Lian, Z.~He, Self-guided information for few-shot
  classification, Pattern Recognition 131 (2022) 108880.

\bibitem{DBLP:journals/corr/FinnAL17}
C.~Finn, P.~Abbeel, S.~Levine, Model-agnostic meta-learning for fast adaptation
  of deep networks, in: Proceedings of the 34th International Conference on
  Machine Learning (ICML), 2017, pp. 1126--1135.

\bibitem{Su2020When}
J.-C. Su, S.~Maji, B.~Hariharan, When does self-supervision improve few-shot
  learning?, in: Proceedings of the European Conference on Computer Vision
  (ECCV), 2020, pp. 645--666.

\bibitem{hou2019cross}
R.~Hou, H.~Chang, B.~Ma, S.~Shan, X.~Chen, Cross attention network for few-shot
  classification, in: Proceedings of the 33rd International Conference on
  Neural Information Processing Systems, 2019, pp. 4005--4016.

\bibitem{Zhao_2018_ECCV}
S.~Y. F.~Zhao, J.~Zhao, J.~Feng, Dynamic conditional networks for few-shot
  learning, in: Proceedings of the European Conference on Computer Vision
  (ECCV), 2018, pp. 19--35.

\bibitem{matching2016}
O.~Vinyals, C.~Blundell, T.~Lillicrap, K.~Kavukcuoglu, D.~Wierstra., Matching
  networks for one shot learning, in: Proceedings of the 30th International
  Conference on Neural Information Processing Systems, 2016, pp. 3637--3645.

\bibitem{LIANG2022108662}
M.~Liang, S.~Huang, S.~Pan, M.~Gong, W.~Liu, Learning multi-level
  weight-centric features for few-shot learning, Pattern Recognition 128 (2022)
  108662.

\bibitem{ijcai2020-152}
Y.~Zhu, C.~Liu, S.~Jiang, Multi-attention meta learning for few-shot
  fine-grained image recognition, in: Proceedings of the International Joint
  Conference On Artificial Intelligence (IJCAI), 2020, pp. 1090--1096.

\bibitem{TANG2022108792}
H.~Tang, C.~Yuan, Z.~Li, J.~Tang, Learning attention-guided pyramidal features
  for few-shot fine-grained recognition, Pattern Recognition 130 (2022) 108792.

\bibitem{SunCZZZWW20}
Y.~Sun, C.~Cheng, Y.~Zhang, C.~Zhang, L.~Zheng, Z.~Wang, Y.~Wei, Circle loss:
  {A} unified perspective of pair similarity optimization, in: Proceedings of
  the IEEE/CVF Conference on Computer Vision and Pattern Recognition (CVPR),
  2020, pp. 6398--6407.

\bibitem{Chuchu2019}
C.~Han, J.~Ye, Y.~Zhong, X.~Tan, C.~Zhang, C.~Gao, N.~Sang, Re-id driven
  localization refinement for person search, in: Proceedings of the IEEE/CVF
  International Conference on Computer Vision (ICCV), 2019, pp. 9814--9823.

\bibitem{munjal2019knowledge}
B.~Munjal, F.~Galasso, S.~Amin, Knowledge distillation for end-to-end person
  search, in: Proceedings of the British Machine Vision Conference (BMVC), BMVA
  Press, 2019, pp. 31.1--31.16.

\bibitem{patcog2022108654}
C.~Liu, H.~Yang, Q.~Zhou, S.~Zheng, Making person search enjoy the merits of
  person re-identification, Pattern Recognition 127 (2022) 108654.

\bibitem{ren2015faster}
S.~Ren, K.~He, R.~Girshick, J.~Sun, Faster r-cnn: Towards real-time object
  detection with region proposal networks, in: Proceedings of the 29th
  International Conference on Neural Information Processing Systems, 2015, pp.
  91--99.

\bibitem{Yan_2021_CVPR}
Y.~Yan, J.~Li, J.~Qin, S.~Bai, S.~Liao, L.~Liu, F.~Zhu, L.~Shao, Anchor-free
  person search, in: Proceedings of the IEEE/CVF Conference on Computer Vision
  and Pattern Recognition (CVPR), 2021, pp. 7690--7699.

\bibitem{dong2020}
W.~Dong, Z.~Zhang, C.~Song, T.~Tan, Bi-directional interaction network for
  person search, in: Proceedings of the IEEE/CVF Conference on Computer Vision
  and Pattern Recognition (CVPR), 2020, pp. 2839--2848.

\bibitem{LI2021107862}
W.~Li, S.~Gong, X.~Zhu, Hierarchical distillation learning for scalable person
  search, Pattern Recognition 114 (2021) 107862.

\bibitem{cheng_2020_CVPR}
C.~Wang, B.~Ma, H.~Chang, S.~Shan, X.~Chen, Tcts: A task-consistent two-stage
  framework for person search, in: Proceedings of the IEEE/CVF Conference on
  Computer Vision and Pattern Recognition (CVPR), 2020, pp. 11952--11961.

\bibitem{Hu_2018_CVPR}
J.~Hu, L.~Shen, G.~Sun, Squeeze-and-excitation networks, in: Proceedings of the
  IEEE Conference on Computer Vision and Pattern Recognition (CVPR), 2018, pp.
  7132--7141.

\bibitem{trans2019}
L.~Qiao, Y.~Shi, J.~Li, Y.~Wang, T.~Huang, Y.~Tian, Transductive episodic-wise
  adaptive metric for few-shot learning, in: Proceedings of the IEEE/CVF
  International Conference on Computer Vision (ICCV), 2019, pp. 3603--3612.

\bibitem{conf/iccv/SelvarajuCDVPB17}
R.~R. Selvaraju, M.~Cogswell, A.~Das, R.~Vedantam, D.~Parikh, D.~Batra,
  Grad-cam: Visual explanations from deep networks via gradient-based
  localization, in: Proceedings of the IEEE International Conference on
  Computer Vision (ICCV), 2017, pp. 618--626.

\bibitem{Zhong2020}
Y.~Zhong, X.~Wang, S.~Zhang, Robust partial matching for person search in the
  wild, in: Proceedings of the IEEE/CVF Conference on Computer Vision and
  Pattern Recognition (CVPR), 2020, pp. 6826--6834.

\bibitem{DBLP:conf/aaai/HanZGSY21}
C.~Han, Z.~Zheng, C.~Gao, N.~Sang, Y.~Yang, Decoupled and memory-reinforced
  networks: Towards effective feature learning for one-step person search, in:
  Proceedings of the AAAI Conference on Artificial Intelligence, 2021, pp.
  1505--1512.

\bibitem{Lin_2017_CVPR}
T.-Y. Lin, P.~Dollar, R.~Girshick, K.~He, B.~Hariharan, S.~Belongie, Feature
  pyramid networks for object detection, in: Proceedings of the IEEE Conference
  on Computer Vision and Pattern Recognition (CVPR), 2017, pp. 2117--2125.

\bibitem{DBLP:conf/aaai/ZhangWBSY21}
X.~Zhang, X.~Wang, J.~Bian, C.~Shen, M.~You, Diverse knowledge distillation for
  end-to-end person search, in: Proceedings of the AAAI Conference on
  Artificial Intelligence, 2021, pp. 3412--3420.

\bibitem{yan2019}
Y.~Yan, Q.~Zhang, B.~Ni, W.~Zhang, M.~Xu, X.~Yang, Learning context graph for
  person search, in: Proceedings of the IEEE/CVF Conference on Computer Vision
  and Pattern Recognition (CVPR), 2019, pp. 2158--2167.

\end{thebibliography}
\end{document}